\documentclass{article} 
\usepackage{iclr2025_conference,times}

\usepackage{amsmath,amsfonts,bm}

\def\eqref#1{equation~\ref{#1}}

\def\1{\bm{1}}

\DeclareMathAlphabet{\mathsfit}{\encodingdefault}{\sfdefault}{m}{sl}
\SetMathAlphabet{\mathsfit}{bold}{\encodingdefault}{\sfdefault}{bx}{n}

\usepackage{hyperref}
\usepackage{url}
\usepackage{subcaption}
\usepackage{mwe} 
\usepackage{enumitem}
\usepackage{booktabs}
\usepackage{wrapfig}
\iclrfinalcopy

\title{Evolution of Concepts in Language Model Pre-Training}

\author{Xuyang Ge$^\dagger$,
\textbf{Wentao Shu}$^\dagger$,
\textbf{Jiaxing Wu}$^\dagger$,
\textbf{Yunhua Zhou}$^\ddagger$,
\textbf{Zhengfu He}$^\dagger$,
\textbf{Xipeng Qiu}$^\dagger$ \thanks{Corresponding author.} \\\\
$^\dagger$ \text{OpenMOSS Team, Shanghai Innovation Institute; Fudan University} \\
$^\ddagger$ \text{Shanghai AI Laboratory} \\
\texttt{xyge24@m.fudan.edu.cn, zfhe19@fudan.edu.cn} \\
}

\begin{document}

\maketitle

\begin{figure}[h]
    \centering
    \includegraphics[width=0.9\textwidth]{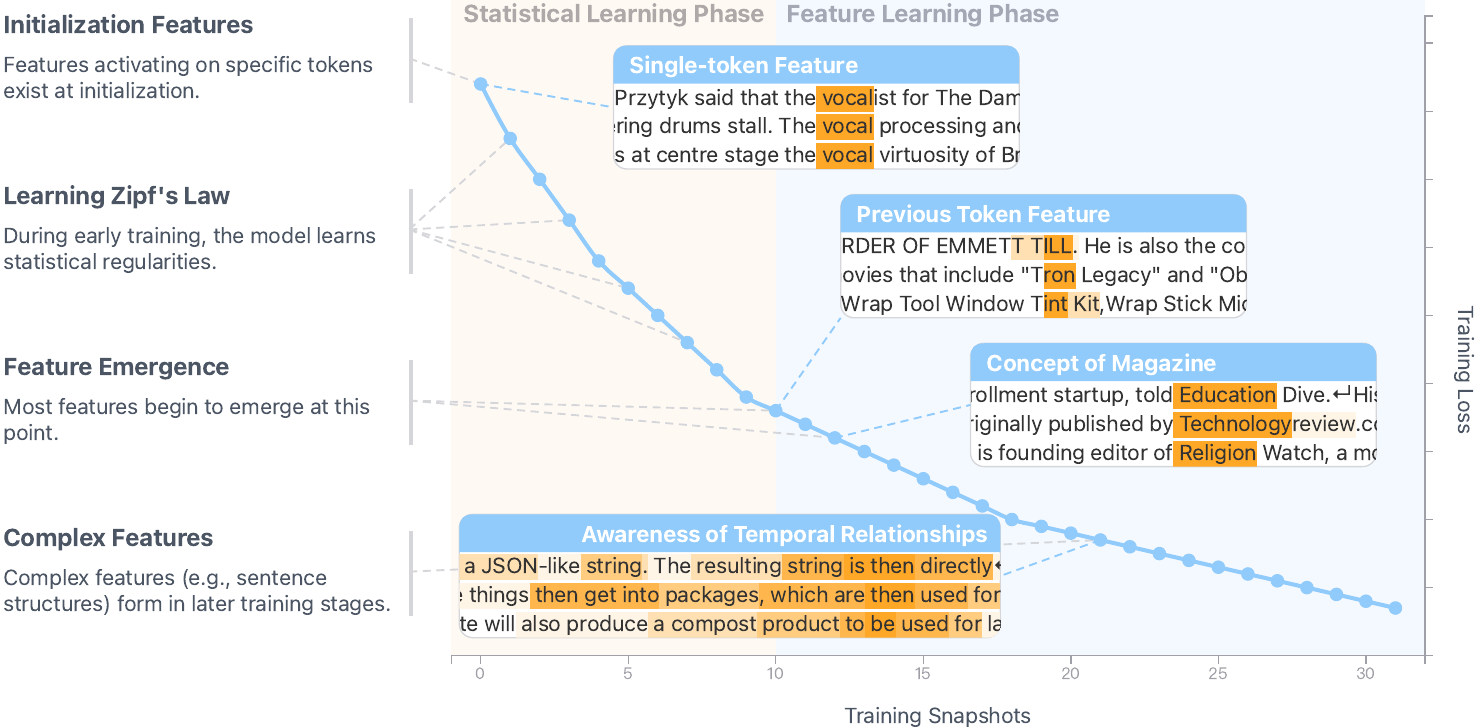}
    \label{fig:head}
\end{figure}

\begin{abstract}
Language models obtain extensive capabilities through pre-training. However, the pre-training dynamics remains a black box. In this work, we track linear interpretable feature evolution across pre-training snapshots using a sparse dictionary learning method called crosscoders. We find that most features begin to form around a specific point, while more complex patterns emerge in later training stages. Feature attribution analyses reveal causal connections between feature evolution and downstream performance. Our feature-level observations are highly consistent with previous findings on Transformer's two-stage learning process, which we term a statistical learning phase and a feature learning phase. Our work opens up the possibility to track fine-grained representation progress during language model learning dynamics. Our code is available at \url{https://github.com/OpenMOSS/Language-Model-SAEs}.
\end{abstract}

\section{Introduction}

Pre-training~\citep{radford2018gpt, devlin2019bert} has emerged as the dominant paradigm for developing frontier large language models (LLM)~\citep{openai2024gpt4, grattafiori2024llama3, yang2025qwen3}. Despite its remarkable success, the pre-training process remains largely a black box. While scaling laws~\citep{hestness2017scaling, kaplan2020scalinglaws, bahri2021explainscalinglaws} reveal the predictable relationships between compute, data, and loss, they offer limited insight into the internal reorganization occurring within the model parameters. Other theoretical frameworks on learning dynamics, including neural tangent kernel~\citep{jacot2018ntk}, information bottleneck theory~\citep{tishby2015informationbottleneck, shwartz2017information}, and singular learning theory~\citep{watanabe2009slt, lau2024llc, wang2025rllc} --- provide high-level explanations on \textit{why} we can observe generalization and grokking~\citep{power2022grokking, nanda2023grokking}. However, a fundamental question --- \textit{how} do LLMs actually develop their capabilities internally during pre-training --- has remained largely opaque.

Recent advances in mechanistic interpretability, particularly through dictionary learning methods based on sparse autoencoders (SAEs), have begun to illuminate the internal representation of neural networks~\citep{bricken2023monosemanticity, templeton2024scalingmonosemanticity, gao2025scalingsae}. By disentangling the phenomenon of superposition~\citep{elhage2022superposition}, sparse autoencoders show their capabilities in extracting millions of features from LLM activations, demonstrating that LLMs encode human-interpretable concepts as linear directions in their activation spaces. However, these analyses have predominantly focused on studying fully-trained models. Consequently, the process of how features initially emerge and evolve throughout the training process remains largely unexplored.

In this paper, we propose to \textbf{track feature evolution} across pre-training snapshots using \textbf{crosscoders}~\citep{lindsey2024crosscoder}, a variant of sparse autoencoders designed to simultaneously identify and align features from a family of correlated model activations. While originally introduced to resolve cross-layer superposition and track features distributed across layers, we adapt this approach to analyze activations from different training checkpoints. By applying cross-snapshot crosscoders, we can observe where features emerge, rotate, and degenerate, thereby providing deeper insight into the internal dynamics of model pre-training. Our main contributions are as follows:

\begin{enumerate}[leftmargin=*]
    \item To the best of our knowledge, this work is the first to adapt crosscoders to study training dynamics (Section~\ref{sec:crosscoders}). We evaluate completeness and faithfulness of interpretations provided by crosscoders in Appendix~\ref{appendix:crosscoder-detail}.
    \item We perform in-depth analyses on cross-snapshot features in Section~\ref{sec:assessing}. We empirically show that the decoder norms can serve as proxies of feature evolution status, and showcase both general and per-feature evolutionary properties.
    \item Our method successfully connects the microscopic features to the macroscopic downstream task metrics using attribution-based circuit tracing techniques (Section~\ref{sec:micro-to-macro}).
    \item We show evidence for the phase transition from a statistical learning phase to a feature learning phase in Section~\ref{sec:phase-transition}.
\end{enumerate}

\section{Related Works}

\paragraph{Learning dynamics and phase transitions.} 
Multiple theoretical frameworks explain neural network training dynamics. Neural Tangent Kernel (NTK) theory~\citep{jacot2018ntk} establishes that infinite-width networks evolve as kernel machines with fixed kernels during gradient descent, with extensions to finite-width corrections~\citep{dyer2020asymptotics} and modern architectures~\citep{yang2020tensorprograms3, yang2021tensorprograms2b}. Recent work identifies phase transitions and lazy regimes in training dynamics~\citep{kumar2024grokkingntk, zhou2025twophase}. Information Bottleneck (IB) theory~\citep{tishby2015informationbottleneck} formulates deep learning as optimizing compression-prediction tradeoffs. \citet{shwartz2017information} empirically demonstrates two distinct training phases—initial fitting followed by compression—which aligns with our findings in Section~\ref{sec:phase-transition}. Singular Learning Theory (SLT) treats neural networks as singular statistical models. \citet{watanabe1999slt, watanabe2009slt} introduces the Real Log Canonical Threshold to provide geometric complexity measures predicting phase transitions, with recent advances enabling practical estimation at scale through Local Learning Coefficients~\citep{lau2024llc, furman2024llc, wang2025rllc}. Our work complements these theoretical frameworks by providing detailed mechanistic accounts of feature evolution during transformer pre-training, bridging high-level theory with empirical observations.

\paragraph{Sparse dictionary learning.}
The superposition hypothesis posits that models use linear representations~\citep{bengio2014representationlearning, alain2017linearprobe, vargas2020linear} to embed more features than neurons~\citep{olah2020zoom, elhage2022superposition}. Sparse Autoencoders (SAEs)~\citep{bricken2023monosemanticity, hubun2024sae,he2024llamascope} extract monosemantic features from superposition using sparse dictionary learning, with subsequent improvements and scaling efforts~\citep{gao2025scalingsae, rajamanoharan2024jumprelu}. \citet{ge2024automatically} and \citet{dunefsky2024transcoders} propose transcoders, an SAE variant that predicts future activations to improve circuit tracability. Recently, \citet{lindsey2024crosscoder} introduces crosscoders to simultaneously read and write to multiple layers. \citet{mishra2024crosscoder, minder2025crosscoders} leverages crosscoders for capturing the difference between pretrained models and their chat-tuned versions.

\section{Tracking the Evolution of Features}
\label{sec:crosscoders}

\begin{figure}[h!]
    \centering
    \includegraphics[width=0.9\textwidth]{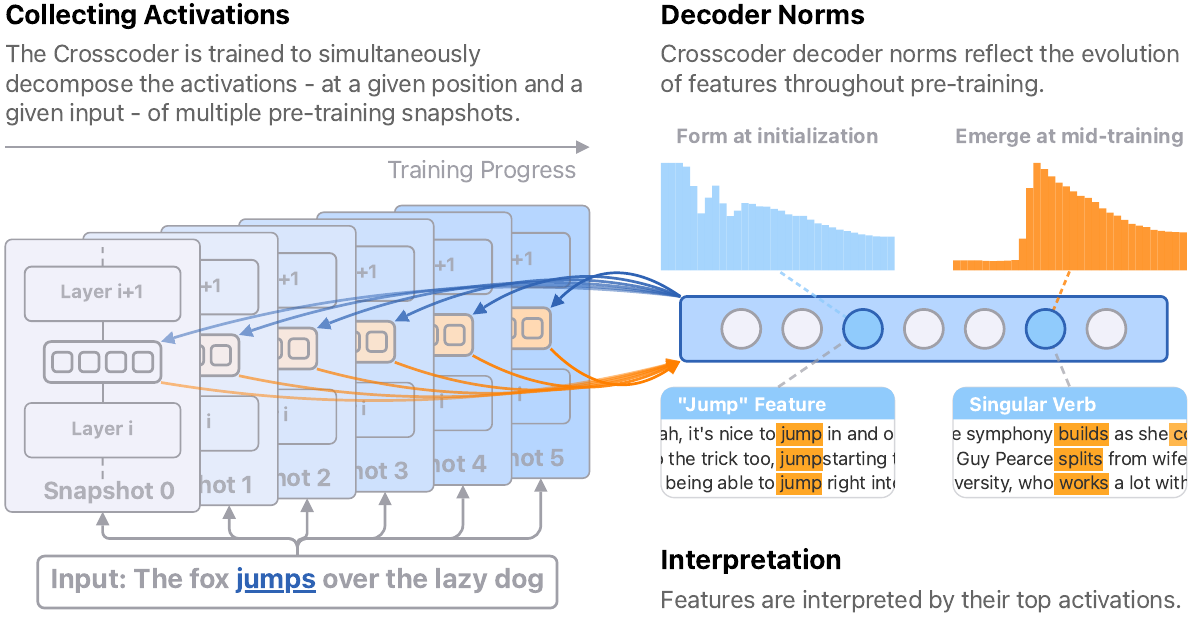}
    \caption{Overview of our method. The crosscoder is trained to decompose activations of multiple pre-training snapshots (left) into sparse features (right).}
    \label{fig:crosscoder-overview}
\end{figure}

\paragraph{Crosscoder architecture.}
For a given text corpus $\mathcal{C}$ and a family of training snapshots $\Theta$ saved during LLM pre-training, let $\mathcal{A}=\{a^\theta:\mathcal{C}\times\mathbb{N}\to\mathbb{R}^{d_\text{model}} \mid \theta\in\Theta\}$ denote a corresponding family of parameterized functions mapping input datapoints to model activations at a specific layer. Each datapoint $x=(c,j)\in \mathcal{C}\times\mathbb{N}$ is a training token in $\mathcal{C}$, indexing the $j$-th token in sequence $c$.

A cross-snapshot crosscoder~\citep{lindsey2024crosscoder} operates on the parameterized function family $\mathcal{A}$ over corpus $\mathcal{C}$ (Figure~\ref{fig:crosscoder-overview}). The crosscoder architecture is defined by:

\begin{equation}
    \begin{aligned}
        f(x)&=\sigma\left(\sum_{\theta \in \Theta} W_\text{enc}^\theta a^\theta(x)+b_\text{enc}\right)\\
        \hat{a}^\theta(x)&=W_\text{dec}^\theta f(x) + b_\text{dec}^\theta
    \end{aligned}
\end{equation}

with $W_\text{enc}^\theta \in \mathbb{R}^{n_\text{features} \times d_\text{model}}$, $b_\text{enc} \in \mathbb{R}^{n_\text{features}}$, $W_\text{dec}^\theta \in \mathbb{R}^{d_\text{model} \times n_\text{features}}$, and $b_\text{dec}^\theta \in \mathbb{R}^{d_\text{model}}$. The parameters $W_\text{enc}^\theta$, $W_\text{dec}^\theta$, and $b_\text{dec}^\theta$ correspond to snapshot-specific encoder and decoder weights for parameter $\theta$. The activation function $\sigma(\cdot)$ produces sparse feature activations $f(x)$ shared for all snapshots, and $\hat{a}^\theta(x)$ denotes the reconstructed term of model activation $a^\theta(x)$.

\paragraph{Training objectives.}
The crosscoder is trained to minimize the loss:

\begin{equation}
    \mathcal{L}(x) = \underbrace{\sum_{\theta\in\Theta}||a^\theta(x) - \hat{a}^\theta(x)||^2}_{\text{Reconstruction loss}} + \underbrace{\lambda_\text{sparsity}\sum_{\theta\in\Theta}\sum_{i=1}^{n_\text{features}}\Omega \left( f_i(x) \cdot || W_{\text{dec}, i}^\theta || \right)}_{\text{Sparsity loss}}
\end{equation}

where $\lambda_\text{sparsity}$ is a hyperparameter to control the trade-off between reconstruction fidelity and feature sparsity. The regularization function $\Omega ( \cdot )$ serves as a differentiable substitute for L0 regularization, penalizing non-sparse feature activations. We include the decoder norm $|| W_{\text{dec}, i}^\theta ||$ in the regularization term to prevent the crosscoder from trivially reducing the feature activation $f_i(x)$ while inflating the decoder norms under imperfect L0 approximations such as L1 regularization. Appendix~\ref{appendix:crosscoder-activation-regularization-selection} shows the details for selecting the proper activation function $\sigma(\cdot)$ and regularization function $\Omega(\cdot)$ to optimize crosscoder feature sparsity.

\paragraph{Experimental setup.}
We use Pythia-160M and Pythia-6.9B~\citep{biderman2023pythia} for our experiments throughout this work. Pythia is a Transformer language model suite with well-controlled training settings and accessibility to training snapshots. We select the middle layers (Layer 6 of Pythia-160M and Layer 16 of Pythia-6.9B) for training crosscoders. To balance training cost with the granularity of feature evolution analysis, we select 32 snapshots out of 154 open-source snapshots ranging from step 0 to 143,000, with a stratified sampling approach: (1) all 20 snapshots before step 10,000 to capture early feature evolution with maximum temporal resolution, and (2) 12 evenly spaced snapshots from later training stages, containing 4 snapshots from steps 14,000–34,000 and 8 snapshots from steps 47,000–143,000. We use SlimPajama~\citep{shen2023slimpajama}, a comprehensive text dataset covering a variety of data sources to sample activations.

\begin{wrapfigure}[12]{r}[0pt]{0.4\textwidth}
  \vspace{-12pt}
  \centering
  \includegraphics[width=0.35\textwidth]{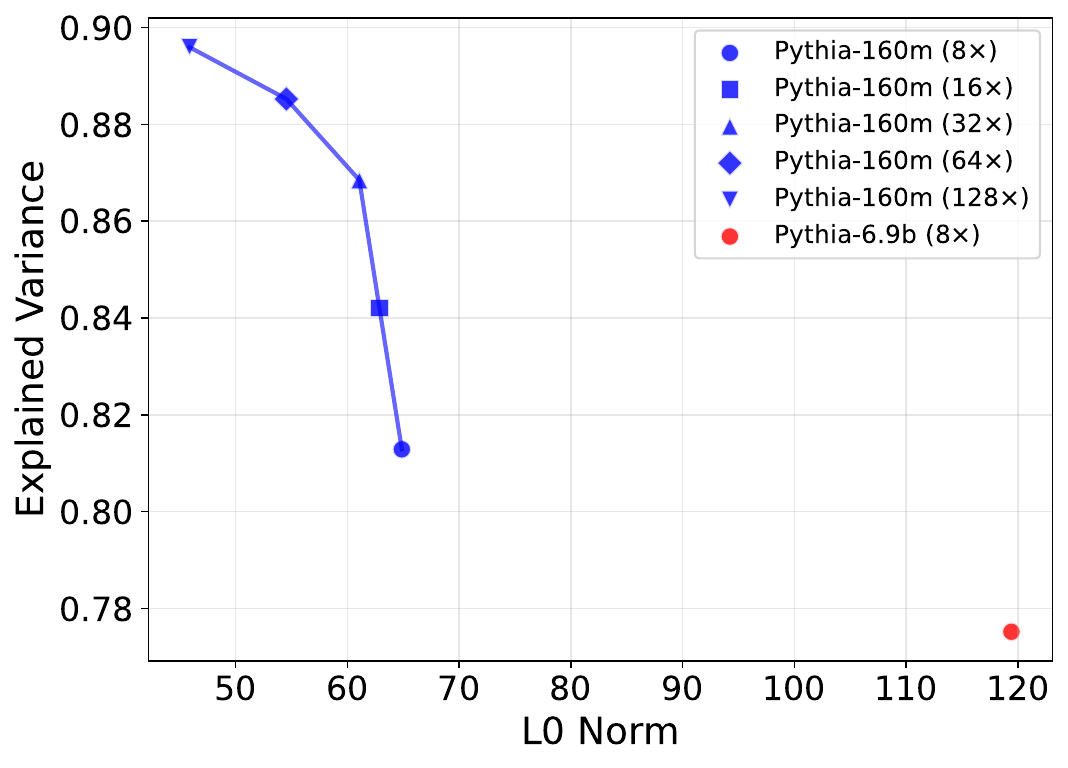}
  \caption{Explained variances versus L0 norms of our crosscoders.}
  \label{fig:ev-l0-across-size-exp}
\end{wrapfigure}

\paragraph{Results.}
Figure~\ref{fig:ev-l0-across-size-exp} demonstrates crosscoder performance in decomposing model activations into sparse feature representations. We evaluate reconstruction quality (measured by activation variance explained) and sparsity (measured by L0 norm averaged across snapshots). Increasing dictionary size $n_\text{features}$ yields significant Pareto improvements across both metrics.

To examine whether crosscoders can extract cross-snapshot features and align identical features in consistent directions within the feature space, we compare crosscoder performance against corresponding SAEs trained individually on each Pythia snapshot. We train SAEs using identical settings and hyperparameters as crosscoders on each snapshot. Figures~\ref{fig:crosscoder-ev-comparison} and~\ref{fig:crosscoder-l0-comparison} demonstrate that crosscoders achieve comparable performance in L0 norm and explained variance at each snapshot, even when SAEs are optimized for individual snapshots where they should have a theoretical advantage.

We further compare the Pareto frontiers (explained variance versus L0 norm) between crosscoders and SAEs trained on the final snapshot. Figure~\ref{fig:crosscoder-pareto-comparison} shows that crosscoders exhibit a slightly superior Pareto frontier, demonstrating their effectiveness in sparse dictionary learning beyond their primary capability of tracking feature evolution across snapshots.

\begin{figure*}[t]
    \centering
    \begin{subfigure}[b]{0.32\textwidth}
        \centering
        \includegraphics[width=\textwidth]{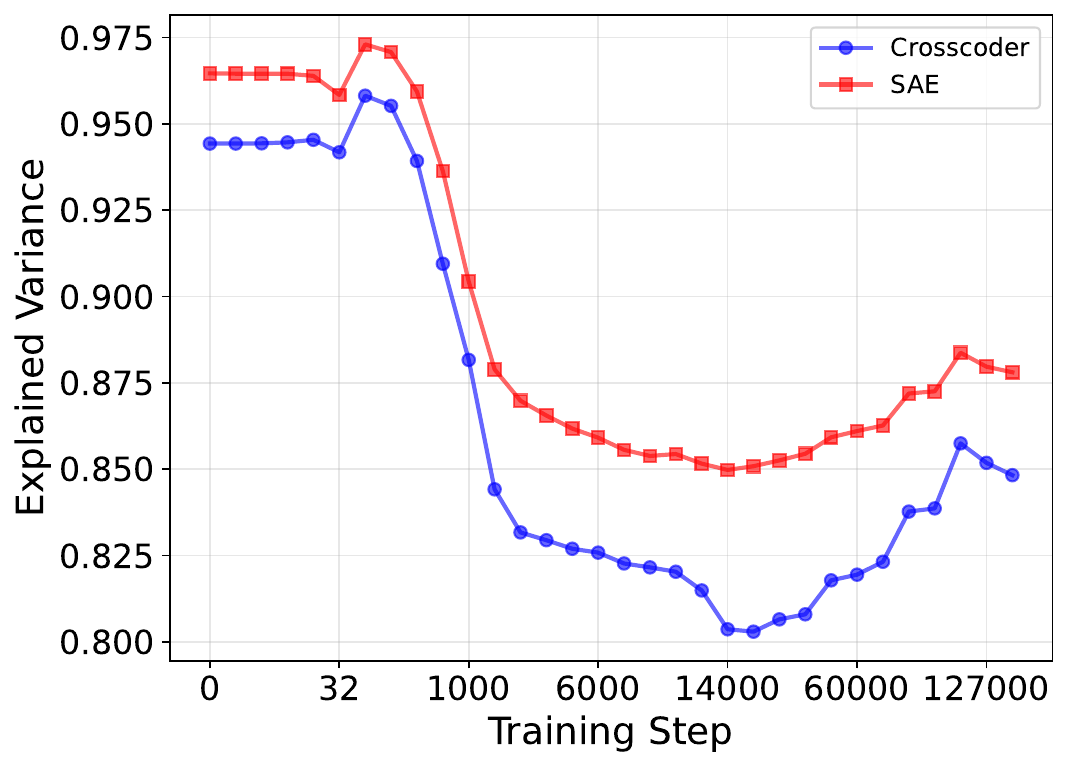}
        \caption{}
        \label{fig:crosscoder-ev-comparison}
    \end{subfigure}
    \hfill
    \begin{subfigure}[b]{0.32\textwidth}
        \centering
        \includegraphics[width=\textwidth]{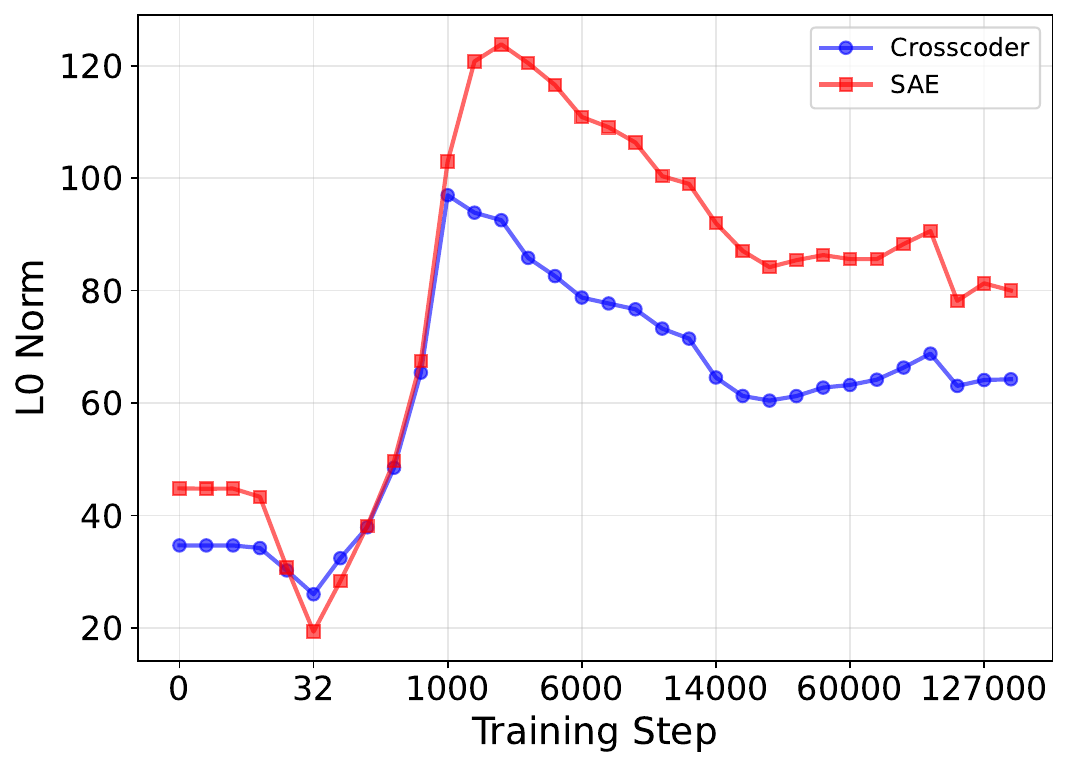}
        \caption{}
        \label{fig:crosscoder-l0-comparison}
    \end{subfigure}
    \hfill
    \begin{subfigure}[b]{0.32\textwidth}
        \centering
        \includegraphics[width=\textwidth]{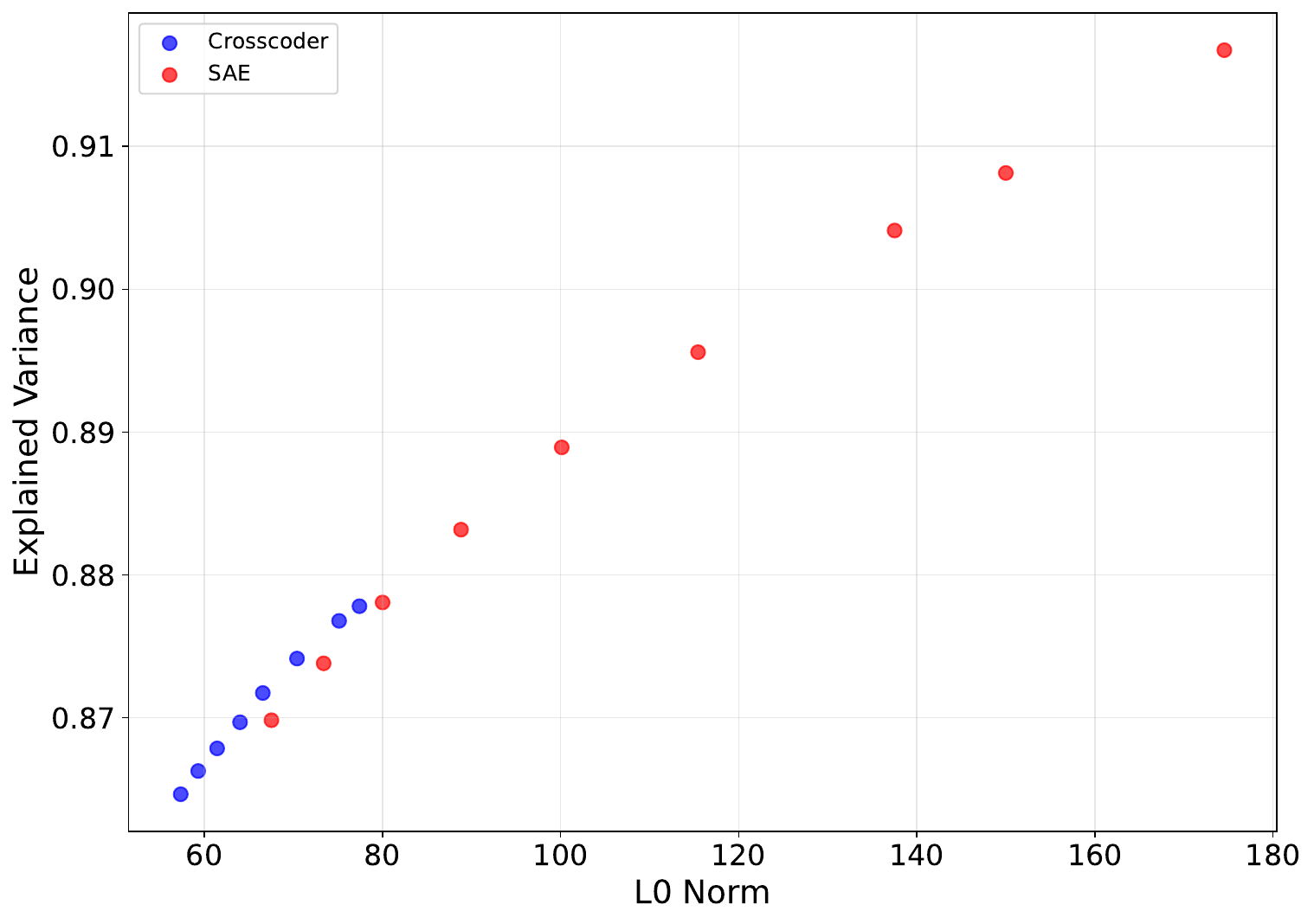}
        \caption{}
        \label{fig:crosscoder-pareto-comparison}
    \end{subfigure}
    \caption{Comparison between crosscoders and per-snapshot SAEs. (a) The explained variance of crosscoders versus SAEs at each snapshot. (b) The L0 norm of crosscoders versus SAEs at each snapshot. (c) The Pareto frontier comparison of crosscoders and SAEs trained on the final snapshot.}
\end{figure*}

\paragraph{Feature evolution revealed by crosscoders.}
An important advantage of crosscoders is the unified feature space (or \textit{sparse codes}) they reveal. The crosscoder encoder aggregates cross-snapshot information to produce shared feature activations. Then these activations are translated back to recover the original activations by a group of independent decoders (or \textit{dictionaries}).

If a feature activates but ``exists" at only a subset of snapshots, the sparse penalty will suppress the decoder norms of this feature at irrelevant snapshots to near-zero so they won't interfere with reconstruction on these snapshots and also reduce sparsity loss.

This design principle leads to a crucial observation: the decoder norm $\|W_{\text{dec}, i}^\theta\|$ directly reflects the strength and presence of feature $i$ at snapshot $\theta$. Therefore, tracking decoder norm changes across snapshots provides a direct window into feature evolution dynamics. Appendix~\ref{appendix:decoder-norm} provides a sanity check of whether $\|W_{\text{dec}, i}^\theta\|$ can indeed serve as a proxy of feature strength using linear probes.

\paragraph{Potential failure modes of crosscoder feature alignment.}
Before assessing crosscoder features, we further discuss two theoretical concerns about cross-snapshot feature alignment:

\textit{Will crosscoders misalign unrelated features?} Such misalignment would strongly contradict the optimization objective of crosscoders. The distinct activation patterns of unrelated features would conflict when forced into the same dimension. Once a feature activates, it would cause others to activate simultaneously, introducing noise in the reconstruction. In terms of results, misalignment would lead to the polysemanticity of crosscoder features, damaging the consistency of their interpretation, which our interpretability evaluation demonstrates does not occur.

\textit{Will crosscoders split shared features?} Suppose features from different snapshots represent the same underlying concept with identical activation patterns, splitting them across multiple dimensions would be suboptimal: it wastes representational capacity in feature space, as both previous works~\citep{gao2025scalingsae, templeton2024scalingmonosemanticity} and Appendix~\ref{appendix:experiments} prove that feature space dimensionality is crucial for better reconstruction fidelity. Nevertheless, in rare cases, we do observe feature splitting among highly active features. These features share semantic and directional similarity but exhibit subtly different activation patterns that justify separate dimensions. We detail this phenomenon in Appendix~\ref{appendix:feature-splitting}.

These considerations suggest that crosscoders should naturally achieve effective feature alignment, mapping semantically equivalent features from different snapshots to consistent positions in the unified feature space.

\section{Assessing Cross-Snapshot Features}
\label{sec:assessing}

\subsection{Overview of Feature Evolution}

\begin{figure*}[h!]
   \centering
   \includegraphics[width=\textwidth]{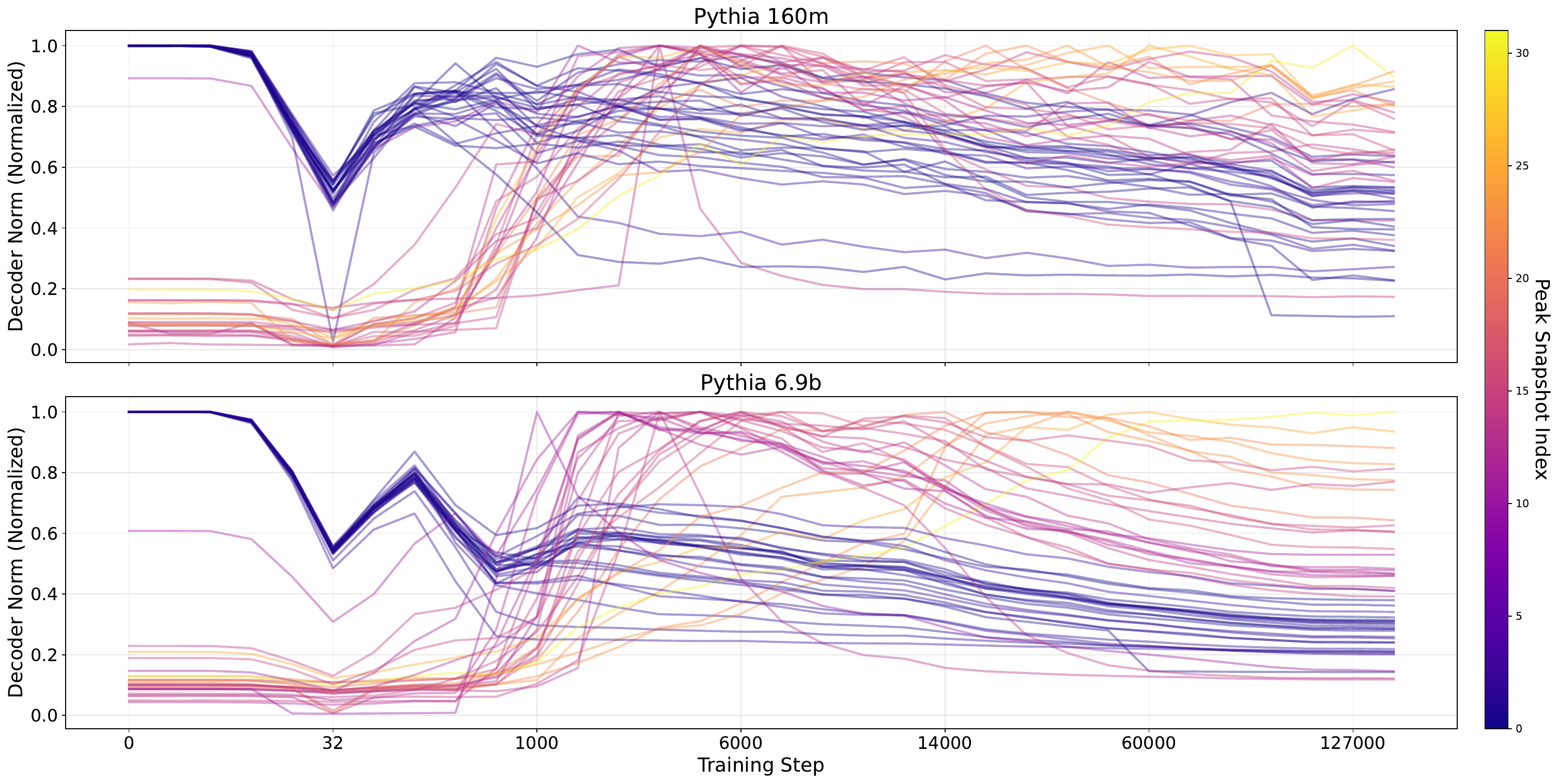}
   \caption{Overview of cross-snapshot feature decoder norm evolution. Features are extracted by a 98,304-feature crosscoder on Pythia-160M (top) and a 32,768-feature crosscoder on Pythia-6.9B (bottom).}
   \label{fig:norm-evolution}
\end{figure*}


Figure~\ref{fig:norm-evolution} shows 50 randomly sampled features and their decoder norm evolution across snapshots. Each feature's decoder norms are linearly rescaled to a maximum of $1$. Most cross-snapshot features exhibit two distinct developmental patterns:

\begin{enumerate}[leftmargin=*]
    \item \textit{Initialization features} that exist from random initialization, exhibit a sudden drop and recovery around step $128$, then gradually decay. The existence of these features has been established by~\citet{bricken2023monosemanticity} and~\citet{heap2025saerandom}.
    \item \textit{Emergent features} that begin forming primarily around step $1000$, reaching peak intensity at various subsequent training steps. There also exists emergent features only appearing in late training, which we discuss in later sections.
\end{enumerate}

\begin{figure*}[h!]
   \centering
   \begin{subfigure}[b]{0.32\textwidth}
       \centering
       \includegraphics[width=\textwidth]{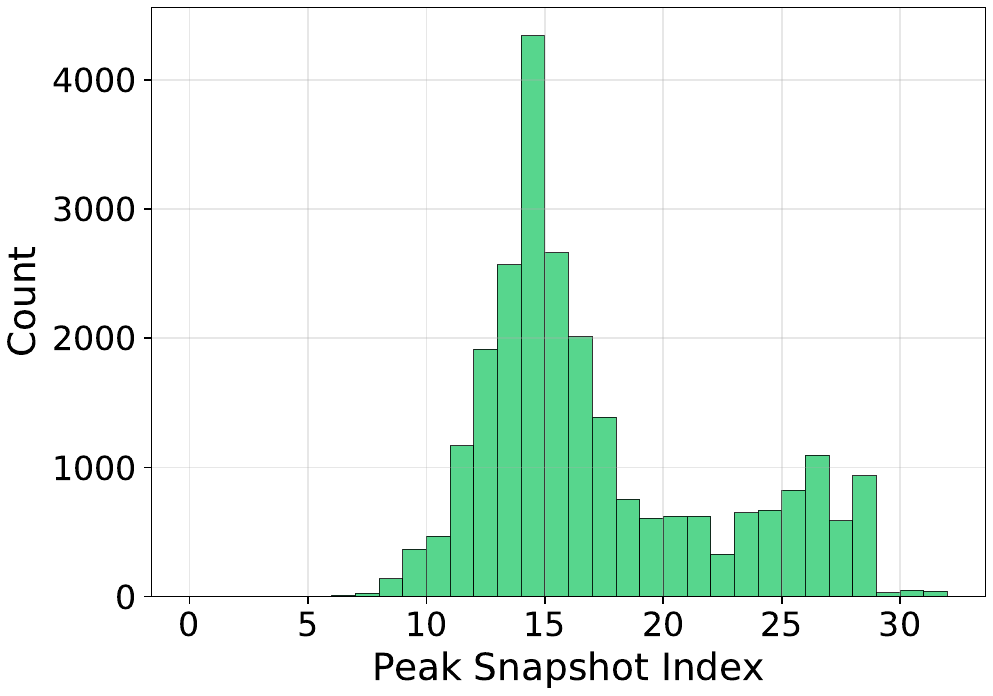}
       \caption{}
       \label{fig:peak-norm-distribution}
   \end{subfigure}
   \hfill
   \begin{subfigure}[b]{0.32\textwidth}
       \centering
       \includegraphics[width=\textwidth]{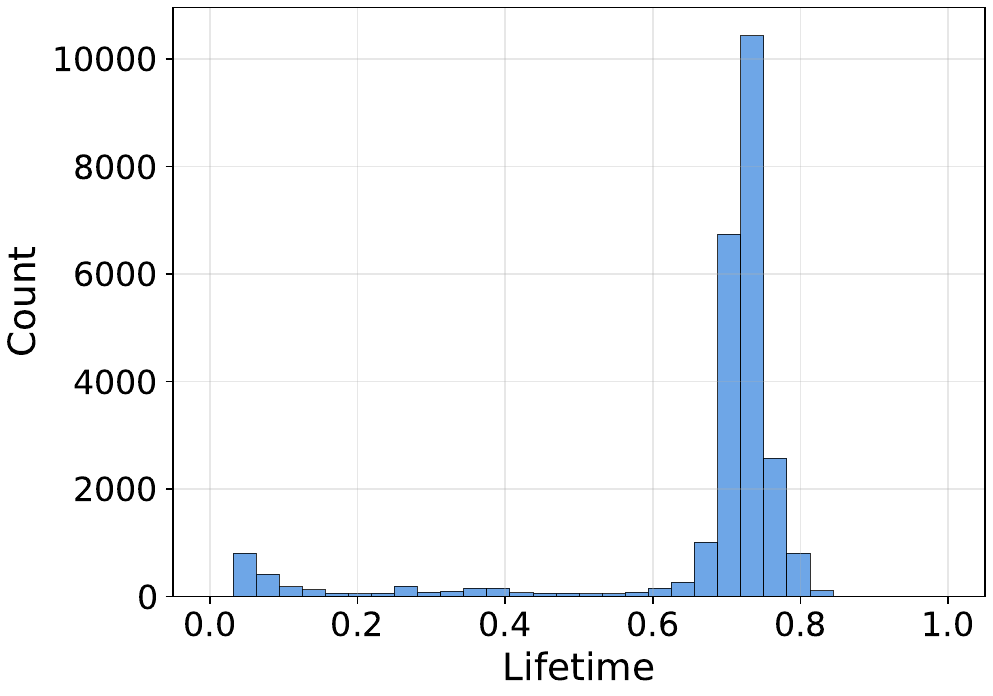}
       \caption{}
       \label{fig:lifetime-distribution}
   \end{subfigure}
   \hfill
   \begin{subfigure}[b]{0.32\textwidth}
       \centering
       \includegraphics[width=\textwidth]{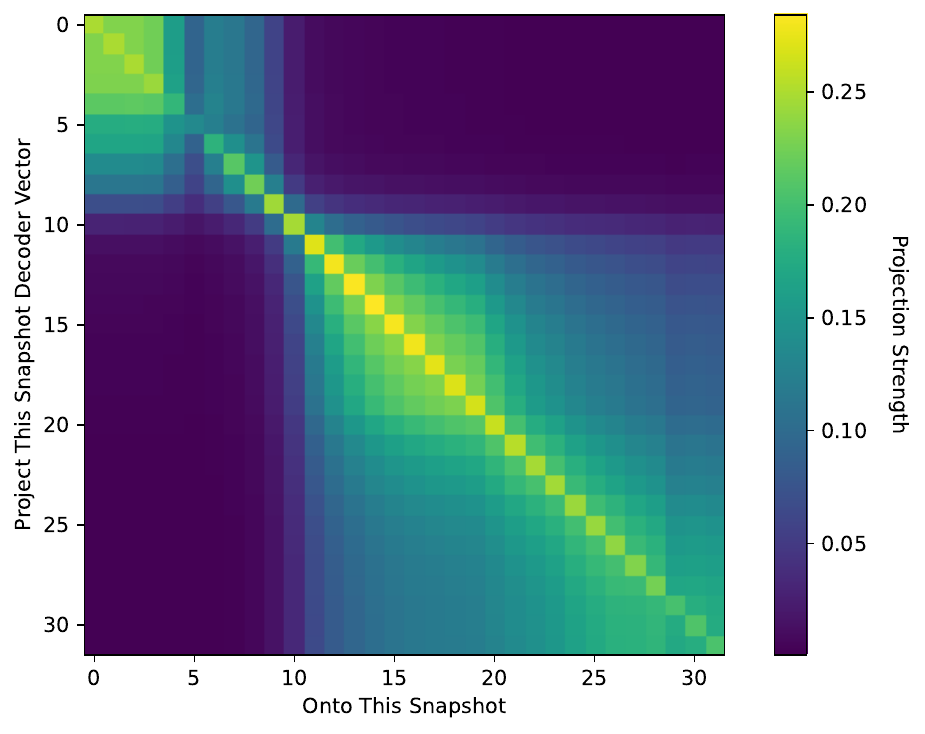}
       \caption{}
       \label{fig:mean-projection}
   \end{subfigure}
   
   \caption{Statistics of emergent features in a 98,304-feature crosscoder on Pythia-160M. (a) Distribution of peak emergence times. (b) Distribution of feature lifetime. (c) Mean projection of each feature's decoder vector of snapshot $\theta_i$ onto its decoder vector of snapshot $\theta_j$.} 
\end{figure*}

\paragraph{Feature emergence steepness.} Previous studies suggest that loss curves comprise discrete phase changes, each contributing to different circuits at distinct training stages, with evidence from toy model features~\citep{elhage2022superposition}, 5-digit addition~\citep{nanda2023grokking}, and in-context learning~\citep{olsson2022induction}. From a feature evolution perspective, we investigate whether features emerge abruptly enough to constitute the fundamental units of these phase transitions. By evaluating the time spent by each feature from initial emergence to peakness, we find that emergence steepness exhibits considerable variation, demonstrating the coexistence of both gradually-formed and abruptly-appearing features.

\paragraph{Feature persistence after emergence.} We next examine whether emergent features persist after formation. We define the lifetime of a crosscoder feature as $|\{j\mid\|W_{\text{dec}, i}^{\theta_j}\|>0.3\}|$, where a threshold is introduced to block out near-zero decoder norms. Figure~\ref{fig:lifetime-distribution} shows the lifetimes of all emergent features. We find that most features persist for extended periods (above 60\% of snapshots) after formation, indicating that: (1) LLMs retain learned features robustly, and (2) our crosscoders successfully align and track these features across snapshots.

\paragraph{A universal directional turning point.} To further investigate the geometry of feature evolution, we compute the projections between feature directions across snapshots. The directional evolution patterns prove remarkably consistent: most features undergo drastic directional shifts around step 1,000, rendering pre- and post-step 1,000 directions nearly orthogonal (Figure~\ref{fig:mean-projection}). Subsequently, features continue to rotate more gradually, with directions at the final snapshot maintaining notable cosine similarities to their initial post-step 1,000 orientations—a significant finding given the high-dimensional activation space.

\subsection{Correlation between Emergence Step and Complexity}
\label{sec:complexity}

To gain deeper insights into features emerging at different training stages, we leverage LLMs to automatically assess their complexity based on top activation patterns. 

\begin{figure*}[h!]
   \centering
   \begin{subfigure}[b]{0.32\textwidth}
       \centering
       \includegraphics[width=\textwidth]{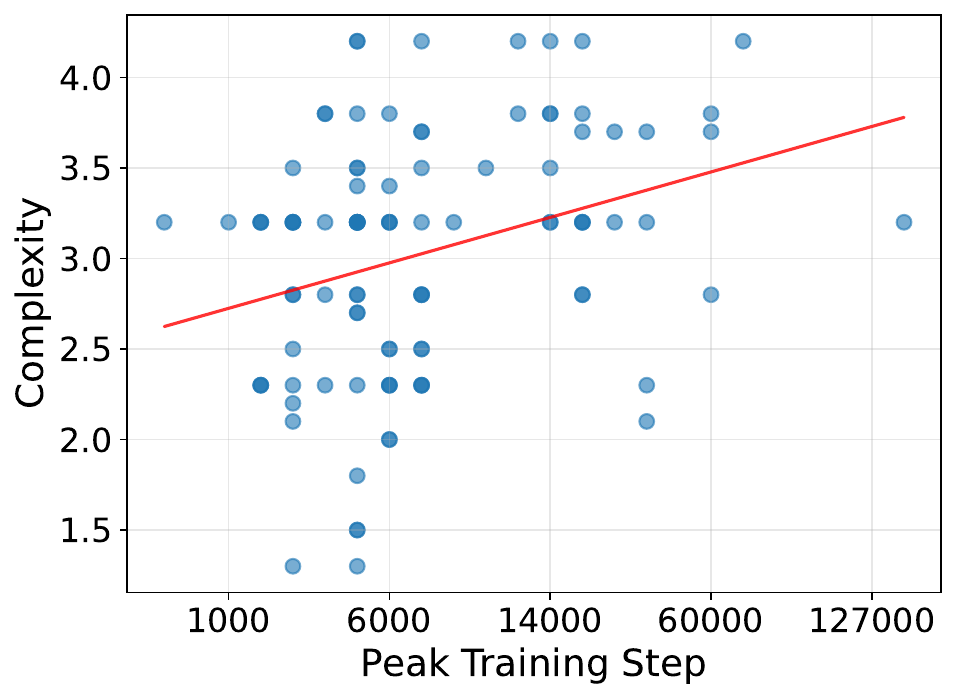}
       \caption{}
       \label{fig:complexity}
   \end{subfigure}
   \hfill
   \begin{subfigure}[b]{0.32\textwidth}
       \centering
       \includegraphics[width=\textwidth]{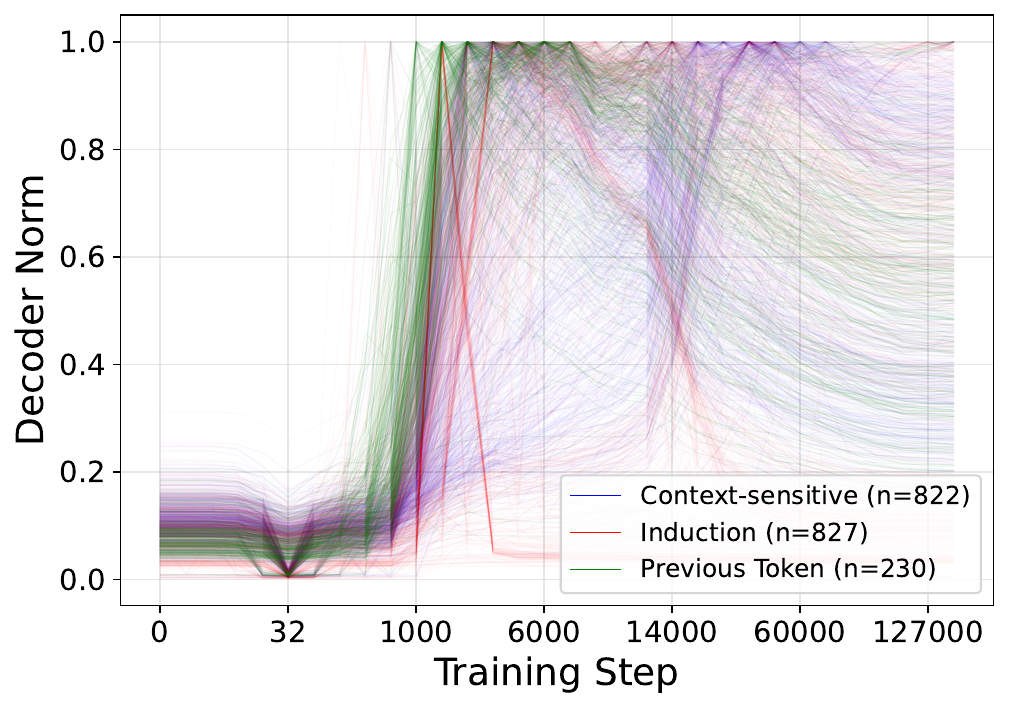}
       \caption{}
       \label{fig:comparison-evolution}
   \end{subfigure}
   \begin{subfigure}[b]{0.32\textwidth}
   \hfill
       \centering
       \includegraphics[width=\textwidth]{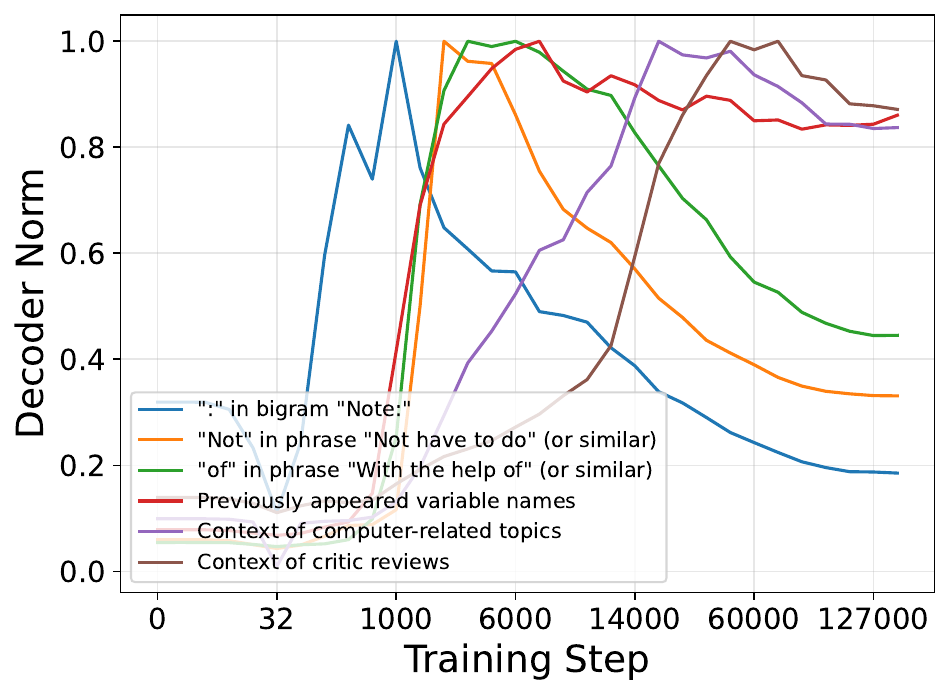}
       \caption{}
       \label{fig:case-norm-evolution}
   \end{subfigure}
   \caption{(a) Complexity scores (evaluated by Claude Sonnet 4) versus peak emergence times, showing a moderate positive correlation (Pearson r = 0.309, p = 0.002). (b) Decoder norm evolution trajectories for all previous token, induction, and context-sensitive features across training. (c) More cases of feature decoder norm evolution trajectories.}
\end{figure*}

We follow~\citet{cunningham2024complexity} to score feature complexity for 100 randomly sampled emergent features based on their top activations, ranging from 1 (simple) to 5 (complex). The result in Figure~\ref{fig:complexity} reveals a non-trivial correlation with moderate strength between peak timing and complexity scores, suggesting that more complex features tend to emerge later in training. The scoring rubrics and complete prompt for automated complexity scoring can be found in Appendix~\ref{appendix:complexity}.

\subsection{Case Study on Typical Cross-Snapshot Features}

We employ simple rule-based approaches to find several well-studied feature types in a crosscoder with 32,768 features trained on Pythia-6.9B, including previous token features (which activate based on the preceding token), induction features (which activate on the second [A] in patterns [A][B]...[A][B] and help predict [B]), and context-sensitive features. These feature categories are extensively documented in prior research on neurons and SAE features~\citep{gurnee2024universalneuron, hubun2024sae, ge2024automatically}. 

Figure~\ref{fig:comparison-evolution} demonstrates the distinct temporal pattern in feature emergence: previous token features arise early (around steps 1,000–5,000), while induction and context-sensitive features appear later and over a wider range (mostly steps 10,000–100,000). This suggests a general emergence hierarchy, from previous token to induction to context-sensitive features, which aligns with their increasing complexity. This finding is also consistent with the causal dependency between induction heads and previous token heads~\citep{olsson2022induction}.

For the majority of features that cannot fit into specific rule-based patterns, we further demonstrate additional cases of random emergent features with different evolutionary patterns (Figure~\ref{fig:case-norm-evolution}). Labels are annotated by summarizing the top activation samples.

\section{Connecting Microscopic Evolution to Macroscopic Behaviors}
\label{sec:micro-to-macro}

One of the ambitious missions of mechanistic interpretability research is to connect feature-level findings with the model's downstream performance. To this end, we employ attribution-based circuit tracing techniques~\citep{syed2023attributionpatching, ge2024automatically, marks2025sparsefeaturecircuits} to investigate the causal effects of crosscoder features formation on downstream task metrics improvements.

\paragraph{Method.} Let metric $m: \mathbb{R}^{d_\text{model}} \to \mathbb{R}$ be an arbitrary scalar-valued function of model activations $a^\theta(x)$.\footnote{The original notation in~\citet{marks2025sparsefeaturecircuits} uses $m$ as a function of input $x$, where the computation graph flows through nodes of interest. Since we focus on feature effects within a single layer (e.g., layer 6 of Pythia-160M), we simplify $m$ as a function of model activations to avoid complex intervention notation.} To quantify the causal effect of each feature activation $f_i(x)$ on metric $m$, we first decompose model activations into per-feature representations:

\begin{equation}
    \begin{aligned}
        a^\theta(x) &= \hat{a}^\theta(x) + \epsilon^\theta(x)\ = \sum_{i=1}^{n_{\text{features}}} f_i(x) \cdot W_{\text{dec},i}^{\theta} + b_{\text{dec}}^\theta + \epsilon^\theta(x)
    \end{aligned}
    \label{eq:decomposition}
\end{equation}

where $\epsilon^\theta(x)\in\mathbb{R}^{d_\text{model}}$ represents the crosscoder reconstruction error. This decomposition incorporates crosscoder features into the computation graph, enabling gradient computation with respect to features. We then estimate each feature's causal effect using its attribution score:

\begin{equation}
    \operatorname{attr}_i^\theta(x) = f_i(x) \cdot \frac{\partial m(a^\theta(x))}{\partial f_i(x)}
    \label{eq:attribution}
\end{equation}

where the gradient $\frac{\partial m}{\partial f_i}$ flows through the decomposition in Eq.~\ref{eq:decomposition}. This attribution score employs first-order Taylor expansion as a linear approximation of model computation.

For structured downstream tasks with clean/corrupted input pairs, such as subject-verb agreement~\citep{matthew2021sva}, we can employ the full framework of attribution patching~\citep{syed2023attributionpatching, marks2025sparsefeaturecircuits} by emphasizing differences between clean and corrupted inputs:

\begin{equation}
    \operatorname{attr}_i^\theta(x, \tilde{x}) = \left[f_i(x) - f_i(\tilde{x})\right] \cdot \frac{\partial m(a^\theta(x))}{\partial f_i(x)}
    \label{eq:attribution-patching}
\end{equation}

where $\tilde{x}$ is the corrupted version of input $x$. Attribution patching isolates components explaining the transition from corrupted to clean performance, excluding the majority of components that contribute to baseline model function. This focused approach improves the validity of the linear approximation.

In practice, we employ the integrated gradient (IG) version of the attribution scores defined in Eq.~\ref{eq:attribution} and Eq.~\ref{eq:attribution-patching}, which compute gradients at evenly-spaced points from $x$ to $\tilde{x}$ ($x$ to zero vector in the pure attribution case). Details of IG computation can be found in Appendix~\ref{appendix:downstream-tasks}.

\begin{figure*}[h!]
   \centering
   \begin{subfigure}[b]{0.32\textwidth}
       \centering
       \includegraphics[width=\textwidth]{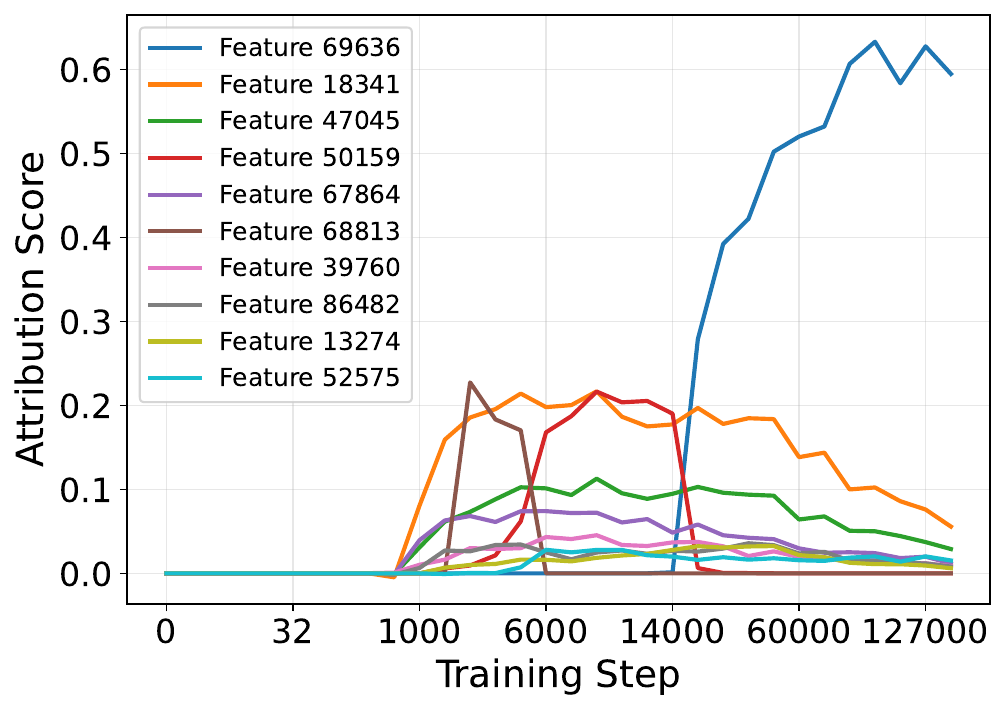}
       \caption{}
       \label{fig:attribution-scores}
   \end{subfigure}
   \hfill
   \begin{subfigure}[b]{0.32\textwidth}
       \centering
       \includegraphics[width=\textwidth]{images/metric-recovery-ablate-sva-nounpp.pdf}
       \caption{}
       \label{fig:metric-recovery-ablate}
   \end{subfigure}
   \hfill
   \begin{subfigure}[b]{0.32\textwidth}
       \centering
       \includegraphics[width=\textwidth]{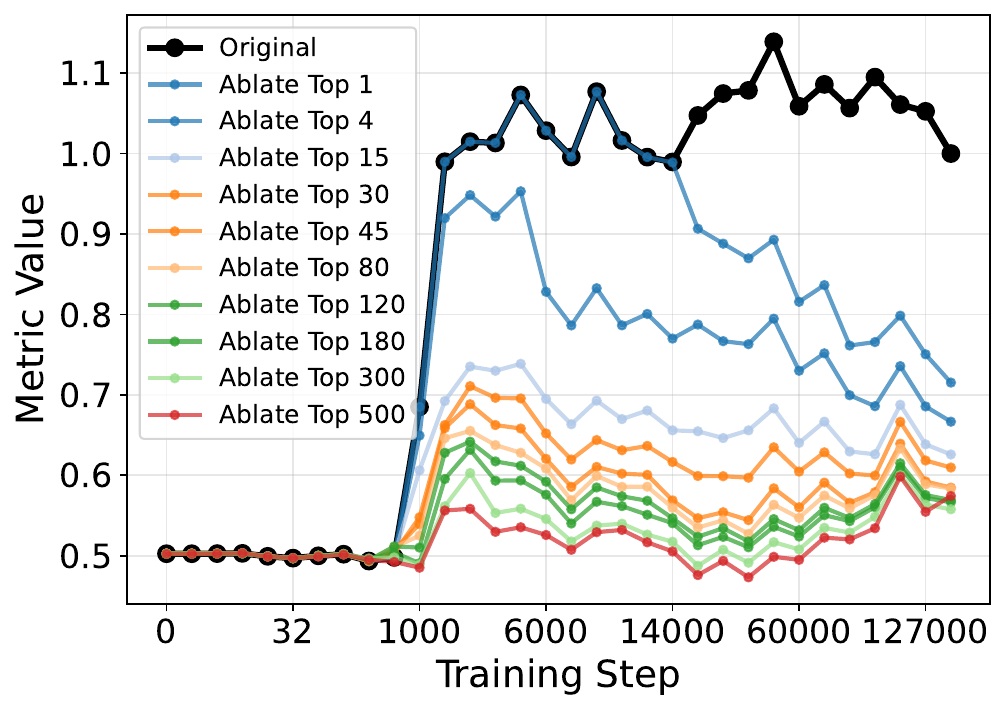}
       \caption{}
       \label{fig:metric-recovery-only}
   \end{subfigure}

   \vspace{1em}
   
   \begin{subfigure}[b]{0.96\textwidth}
       \centering
       \includegraphics[width=\textwidth]{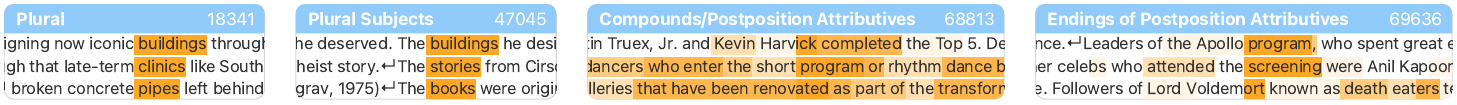}
       \caption{}
       \label{fig:subfigure-d}
   \end{subfigure}
   \caption{Crosscoder feature attribution on the Across-PP variant of the subject-verb agreement task, e.g. ``The teachers near the desk are". We use a crosscoder train at layer 6 of Pythia-160M with 98,304 features. (a) The attribution scores of top contributing features over time. (b) The metric recovery when ablating all features except the top $k$ contributing features. (c) The metric recovery when ablating the top $k$ contributing features. (d) Top activation samples of key features contributing to this task. Features recognizing plural nouns appear before features recognizing postposition attributives.}
   \label{fig:attribution}
\end{figure*}

\paragraph{Experimental setup.} We evaluate our method on subject-verb agreement (SVA)~\citep{matthew2021sva}, induction, and indirect object identification (IOI)~\citep{wang2023ioi} tasks, using 1000 samples each to identify critical features. 

For SVA and IOI, we create corrupted controls by swapping singular/plural forms (SVA) or altering the second subject (IOI), using logit differences between clean and corrupted answers as metrics. For induction, which lacks natural corruptions, we use target answer log probabilities. We compute IG attribution scores for all features on a 98,304-feature crosscoder trained on Pythia-160M, and rank crosscoder feature contribution by mean attribution scores across all snapshots. We then perform complementary ablations: (1) removing top-ranked features, and (2) removing all except top-ranked features.

\paragraph{Results.} Figure~\ref{fig:attribution} shows results for the Across-PP variant of SVA, where postpositional attributives separate subjects and verbs, e.g. ``The teachers near the desk are". We identify key contributing features ordered by emergence time: (1) Features 18341 and 47045 capture plural nouns, with 47045 specialized for plural subjects; (2) Feature 68813 marks compound subjects and postpositional attributives; (3) Features 50159 and 69636 identify endings of postpositional attributives, with 69636 showing higher accuracy. Additional features include subject-specific and context-specialized plural markers with lower task contributions. Notably, Features 68813, 50159, and 69636 alternately dominate the metric, revealing circuit-level model evolution through component iteration.

Ablation experiments across all tasks demonstrate that within tens of features, we can consistently disrupt or recover model performance on downstream tasks across training snapshots, confirming our method identifies necessary and sufficient task components.

\section{Observations of A Statistical-to-Superposition Transition}
\label{sec:phase-transition}

\begin{figure*}[h!]
   \centering
   \begin{subfigure}[b]{0.32\textwidth}
       \centering
       \includegraphics[width=\textwidth]{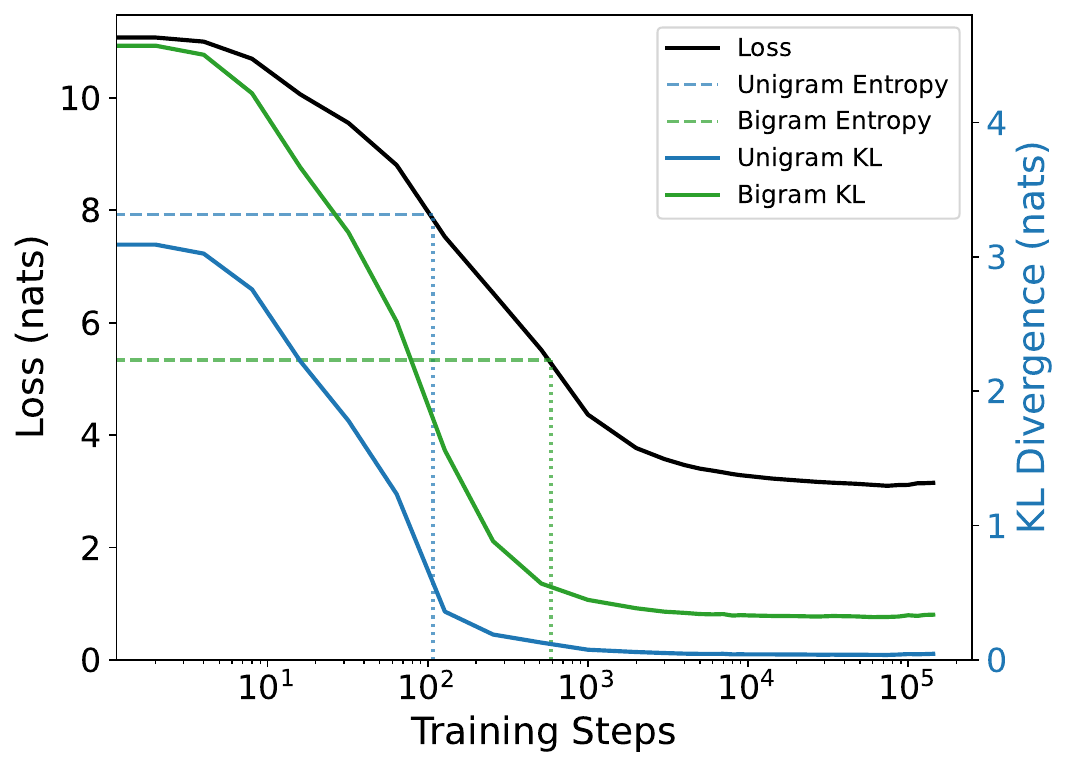}
       \caption{}
       \label{fig:zipfs-160m}
   \end{subfigure}
   \hfill
   \begin{subfigure}[b]{0.32\textwidth}
       \centering
       \includegraphics[width=\textwidth]{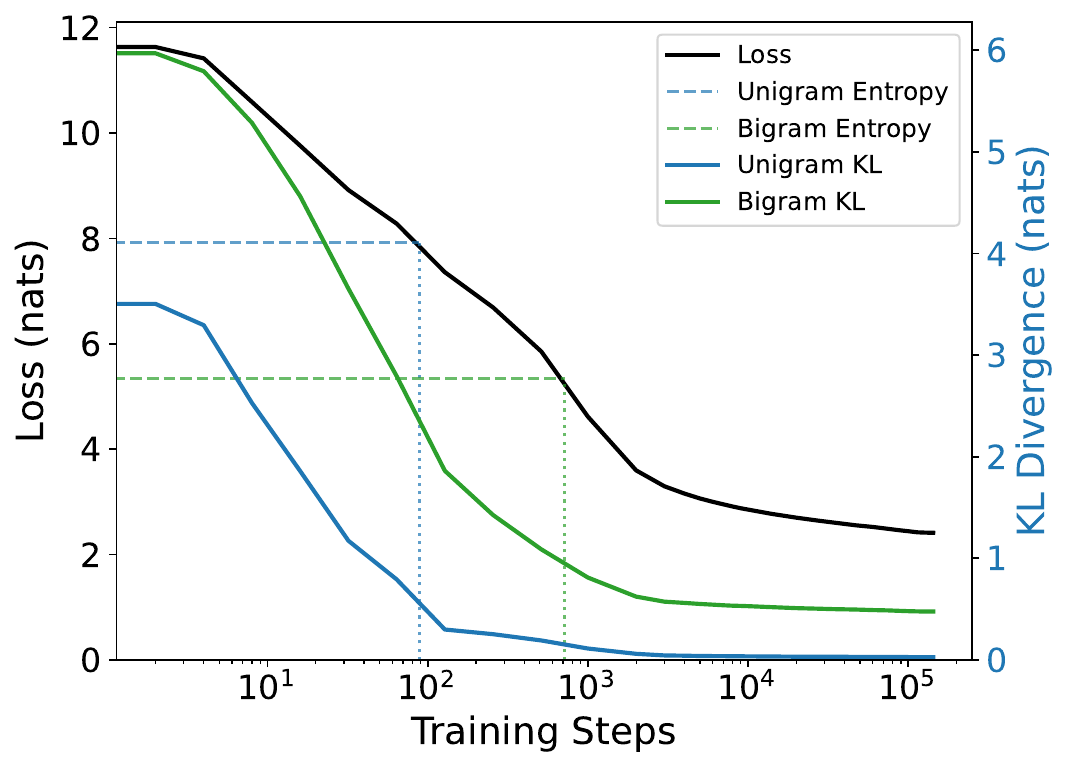}
       \caption{}
       \label{fig:zipfs-6.9b}
   \end{subfigure}
   \hfill
   \begin{subfigure}[b]{0.32\textwidth}
       \centering
       \includegraphics[width=\textwidth]{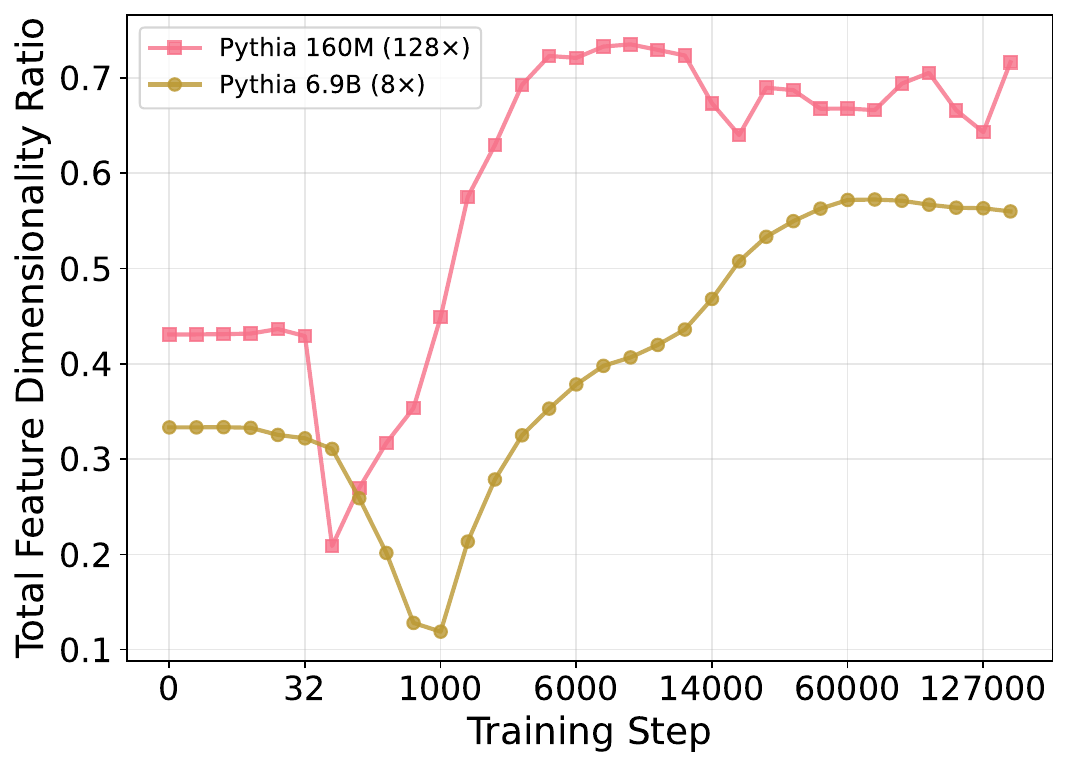}
       \caption{}
       \label{fig:feature-dimensionalities}
   \end{subfigure}
   \caption{Observations supporting a statistical-to-superposition transition in Pythia-160M and Pythia-6.9B. (a) Unigram and bigram KL divergence evolution in Pythia-160M. The convergence timing of KL divergences coincides with when training loss approaches the theoretical minimum (unigram/bigram entropy). (b) Unigram and bigram KL divergence evolution in Pythia-6.9B. (c) Total feature dimensionality ratio (relative to activation space dimension) over training time.}
   \label{fig:phase-transition}
\end{figure*}

What is the model learning at the beginning of training while the training loss rapidly decreases, if no features are formed at this period? We hypothesize that the rapid loss decrease in early training is driven by learning coarse statistical patterns, rather than by forming distinct features. After this initial optimization nears completion, sparse features emerge in superposition for further loss reduction.

This two-stage structure is deeply consistent with the fitting-to-compressing phase transition predicted by the information bottleneck theory~\citep{shwartz2017information}.

\paragraph{Early training almost exclusively learns uni- and bi-gram distributions.} Following previous work on language model learning statistical patterns~\citep{takahashi2017statistical, xu2019frequency, choshen2022grammar, belrose2024statistics, svete2024ngram, nguyen2024ngram, chen2024jetexpansionsngram}, we compute the KL divergence between the true token distribution $Q$ and the model's predicted distribution $P$. We randomly sample 10M tokens from SlimPajama to approximate both distributions. We then evaluate unigram KL divergence $\mathrm{D}_{\mathsf{KL}}(P(x) \parallel Q(x))$ and bigram KL divergence $\mathrm{D}_{\mathsf{KL}}(P(x_i \mid x_{i-1}) \parallel Q(x_i \mid x_{i-1}))$.

The results show that both unigram and bigram KL divergences rapidly converge to low values during early training (Figure~\ref{fig:zipfs-160m} and~\ref{fig:zipfs-6.9b}). Furthermore, the training losses during this period approach the theoretical minimum achievable if the model perfectly matched the true token distributions—i.e., the entropy of these distributions. This suggests that the model primarily learns to fit statistical regularities (Zipf's law~\citep{zipf1935psychobiology, piantadosi2014zipfs}) during the early training stage, which explains the dense nature of internal representations at this phase.

\paragraph{Total feature dimensionality undergoes compression and expansion.} To directly measure superposition status at each snapshot, we adapt the approach from~\citet{elhage2022superposition} and compute the dimensionality of each feature at snapshot $\theta$ as:

\begin{equation}
    D_i = \frac{\|W_{\text{dec},i}^\theta\|^2}{\sum_{j=1}^{n_\text{features}} \left( \hat{W}_{\text{dec},i}^\theta \cdot W_{\text{dec},j}^\theta \right)^2}
\end{equation}

where $\hat{W}_{\text{dec},i}^\theta$ is the normalized version of decoder vector $W_{\text{dec},i}^\theta$. In contrast to its original application in toy models with ground-truth features, we apply this metric to crosscoder features due to their consistent alignment across training snapshots, providing insight into superposition dynamics.

We compute total feature dimensionalities for crosscoders trained on Pythia-160M and Pythia-6.9B (Figure~\ref{fig:feature-dimensionalities}). Under ideal superposition, features should form symmetric arrangements with total dimensionalities summing to the activation space dimension. Feature dimensionalities should go down if large interference exists among features. We observe that total feature dimensionality first decreases then increases around the turning point, eventually reaching approximately 70\% of available dimensions in the crosscoder with 98,304 features for Pythia-160M. Features from Pythia-6.9B account for a smaller proportion of activation space dimensions, likely due to the limited representational capacity of the 32,768-feature crosscoder. Nevertheless, both models exhibit the same trend, suggesting that initialization features initially form weak superposition, then become compressed to accommodate emergent features. This indicates that the model develops into a feature learning phase.

\section{Limitations}

\paragraph{Scope and generalizability.}
Superposition has been proven to be a general phenomenon in deep neural networks~\citep{elhage2022superposition}. However, our analysis focuses on feature evolution in Pythia suite models using their open-source training snapshots. While previous research on feature universality~\citep{wang2025universality} suggest our method might be able to generalize to different architectures, datasets, post training dynamics, and tasks beyond language modeling, the extent to which feature evolution patterns are consistent across diverse settings remains to be established. We leave broader generalization studies to future work.

\paragraph{Limited downstream task complexity.}
Section~\ref{sec:micro-to-macro} establishes the connection between feature evolution and model behavior. However, the downstream tasks we examine are relatively simple, constrained by multiple factors including Pythia model capabilities, sparse dictionary quality, and the current state of circuit tracing methodologies. Scaling to more complex downstream tasks represents a natural direction for future work.

\paragraph{Discrete snapshot requirement.} 
Crosscoder training requires activations from discrete training snapshots, with memory and computational costs scaling linearly with snapshot count, which limits observational granularity. Potential solutions include architectural modifications for online multi-snapshot processing or incorporating gradient information to capture continuous training dynamics.


\section{Conclusion}

We introduce crosscoders to study feature evolution in LLM pre-training. Our analysis reveals two patterns: initialization-dependent features and emergent features, with complex patterns emerging later. We establish causal connections between feature evolution and downstream performance. Supported by uni- and bi-gram distribution analysis and feature dimensionality dynamics, we propose that model pre-training can be roughly divided into a statistical learning phase and a feature learning phase. This work bridges mechanistic interpretability with training dynamics.

\newpage

\section*{Reproducibility Statement}
All our crosscoders are based on open-source models (Pythia suite) and datasets (SlimPajama) with public accessibility. We provide source code for (1) generating model activations on each snapshot, (2) training crosscoders, and (3) analyzing trained crosscoders at \url{https://github.com/OpenMOSS/Language-Model-SAEs}. Detailed instructions for replicating our results are provided in the \texttt{examples/reproduce\_evolution\_of\_concepts/README.md} file. We note that training large crosscoders requires substantial computational resources and disk space, as also listed in the \texttt{README.md} file.

\bibliography{iclr2025_conference}

@misc{grattafiori2024llama3,
  title         = {The Llama 3 Herd of Models},
  author        = {Aaron Grattafiori and Abhimanyu Dubey and Abhinav Jauhri and Abhinav Pandey and Abhishek Kadian and Ahmad Al-Dahle and Aiesha Letman and Akhil Mathur and Alan Schelten and Alex Vaughan and Amy Yang and Angela Fan and Anirudh Goyal and Anthony Hartshorn and Aobo Yang and Archi Mitra and Archie Sravankumar and Artem Korenev and Arthur Hinsvark and Arun Rao and Aston Zhang and Aurelien Rodriguez and Austen Gregerson and Ava Spataru and Baptiste Roziere and Bethany Biron and Binh Tang and Bobbie Chern and Charlotte Caucheteux and Chaya Nayak and Chloe Bi and Chris Marra and Chris McConnell and Christian Keller and Christophe Touret and Chunyang Wu and Corinne Wong and Cristian Canton Ferrer and Cyrus Nikolaidis and Damien Allonsius and Daniel Song and Danielle Pintz and Danny Livshits and Danny Wyatt and David Esiobu and Dhruv Choudhary and Dhruv Mahajan and Diego Garcia-Olano and Diego Perino and Dieuwke Hupkes and Egor Lakomkin and Ehab AlBadawy and Elina Lobanova and Emily Dinan and Eric Michael Smith and Filip Radenovic and Francisco Guzmán and Frank Zhang and Gabriel Synnaeve and Gabrielle Lee and Georgia Lewis Anderson and Govind Thattai and Graeme Nail and Gregoire Mialon and Guan Pang and Guillem Cucurell and Hailey Nguyen and Hannah Korevaar and Hu Xu and Hugo Touvron and Iliyan Zarov and Imanol Arrieta Ibarra and Isabel Kloumann and Ishan Misra and Ivan Evtimov and Jack Zhang and Jade Copet and Jaewon Lee and Jan Geffert and Jana Vranes and Jason Park and Jay Mahadeokar and Jeet Shah and Jelmer van der Linde and Jennifer Billock and Jenny Hong and Jenya Lee and Jeremy Fu and Jianfeng Chi and Jianyu Huang and Jiawen Liu and Jie Wang and Jiecao Yu and Joanna Bitton and Joe Spisak and Jongsoo Park and Joseph Rocca and Joshua Johnstun and Joshua Saxe and Junteng Jia and Kalyan Vasuden Alwala and Karthik Prasad and Kartikeya Upasani and Kate Plawiak and Ke Li and Kenneth Heafield and Kevin Stone and Khalid El-Arini and Krithika Iyer and Kshitiz Malik and Kuenley Chiu and Kunal Bhalla and Kushal Lakhotia and Lauren Rantala-Yeary and Laurens van der Maaten and Lawrence Chen and Liang Tan and Liz Jenkins and Louis Martin and Lovish Madaan and Lubo Malo and Lukas Blecher and Lukas Landzaat and Luke de Oliveira and Madeline Muzzi and Mahesh Pasupuleti and Mannat Singh and Manohar Paluri and Marcin Kardas and Maria Tsimpoukelli and Mathew Oldham and Mathieu Rita and Maya Pavlova and Melanie Kambadur and Mike Lewis and Min Si and Mitesh Kumar Singh and Mona Hassan and Naman Goyal and Narjes Torabi and Nikolay Bashlykov and Nikolay Bogoychev and Niladri Chatterji and Ning Zhang and Olivier Duchenne and Onur Çelebi and Patrick Alrassy and Pengchuan Zhang and Pengwei Li and Petar Vasic and Peter Weng and Prajjwal Bhargava and Pratik Dubal and Praveen Krishnan and Punit Singh Koura and Puxin Xu and Qing He and Qingxiao Dong and Ragavan Srinivasan and Raj Ganapathy and Ramon Calderer and Ricardo Silveira Cabral and Robert Stojnic and Roberta Raileanu and Rohan Maheswari and Rohit Girdhar and Rohit Patel and Romain Sauvestre and Ronnie Polidoro and Roshan Sumbaly and Ross Taylor and Ruan Silva and Rui Hou and Rui Wang and Saghar Hosseini and Sahana Chennabasappa and Sanjay Singh and Sean Bell and Seohyun Sonia Kim and Sergey Edunov and Shaoliang Nie and Sharan Narang and Sharath Raparthy and Sheng Shen and Shengye Wan and Shruti Bhosale and Shun Zhang and Simon Vandenhende and Soumya Batra and Spencer Whitman and Sten Sootla and Stephane Collot and Suchin Gururangan and Sydney Borodinsky and Tamar Herman and Tara Fowler and Tarek Sheasha and Thomas Georgiou and Thomas Scialom and Tobias Speckbacher and Todor Mihaylov and Tong Xiao and Ujjwal Karn and Vedanuj Goswami and Vibhor Gupta and Vignesh Ramanathan and Viktor Kerkez and Vincent Gonguet and Virginie Do and Vish Vogeti and Vítor Albiero and Vladan Petrovic and Weiwei Chu and Wenhan Xiong and Wenyin Fu and Whitney Meers and Xavier Martinet and Xiaodong Wang and Xiaofang Wang and Xiaoqing Ellen Tan and Xide Xia and Xinfeng Xie and Xuchao Jia and Xuewei Wang and Yaelle Goldschlag and Yashesh Gaur and Yasmine Babaei and Yi Wen and Yiwen Song and Yuchen Zhang and Yue Li and Yuning Mao and Zacharie Delpierre Coudert and Zheng Yan and Zhengxing Chen and Zoe Papakipos and Aaditya Singh and Aayushi Srivastava and Abha Jain and Adam Kelsey and Adam Shajnfeld and Adithya Gangidi and Adolfo Victoria and Ahuva Goldstand and Ajay Menon and Ajay Sharma and Alex Boesenberg and Alexei Baevski and Allie Feinstein and Amanda Kallet and Amit Sangani and Amos Teo and Anam Yunus and Andrei Lupu and Andres Alvarado and Andrew Caples and Andrew Gu and Andrew Ho and Andrew Poulton and Andrew Ryan and Ankit Ramchandani and Annie Dong and Annie Franco and Anuj Goyal and Aparajita Saraf and Arkabandhu Chowdhury and Ashley Gabriel and Ashwin Bharambe and Assaf Eisenman and Azadeh Yazdan and Beau James and Ben Maurer and Benjamin Leonhardi and Bernie Huang and Beth Loyd and Beto De Paola and Bhargavi Paranjape and Bing Liu and Bo Wu and Boyu Ni and Braden Hancock and Bram Wasti and Brandon Spence and Brani Stojkovic and Brian Gamido and Britt Montalvo and Carl Parker and Carly Burton and Catalina Mejia and Ce Liu and Changhan Wang and Changkyu Kim and Chao Zhou and Chester Hu and Ching-Hsiang Chu and Chris Cai and Chris Tindal and Christoph Feichtenhofer and Cynthia Gao and Damon Civin and Dana Beaty and Daniel Kreymer and Daniel Li and David Adkins and David Xu and Davide Testuggine and Delia David and Devi Parikh and Diana Liskovich and Didem Foss and Dingkang Wang and Duc Le and Dustin Holland and Edward Dowling and Eissa Jamil and Elaine Montgomery and Eleonora Presani and Emily Hahn and Emily Wood and Eric-Tuan Le and Erik Brinkman and Esteban Arcaute and Evan Dunbar and Evan Smothers and Fei Sun and Felix Kreuk and Feng Tian and Filippos Kokkinos and Firat Ozgenel and Francesco Caggioni and Frank Kanayet and Frank Seide and Gabriela Medina Florez and Gabriella Schwarz and Gada Badeer and Georgia Swee and Gil Halpern and Grant Herman and Grigory Sizov and Guangyi and Zhang and Guna Lakshminarayanan and Hakan Inan and Hamid Shojanazeri and Han Zou and Hannah Wang and Hanwen Zha and Haroun Habeeb and Harrison Rudolph and Helen Suk and Henry Aspegren and Hunter Goldman and Hongyuan Zhan and Ibrahim Damlaj and Igor Molybog and Igor Tufanov and Ilias Leontiadis and Irina-Elena Veliche and Itai Gat and Jake Weissman and James Geboski and James Kohli and Janice Lam and Japhet Asher and Jean-Baptiste Gaya and Jeff Marcus and Jeff Tang and Jennifer Chan and Jenny Zhen and Jeremy Reizenstein and Jeremy Teboul and Jessica Zhong and Jian Jin and Jingyi Yang and Joe Cummings and Jon Carvill and Jon Shepard and Jonathan McPhie and Jonathan Torres and Josh Ginsburg and Junjie Wang and Kai Wu and Kam Hou U and Karan Saxena and Kartikay Khandelwal and Katayoun Zand and Kathy Matosich and Kaushik Veeraraghavan and Kelly Michelena and Keqian Li and Kiran Jagadeesh and Kun Huang and Kunal Chawla and Kyle Huang and Lailin Chen and Lakshya Garg and Lavender A and Leandro Silva and Lee Bell and Lei Zhang and Liangpeng Guo and Licheng Yu and Liron Moshkovich and Luca Wehrstedt and Madian Khabsa and Manav Avalani and Manish Bhatt and Martynas Mankus and Matan Hasson and Matthew Lennie and Matthias Reso and Maxim Groshev and Maxim Naumov and Maya Lathi and Meghan Keneally and Miao Liu and Michael L. Seltzer and Michal Valko and Michelle Restrepo and Mihir Patel and Mik Vyatskov and Mikayel Samvelyan and Mike Clark and Mike Macey and Mike Wang and Miquel Jubert Hermoso and Mo Metanat and Mohammad Rastegari and Munish Bansal and Nandhini Santhanam and Natascha Parks and Natasha White and Navyata Bawa and Nayan Singhal and Nick Egebo and Nicolas Usunier and Nikhil Mehta and Nikolay Pavlovich Laptev and Ning Dong and Norman Cheng and Oleg Chernoguz and Olivia Hart and Omkar Salpekar and Ozlem Kalinli and Parkin Kent and Parth Parekh and Paul Saab and Pavan Balaji and Pedro Rittner and Philip Bontrager and Pierre Roux and Piotr Dollar and Polina Zvyagina and Prashant Ratanchandani and Pritish Yuvraj and Qian Liang and Rachad Alao and Rachel Rodriguez and Rafi Ayub and Raghotham Murthy and Raghu Nayani and Rahul Mitra and Rangaprabhu Parthasarathy and Raymond Li and Rebekkah Hogan and Robin Battey and Rocky Wang and Russ Howes and Ruty Rinott and Sachin Mehta and Sachin Siby and Sai Jayesh Bondu and Samyak Datta and Sara Chugh and Sara Hunt and Sargun Dhillon and Sasha Sidorov and Satadru Pan and Saurabh Mahajan and Saurabh Verma and Seiji Yamamoto and Sharadh Ramaswamy and Shaun Lindsay and Shaun Lindsay and Sheng Feng and Shenghao Lin and Shengxin Cindy Zha and Shishir Patil and Shiva Shankar and Shuqiang Zhang and Shuqiang Zhang and Sinong Wang and Sneha Agarwal and Soji Sajuyigbe and Soumith Chintala and Stephanie Max and Stephen Chen and Steve Kehoe and Steve Satterfield and Sudarshan Govindaprasad and Sumit Gupta and Summer Deng and Sungmin Cho and Sunny Virk and Suraj Subramanian and Sy Choudhury and Sydney Goldman and Tal Remez and Tamar Glaser and Tamara Best and Thilo Koehler and Thomas Robinson and Tianhe Li and Tianjun Zhang and Tim Matthews and Timothy Chou and Tzook Shaked and Varun Vontimitta and Victoria Ajayi and Victoria Montanez and Vijai Mohan and Vinay Satish Kumar and Vishal Mangla and Vlad Ionescu and Vlad Poenaru and Vlad Tiberiu Mihailescu and Vladimir Ivanov and Wei Li and Wenchen Wang and Wenwen Jiang and Wes Bouaziz and Will Constable and Xiaocheng Tang and Xiaojian Wu and Xiaolan Wang and Xilun Wu and Xinbo Gao and Yaniv Kleinman and Yanjun Chen and Ye Hu and Ye Jia and Ye Qi and Yenda Li and Yilin Zhang and Ying Zhang and Yossi Adi and Youngjin Nam and Yu and Wang and Yu Zhao and Yuchen Hao and Yundi Qian and Yunlu Li and Yuzi He and Zach Rait and Zachary DeVito and Zef Rosnbrick and Zhaoduo Wen and Zhenyu Yang and Zhiwei Zhao and Zhiyu Ma},
  year          = {2024},
  eprint        = {2407.21783},
  archiveprefix = {arXiv},
  primaryclass  = {cs.AI},
  url           = {https://arxiv.org/abs/2407.21783}
}

@misc{yang2025qwen3,
  title         = {Qwen3 Technical Report},
  author        = {An Yang and Anfeng Li and Baosong Yang and Beichen Zhang and Binyuan Hui and Bo Zheng and Bowen Yu and Chang Gao and Chengen Huang and Chenxu Lv and Chujie Zheng and Dayiheng Liu and Fan Zhou and Fei Huang and Feng Hu and Hao Ge and Haoran Wei and Huan Lin and Jialong Tang and Jian Yang and Jianhong Tu and Jianwei Zhang and Jianxin Yang and Jiaxi Yang and Jing Zhou and Jingren Zhou and Junyang Lin and Kai Dang and Keqin Bao and Kexin Yang and Le Yu and Lianghao Deng and Mei Li and Mingfeng Xue and Mingze Li and Pei Zhang and Peng Wang and Qin Zhu and Rui Men and Ruize Gao and Shixuan Liu and Shuang Luo and Tianhao Li and Tianyi Tang and Wenbiao Yin and Xingzhang Ren and Xinyu Wang and Xinyu Zhang and Xuancheng Ren and Yang Fan and Yang Su and Yichang Zhang and Yinger Zhang and Yu Wan and Yuqiong Liu and Zekun Wang and Zeyu Cui and Zhenru Zhang and Zhipeng Zhou and Zihan Qiu},
  year          = {2025},
  eprint        = {2505.09388},
  archiveprefix = {arXiv},
  primaryclass  = {cs.CL},
  url           = {https://arxiv.org/abs/2505.09388}
}

@misc{openai2024gpt4,
  title         = {{GPT-4} Technical Report},
  author        = {OpenAI},
  year          = {2023},
  eprint        = {2303.08774},
  archiveprefix = {arXiv},
  primaryclass  = {cs.CL},
  url           = {https://arxiv.org/abs/2303.08774}
}

@article{lindsey2024crosscoder,
  title   = {Sparse crosscoders for cross-layer features and model diffing},
  author  = {Lindsey, Jack and Templeton, Adly and Marcus, Jonathan and Conerly, Thomas and Batson, Joshua and Olah, Christopher},
  journal = {Transformer Circuits Thread},
  note    = {https://transformer-circuits.pub/2024/crosscoders/index.html},
  year    = {2024}
}

@article{bricken2023monosemanticity,
  title   = {Towards Monosemanticity: Decomposing Language Models With Dictionary Learning},
  author  = {Bricken, Trenton and Templeton, Adly and Batson, Joshua and Chen, Brian and Jermyn, Adam and Conerly, Tom and Turner, Nick and Anil, Cem and Denison, Carson and Askell, Amanda and Lasenby, Robert and Wu, Yifan and Kravec, Shauna and Schiefer, Nicholas and Maxwell, Tim and Joseph, Nicholas and Hatfield-Dodds, Zac and Tamkin, Alex and Nguyen, Karina and McLean, Brayden and Burke, Josiah E and Hume, Tristan and Carter, Shan and Henighan, Tom and Olah, Christopher},
  year    = {2023},
  journal = {Transformer Circuits Thread},
  note    = {https://transformer-circuits.pub/2023/monosemantic-features/index.html}
}

@article{templeton2024scalingmonosemanticity,
  title   = {Scaling Monosemanticity: Extracting Interpretable Features from Claude 3 Sonnet},
  author  = {Templeton, Adly and Conerly, Tom and Marcus, Jonathan and Lindsey, Jack and Bricken, Trenton and Chen, Brian and Pearce, Adam and Citro, Craig and Ameisen, Emmanuel and Jones, Andy and Cunningham, Hoagy and Turner, Nicholas L and McDougall, Callum and MacDiarmid, Monte and Freeman, C. Daniel and Sumers, Theodore R. and Rees, Edward and Batson, Joshua and Jermyn, Adam and Carter, Shan and Olah, Chris and Henighan, Tom},
  year    = {2024},
  journal = {Transformer Circuits Thread},
  url     = {https://transformer-circuits.pub/2024/scaling-monosemanticity/index.html}
}

@inproceedings{gao2025scalingsae,
  author    = {Leo Gao and
               Tom Dupr{\'{e}} la Tour and
               Henk Tillman and
               Gabriel Goh and
               Rajan Troll and
               Alec Radford and
               Ilya Sutskever and
               Jan Leike and
               Jeffrey Wu},
  title     = {Scaling and evaluating sparse autoencoders},
  booktitle = {The Thirteenth International Conference on Learning Representations,
               {ICLR} 2025, Singapore, April 24-28, 2025},
  publisher = {OpenReview.net},
  year      = {2025},
  url       = {https://openreview.net/forum?id=tcsZt9ZNKD},
  bibsource = {dblp computer science bibliography, https://dblp.org}
}

@article{elhage2022superposition,
  title   = {Toy Models of Superposition},
  author  = {Elhage, Nelson and Hume, Tristan and Olsson, Catherine and Schiefer, Nicholas and Henighan, Tom and Kravec, Shauna and Hatfield-Dodds, Zac and Lasenby, Robert and Drain, Dawn and Chen, Carol and Grosse, Roger and McCandlish, Sam and Kaplan, Jared and Amodei, Dario and Wattenberg, Martin and Olah, Christopher},
  year    = {2022},
  journal = {Transformer Circuits Thread},
  note    = {https://transformer-circuits.pub/2022/toy\_model/index.html}
}

@article{hestness2017scaling,
  author     = {Joel Hestness and
                Sharan Narang and
                Newsha Ardalani and
                Gregory F. Diamos and
                Heewoo Jun and
                Hassan Kianinejad and
                Md. Mostofa Ali Patwary and
                Yang Yang and
                Yanqi Zhou},
  title      = {Deep Learning Scaling is Predictable, Empirically},
  journal    = {CoRR},
  volume     = {abs/1712.00409},
  year       = {2017},
  url        = {http://arxiv.org/abs/1712.00409},
  eprinttype = {arXiv},
  eprint     = {1712.00409},
  bibsource  = {dblp computer science bibliography, https://dblp.org}
}

@article{kaplan2020scalinglaws,
  author     = {Jared Kaplan and
                Sam McCandlish and
                Tom Henighan and
                Tom B. Brown and
                Benjamin Chess and
                Rewon Child and
                Scott Gray and
                Alec Radford and
                Jeffrey Wu and
                Dario Amodei},
  title      = {Scaling Laws for Neural Language Models},
  journal    = {CoRR},
  volume     = {abs/2001.08361},
  year       = {2020},
  url        = {https://arxiv.org/abs/2001.08361},
  eprinttype = {arXiv},
  eprint     = {2001.08361},
  bibsource  = {dblp computer science bibliography, https://dblp.org}
}

@article{bahri2021explainscalinglaws,
  author     = {Yasaman Bahri and
                Ethan Dyer and
                Jared Kaplan and
                Jaehoon Lee and
                Utkarsh Sharma},
  title      = {Explaining Neural Scaling Laws},
  journal    = {CoRR},
  volume     = {abs/2102.06701},
  year       = {2021},
  url        = {https://arxiv.org/abs/2102.06701},
  eprinttype = {arXiv},
  eprint     = {2102.06701},
  bibsource  = {dblp computer science bibliography, https://dblp.org}
}

@inproceedings{jacot2018ntk,
  author    = {Arthur Jacot and
               Cl{\'{e}}ment Hongler and
               Franck Gabriel},
  editor    = {Samy Bengio and
               Hanna M. Wallach and
               Hugo Larochelle and
               Kristen Grauman and
               Nicol{\`{o}} Cesa{-}Bianchi and
               Roman Garnett},
  title     = {Neural Tangent Kernel: Convergence and Generalization in Neural Networks},
  booktitle = {Advances in Neural Information Processing Systems 31: Annual Conference
               on Neural Information Processing Systems 2018, NeurIPS 2018, December
               3-8, 2018, Montr{\'{e}}al, Canada},
  pages     = {8580--8589},
  year      = {2018},
  url       = {https://proceedings.neurips.cc/paper/2018/hash/5a4be1fa34e62bb8a6ec6b91d2462f5a-Abstract.html},
  bibsource = {dblp computer science bibliography, https://dblp.org}
}

@inproceedings{tishby2015informationbottleneck,
  author    = {Naftali Tishby and
               Noga Zaslavsky},
  title     = {Deep learning and the information bottleneck principle},
  booktitle = {2015 {IEEE} Information Theory Workshop, {ITW} 2015, Jerusalem, Israel,
               April 26 - May 1, 2015},
  pages     = {1--5},
  publisher = {{IEEE}},
  year      = {2015},
  url       = {https://doi.org/10.1109/ITW.2015.7133169},
  doi       = {10.1109/ITW.2015.7133169},
  bibsource = {dblp computer science bibliography, https://dblp.org}
}

@article{shwartz2017information,
  author     = {Ravid Shwartz{-}Ziv and
                Naftali Tishby},
  title      = {Opening the Black Box of Deep Neural Networks via Information},
  journal    = {CoRR},
  volume     = {abs/1703.00810},
  year       = {2017},
  url        = {http://arxiv.org/abs/1703.00810},
  eprinttype = {arXiv},
  eprint     = {1703.00810},
  bibsource  = {dblp computer science bibliography, https://dblp.org}
}

@book{watanabe2009slt,
  place      = {Cambridge},
  series     = {Cambridge Monographs on Applied and Computational Mathematics},
  title      = {Algebraic Geometry and Statistical Learning Theory},
  publisher  = {Cambridge University Press},
  author     = {Watanabe, Sumio},
  year       = {2009},
  collection = {Cambridge Monographs on Applied and Computational Mathematics}
}

@misc{lau2024llc,
      title={The Local Learning Coefficient: A Singularity-Aware Complexity Measure}, 
      author={Edmund Lau and Zach Furman and George Wang and Daniel Murfet and Susan Wei},
      year={2024},
      eprint={2308.12108},
      archivePrefix={arXiv},
      primaryClass={stat.ML},
      url={https://arxiv.org/abs/2308.12108}, 
}

@inproceedings{wang2025rllc,
  author       = {George Wang and
                  Jesse Hoogland and
                  Stan van Wingerden and
                  Zach Furman and
                  Daniel Murfet},
  title        = {Differentiation and Specialization of Attention Heads via the Refined
                  Local Learning Coefficient},
  booktitle    = {The Thirteenth International Conference on Learning Representations,
                  {ICLR} 2025, Singapore, April 24-28, 2025},
  publisher    = {OpenReview.net},
  year         = {2025},
  url          = {https://openreview.net/forum?id=SUc1UOWndp},
  bibsource    = {dblp computer science bibliography, https://dblp.org}
}

@inproceedings{biderman2023pythia,
  author       = {Stella Biderman and
                  Hailey Schoelkopf and
                  Quentin Gregory Anthony and
                  Herbie Bradley and
                  Kyle O'Brien and
                  Eric Hallahan and
                  Mohammad Aflah Khan and
                  Shivanshu Purohit and
                  USVSN Sai Prashanth and
                  Edward Raff and
                  Aviya Skowron and
                  Lintang Sutawika and
                  Oskar van der Wal},
  editor       = {Andreas Krause and
                  Emma Brunskill and
                  Kyunghyun Cho and
                  Barbara Engelhardt and
                  Sivan Sabato and
                  Jonathan Scarlett},
  title        = {Pythia: {A} Suite for Analyzing Large Language Models Across Training
                  and Scaling},
  booktitle    = {International Conference on Machine Learning, {ICML} 2023, 23-29 July
                  2023, Honolulu, Hawaii, {USA}},
  series       = {Proceedings of Machine Learning Research},
  volume       = {202},
  pages        = {2397--2430},
  publisher    = {{PMLR}},
  year         = {2023},
  url          = {https://proceedings.mlr.press/v202/biderman23a.html},
  bibsource    = {dblp computer science bibliography, https://dblp.org}
}

@article{power2022grokking,
  author       = {Alethea Power and
                  Yuri Burda and
                  Harri Edwards and
                  Igor Babuschkin and
                  Vedant Misra},
  title        = {Grokking: Generalization Beyond Overfitting on Small Algorithmic Datasets},
  journal      = {CoRR},
  volume       = {abs/2201.02177},
  year         = {2022},
  url          = {https://arxiv.org/abs/2201.02177},
  eprinttype    = {arXiv},
  eprint       = {2201.02177},
  bibsource    = {dblp computer science bibliography, https://dblp.org}
}

@misc{nanda2023grokking,
  doi = {10.48550/ARXIV.2301.05217},
  
  url = {https://arxiv.org/abs/2301.05217},
  
  author = {Nanda, Neel and Chan, Lawrence and Lieberum, Tom and Smith, Jess and Steinhardt, Jacob},
  
  title = {Progress measures for grokking via mechanistic interpretability},
  
  publisher = {arXiv},
  
  year = {2023},
  
  copyright = {arXiv.org perpetual, non-exclusive license}
}

@article{radford2018gpt,
  title={Improving Language Understanding by Generative Pre-Training},
  author={Alec Radford and Karthik Narasimhan and Tim Salimans and Ilya Sutskever},
  journal={Technical Report},
  publisher = {OpenAI},
  year={2018}
}

@inproceedings{devlin2019bert,
  author       = {Jacob Devlin and
                  Ming{-}Wei Chang and
                  Kenton Lee and
                  Kristina Toutanova},
  editor       = {Jill Burstein and
                  Christy Doran and
                  Thamar Solorio},
  title        = {{BERT:} Pre-training of Deep Bidirectional Transformers for Language
                  Understanding},
  booktitle    = {Proceedings of the 2019 Conference of the North American Chapter of
                  the Association for Computational Linguistics: Human Language Technologies,
                  {NAACL-HLT} 2019, Minneapolis, MN, USA, June 2-7, 2019, Volume 1 (Long
                  and Short Papers)},
  pages        = {4171--4186},
  publisher    = {Association for Computational Linguistics},
  year         = {2019},
  url          = {https://doi.org/10.18653/v1/n19-1423},
  doi          = {10.18653/V1/N19-1423},
  bibsource    = {dblp computer science bibliography, https://dblp.org}
}

@article{shen2023slimpajama,
  author       = {Zhiqiang Shen and
                  Tianhua Tao and
                  Liqun Ma and
                  Willie Neiswanger and
                  Zhengzhong Liu and
                  Hongyi Wang and
                  Bowen Tan and
                  Joel Hestness and
                  Natalia Vassilieva and
                  Daria Soboleva and
                  Eric P. Xing},
  title        = {SlimPajama-DC: Understanding Data Combinations for {LLM} Training},
  journal      = {CoRR},
  volume       = {abs/2309.10818},
  year         = {2023},
  url          = {https://doi.org/10.48550/arXiv.2309.10818},
  doi          = {10.48550/ARXIV.2309.10818},
  eprinttype    = {arXiv},
  eprint       = {2309.10818},
  bibsource    = {dblp computer science bibliography, https://dblp.org}
}

@article{olshausen1997sparsecoding,
    title = {Sparse coding with an overcomplete basis set: A strategy employed by V1?},
    journal = {Vision Research},
    volume = {37},
    number = {23},
    pages = {3311-3325},
    year = {1997},
    issn = {0042-6989},
    doi = {https://doi.org/10.1016/S0042-6989(97)00169-7},
    url = {https://www.sciencedirect.com/science/article/pii/S0042698997001697},
    author = {Bruno A. Olshausen and David J. Field}
}

@inproceedings{hubun2024sae,
  author       = {Robert Huben and
                  Hoagy Cunningham and
                  Logan Riggs Smith and
                  Aidan Ewart and
                  Lee Sharkey},
  title        = {Sparse Autoencoders Find Highly Interpretable Features in Language
                  Models},
  booktitle    = {The Twelfth International Conference on Learning Representations,
                  {ICLR} 2024, Vienna, Austria, May 7-11, 2024},
  publisher    = {OpenReview.net},
  year         = {2024},
  url          = {https://openreview.net/forum?id=F76bwRSLeK},
  bibsource    = {dblp computer science bibliography, https://dblp.org}
}

@article{rajamanoharan2024jumprelu,
  author       = {Senthooran Rajamanoharan and
                  Tom Lieberum and
                  Nicolas Sonnerat and
                  Arthur Conmy and
                  Vikrant Varma and
                  J{\'{a}}nos Kram{\'{a}}r and
                  Neel Nanda},
  title        = {Jumping Ahead: Improving Reconstruction Fidelity with JumpReLU Sparse
                  Autoencoders},
  journal      = {CoRR},
  volume       = {abs/2407.14435},
  year         = {2024},
  url          = {https://doi.org/10.48550/arXiv.2407.14435},
  doi          = {10.48550/ARXIV.2407.14435},
  eprinttype    = {arXiv},
  eprint       = {2407.14435},
  bibsource    = {dblp computer science bibliography, https://dblp.org}
}

@misc{smith2025negativesae,
  author = {Lewis Smith and Sen Rajamanoharan and Arthur Conmy and Callum McDougall and Janos Kramar and Tom Lieberum and Rohin Shah and Neel Nanda},
  title = {Negative Results for {SAEs} on Downstream Tasks and Deprioritising {SAE} Research ({GDM} Mech Interp Team Progress Update \#2)},
  year = {2025},
  url = {https://www.lesswrong.com/posts/4uXCAJNuPKtKBsi28/negative-results-for-saes-on-downstream-tasks},
  note = {LessWrong},
  urldate = {2025-08-15}
}

@article{bengio2013ste,
  author       = {Yoshua Bengio and
                  Nicholas L{\'{e}}onard and
                  Aaron C. Courville},
  title        = {Estimating or Propagating Gradients Through Stochastic Neurons for
                  Conditional Computation},
  journal      = {CoRR},
  volume       = {abs/1308.3432},
  year         = {2013},
  url          = {http://arxiv.org/abs/1308.3432},
  eprinttype    = {arXiv},
  eprint       = {1308.3432},
  bibsource    = {dblp computer science bibliography, https://dblp.org}
}

@misc{heap2025saerandom,
      title={Sparse Autoencoders Can Interpret Randomly Initialized Transformers}, 
      author={Thomas Heap and Tim Lawson and Lucy Farnik and Laurence Aitchison},
      year={2025},
      eprint={2501.17727},
      archivePrefix={arXiv},
      primaryClass={cs.LG},
      url={https://arxiv.org/abs/2501.17727}, 
}

@article{olsson2022induction,
   title={In-context Learning and Induction Heads},
   author={Olsson, Catherine and Elhage, Nelson and Nanda, Neel and Joseph, Nicholas and DasSarma, Nova and Henighan, Tom and Mann, Ben and Askell, Amanda and Bai, Yuntao and Chen, Anna and Conerly, Tom and Drain, Dawn and Ganguli, Deep and Hatfield-Dodds, Zac and Hernandez, Danny and Johnston, Scott and Jones, Andy and Kernion, Jackson and Lovitt, Liane and Ndousse, Kamal and Amodei, Dario and Brown, Tom and Clark, Jack and Kaplan, Jared and McCandlish, Sam and Olah, Chris},
   year={2022},
   journal={Transformer Circuits Thread},
   note={https://transformer-circuits.pub/2022/in-context-learning-and-induction-heads/index.html}
}

@article{gurnee2024universalneuron,
  author       = {Wes Gurnee and
                  Theo Horsley and
                  Zifan Carl Guo and
                  Tara Rezaei Kheirkhah and
                  Qinyi Sun and
                  Will Hathaway and
                  Neel Nanda and
                  Dimitris Bertsimas},
  title        = {Universal Neurons in {GPT2} Language Models},
  journal      = {Trans. Mach. Learn. Res.},
  volume       = {2024},
  year         = {2024},
  url          = {https://openreview.net/forum?id=ZeI104QZ8I},
  bibsource    = {dblp computer science bibliography, https://dblp.org}
}

@inproceedings{marks2025sparsefeaturecircuits,
  author       = {Samuel Marks and
                  Can Rager and
                  Eric J. Michaud and
                  Yonatan Belinkov and
                  David Bau and
                  Aaron Mueller},
  title        = {Sparse Feature Circuits: Discovering and Editing Interpretable Causal
                  Graphs in Language Models},
  booktitle    = {The Thirteenth International Conference on Learning Representations,
                  {ICLR} 2025, Singapore, April 24-28, 2025},
  publisher    = {OpenReview.net},
  year         = {2025},
  url          = {https://openreview.net/forum?id=I4e82CIDxv},
  bibsource    = {dblp computer science bibliography, https://dblp.org}
}

@article{syed2023attributionpatching,
  author       = {Aaquib Syed and
                  Can Rager and
                  Arthur Conmy},
  title        = {Attribution Patching Outperforms Automated Circuit Discovery},
  journal      = {CoRR},
  volume       = {abs/2310.10348},
  year         = {2023},
  url          = {https://doi.org/10.48550/arXiv.2310.10348},
  doi          = {10.48550/ARXIV.2310.10348},
  eprinttype    = {arXiv},
  eprint       = {2310.10348},
  bibsource    = {dblp computer science bibliography, https://dblp.org}
}

@misc{ge2024automatically,
    title={Automatically Identifying Local and Global Circuits with Linear Computation Graphs},
    author={Xuyang Ge and Fukang Zhu and Wentao Shu and Junxuan Wang and Zhengfu He and Xipeng Qiu},
    year={2024},
    eprint={2405.13868},
    archivePrefix={arXiv},
    primaryClass={cs.LG}
}

@inproceedings{wang2023ioi,
  author       = {Kevin Ro Wang and
                  Alexandre Variengien and
                  Arthur Conmy and
                  Buck Shlegeris and
                  Jacob Steinhardt},
  title        = {Interpretability in the Wild: a Circuit for Indirect Object Identification
                  in {GPT-2} Small},
  booktitle    = {The Eleventh International Conference on Learning Representations,
                  {ICLR} 2023, Kigali, Rwanda, May 1-5, 2023},
  publisher    = {OpenReview.net},
  year         = {2023},
  url          = {https://openreview.net/forum?id=NpsVSN6o4ul},
  bibsource    = {dblp computer science bibliography, https://dblp.org}
}

@inproceedings{matthew2021sva,
    title = "Causal Analysis of Syntactic Agreement Mechanisms in Neural Language Models",
    author = "Finlayson, Matthew  and
      Mueller, Aaron  and
      Gehrmann, Sebastian  and
      Shieber, Stuart  and
      Linzen, Tal  and
      Belinkov, Yonatan",
    editor = "Zong, Chengqing  and
      Xia, Fei  and
      Li, Wenjie  and
      Navigli, Roberto",
    booktitle = "Proceedings of the 59th Annual Meeting of the Association for Computational Linguistics and the 11th International Joint Conference on Natural Language Processing (Volume 1: Long Papers)",
    month = aug,
    year = "2021",
    address = "Online",
    publisher = "Association for Computational Linguistics",
    url = "https://aclanthology.org/2021.acl-long.144/",
    doi = "10.18653/v1/2021.acl-long.144",
    pages = "1828--1843"
}

@book{zipf1935psychobiology,
  address = {New York, NY, USA},
  author = {Zipf, George Kingsley},
  publisher = {Houghton-Mifflin},
  title = {The Psychobiology of Language},
  year = 1935
}

@article{piantadosi2014zipfs,
  title={Zipf’s word frequency law in natural language: A critical review and future directions},
  author={Steven T. Piantadosi},
  journal={Psychonomic Bulletin \& Review},
  year={2014},
  volume={21},
  pages={1112 - 1130},
  url={https://api.semanticscholar.org/CorpusID:14264582}
}

@inproceedings{dyer2020asymptotics,
  author       = {Ethan Dyer and
                  Guy Gur{-}Ari},
  title        = {Asymptotics of Wide Networks from Feynman Diagrams},
  booktitle    = {8th International Conference on Learning Representations, {ICLR} 2020,
                  Addis Ababa, Ethiopia, April 26-30, 2020},
  publisher    = {OpenReview.net},
  year         = {2020},
  url          = {https://openreview.net/forum?id=S1gFvANKDS},
  bibsource    = {dblp computer science bibliography, https://dblp.org}
}

@InProceedings{yang2021tensorprograms2b,
  title = 	 {Tensor Programs IIb: Architectural Universality Of Neural Tangent Kernel Training Dynamics},
  author =       {Yang, Greg and Littwin, Etai},
  booktitle = 	 {Proceedings of the 38th International Conference on Machine Learning},
  pages = 	 {11762--11772},
  year = 	 {2021},
  editor = 	 {Meila, Marina and Zhang, Tong},
  volume = 	 {139},
  series = 	 {Proceedings of Machine Learning Research},
  month = 	 {18--24 Jul},
  publisher =    {PMLR},
  url = 	 {https://proceedings.mlr.press/v139/yang21f.html}
}

@article{yang2020tensorprograms3,
  author       = {Greg Yang},
  title        = {Tensor Programs {III:} Neural Matrix Laws},
  journal      = {CoRR},
  volume       = {abs/2009.10685},
  year         = {2020},
  url          = {https://arxiv.org/abs/2009.10685},
  eprinttype    = {arXiv},
  eprint       = {2009.10685},
  bibsource    = {dblp computer science bibliography, https://dblp.org}
}

@inproceedings{kumar2024grokkingntk,
  author       = {Tanishq Kumar and
                  Blake Bordelon and
                  Samuel J. Gershman and
                  Cengiz Pehlevan},
  title        = {Grokking as the transition from lazy to rich training dynamics},
  booktitle    = {The Twelfth International Conference on Learning Representations,
                  {ICLR} 2024, Vienna, Austria, May 7-11, 2024},
  publisher    = {OpenReview.net},
  year         = {2024},
  url          = {https://openreview.net/forum?id=vt5mnLVIVo},
  bibsource    = {dblp computer science bibliography, https://dblp.org}
}

@misc{zhou2025twophase,
      title={New Evidence of the Two-Phase Learning Dynamics of Neural Networks}, 
      author={Zhanpeng Zhou and Yongyi Yang and Mahito Sugiyama and Junchi Yan},
      year={2025},
      eprint={2505.13900},
      archivePrefix={arXiv},
      primaryClass={cs.LG},
      url={https://arxiv.org/abs/2505.13900}, 
}

@inproceedings{watanabe1999slt,
  author       = {Sumio Watanabe},
  editor       = {Sara A. Solla and
                  Todd K. Leen and
                  Klaus{-}Robert M{\"{u}}ller},
  title        = {Algebraic Analysis for Non-regular Learning Machines},
  booktitle    = {Advances in Neural Information Processing Systems 12, {[NIPS} Conference,
                  Denver, Colorado, USA, November 29 - December 4, 1999]},
  pages        = {356--362},
  publisher    = {The {MIT} Press},
  year         = {1999},
  url          = {http://papers.nips.cc/paper/1739-algebraic-analysis-for-non-regular-learning-machines},
  bibsource    = {dblp computer science bibliography, https://dblp.org}
}

@misc{furman2024llc,
      title={Estimating the Local Learning Coefficient at Scale}, 
      author={Zach Furman and Edmund Lau},
      year={2024},
      eprint={2402.03698},
      archivePrefix={arXiv},
      primaryClass={cs.LG},
      url={https://arxiv.org/abs/2402.03698}, 
}

@misc{bengio2014representationlearning,
      title={Representation Learning: A Review and New Perspectives}, 
      author={Yoshua Bengio and Aaron Courville and Pascal Vincent},
      year={2014},
      eprint={1206.5538},
      archivePrefix={arXiv},
      primaryClass={cs.LG},
      url={https://arxiv.org/abs/1206.5538}, 
}

@article{olah2020zoom,
  author = {Olah, Chris and Cammarata, Nick and Schubert, Ludwig and Goh, Gabriel and Petrov, Michael and Carter, Shan},
  title = {Zoom In: An Introduction to Circuits},
  journal = {Distill},
  year = {2020},
  note = {https://distill.pub/2020/circuits/zoom-in},
  doi = {10.23915/distill.00024.001}
}

@inproceedings{alain2017linearprobe,
  author       = {Guillaume Alain and
                  Yoshua Bengio},
  title        = {Understanding intermediate layers using linear classifier probes},
  booktitle    = {5th International Conference on Learning Representations, {ICLR} 2017,
                  Toulon, France, April 24-26, 2017, Workshop Track Proceedings},
  publisher    = {OpenReview.net},
  year         = {2017},
  url          = {https://openreview.net/forum?id=HJ4-rAVtl},
  bibsource    = {dblp computer science bibliography, https://dblp.org}
}

@inproceedings{vargas2020linear,
  author       = {Francisco Vargas and
                  Ryan Cotterell},
  editor       = {Bonnie Webber and
                  Trevor Cohn and
                  Yulan He and
                  Yang Liu},
  title        = {Exploring the Linear Subspace Hypothesis in Gender Bias Mitigation},
  booktitle    = {Proceedings of the 2020 Conference on Empirical Methods in Natural
                  Language Processing, {EMNLP} 2020, Online, November 16-20, 2020},
  pages        = {2902--2913},
  publisher    = {Association for Computational Linguistics},
  year         = {2020},
  url          = {https://doi.org/10.18653/v1/2020.emnlp-main.232},
  doi          = {10.18653/V1/2020.EMNLP-MAIN.232},
  bibsource    = {dblp computer science bibliography, https://dblp.org}
}

@inproceedings{dunefsky2024transcoders,
  author       = {Jacob Dunefsky and
                  Philippe Chlenski and
                  Neel Nanda},
  editor       = {Amir Globersons and
                  Lester Mackey and
                  Danielle Belgrave and
                  Angela Fan and
                  Ulrich Paquet and
                  Jakub M. Tomczak and
                  Cheng Zhang},
  title        = {Transcoders find interpretable {LLM} feature circuits},
  booktitle    = {Advances in Neural Information Processing Systems 38: Annual Conference
                  on Neural Information Processing Systems 2024, NeurIPS 2024, Vancouver,
                  BC, Canada, December 10 - 15, 2024},
  year         = {2024},
  url          = {http://papers.nips.cc/paper\_files/paper/2024/hash/2b8f4db0464cc5b6e9d5e6bea4b9f308-Abstract-Conference.html},
  bibsource    = {dblp computer science bibliography, https://dblp.org}
}

@misc{minder2025crosscoders,
      title={Overcoming Sparsity Artifacts in Crosscoders to Interpret Chat-Tuning}, 
      author={Julian Minder and Clément Dumas and Caden Juang and Bilal Chugtai and Neel Nanda},
      year={2025},
      eprint={2504.02922},
      archivePrefix={arXiv},
      primaryClass={cs.LG},
      url={https://arxiv.org/abs/2504.02922}, 
}

@article{mishra2024crosscoder,
  title   = {Sparse crosscoders for cross-layer features and model diffing},
  author  = {Mishra-Sharma, Siddharth and Bricken, Trenton and Lindsey, Jack and Jermyn, Adam and Marcus, Jonathan and Rivoire, Kelley and Olah, Christopher and Henighan, Thomas},
  journal = {Transformer Circuits Thread},
  note    = {https://transformer-circuits.pub/2025/crosscoder-diffing-update/index.html},
  year    = {2024}
}

@inproceedings{
    wang2025universality,
    title={Towards Universality: Studying Mechanistic Similarity Across Language Model Architectures},
    author={Junxuan Wang and Xuyang Ge and Wentao Shu and Qiong Tang and Yunhua Zhou and Zhengfu He and Xipeng Qiu},
    booktitle={The Thirteenth International Conference on Learning Representations},
    year={2025},
    url={https://openreview.net/forum?id=2J18i8T0oI}
}

@article{he2024llamascope,
  author       = {Zhengfu He and
                  Wentao Shu and
                  Xuyang Ge and
                  Lingjie Chen and
                  Junxuan Wang and
                  Yunhua Zhou and
                  Frances Liu and
                  Qipeng Guo and
                  Xuanjing Huang and
                  Zuxuan Wu and
                  Yu{-}Gang Jiang and
                  Xipeng Qiu},
  title        = {Llama Scope: Extracting Millions of Features from Llama-3.1-8B with
                  Sparse Autoencoders},
  journal      = {CoRR},
  volume       = {abs/2410.20526},
  year         = {2024},
  url          = {https://doi.org/10.48550/arXiv.2410.20526},
  doi          = {10.48550/ARXIV.2410.20526},
  eprinttype    = {arXiv},
  eprint       = {2410.20526},
  bibsource    = {dblp computer science bibliography, https://dblp.org}
}

@article{shoeybi2019megatron,
  author       = {Mohammad Shoeybi and
                  Mostofa Patwary and
                  Raul Puri and
                  Patrick LeGresley and
                  Jared Casper and
                  Bryan Catanzaro},
  title        = {Megatron-LM: Training Multi-Billion Parameter Language Models Using
                  Model Parallelism},
  journal      = {CoRR},
  volume       = {abs/1909.08053},
  year         = {2019},
  url          = {http://arxiv.org/abs/1909.08053},
  eprinttype    = {arXiv},
  eprint       = {1909.08053},
  bibsource    = {dblp computer science bibliography, https://dblp.org}
}

@book{sanders2019parallel,
author = {Sanders, Peter and Mehlhorn, Kurt and Dietzfelbinger, Martin and Dementiev, Roman},
title = {Sequential and Parallel Algorithms and Data Structures: The Basic Toolbox},
year = {2019},
isbn = {3030252086},
publisher = {Springer Publishing Company, Incorporated},
edition = {1st}
}

@misc{kingma2017adam,
      title={Adam: A Method for Stochastic Optimization}, 
      author={Diederik P. Kingma and Jimmy Ba},
      year={2017},
      eprint={1412.6980},
      archivePrefix={arXiv},
      primaryClass={cs.LG},
      url={https://arxiv.org/abs/1412.6980}, 
}

@article{cunningham2024complexity,
   title={Circuits Updates - June 2024},
   author={Hoagy Cunningham and Tom Conerly},
   year={2024},
   journal={Transformer Circuits Thread},
   url={https://transformer-circuits.pub/2024/june-update/index.html#hurdles}
}

@inproceedings{xu2019frequency,
  author       = {Zhi{-}Qin John Xu and
                  Yaoyu Zhang and
                  Yanyang Xiao},
  editor       = {Tom Gedeon and
                  Kok Wai Wong and
                  Minho Lee},
  title        = {Training Behavior of Deep Neural Network in Frequency Domain},
  booktitle    = {Neural Information Processing - 26th International Conference, {ICONIP}
                  2019, Sydney, NSW, Australia, December 12-15, 2019, Proceedings, Part
                  {I}},
  series       = {Lecture Notes in Computer Science},
  volume       = {11953},
  pages        = {264--274},
  publisher    = {Springer},
  year         = {2019},
  url          = {https://doi.org/10.1007/978-3-030-36708-4\_22},
  doi          = {10.1007/978-3-030-36708-4\_22},
  bibsource    = {dblp computer science bibliography, https://dblp.org}
}

@inproceedings{choshen2022grammar,
  author       = {Leshem Choshen and
                  Guy Hacohen and
                  Daphna Weinshall and
                  Omri Abend},
  editor       = {Smaranda Muresan and
                  Preslav Nakov and
                  Aline Villavicencio},
  title        = {The Grammar-Learning Trajectories of Neural Language Models},
  booktitle    = {Proceedings of the 60th Annual Meeting of the Association for Computational
                  Linguistics (Volume 1: Long Papers), {ACL} 2022, Dublin, Ireland,
                  May 22-27, 2022},
  pages        = {8281--8297},
  publisher    = {Association for Computational Linguistics},
  year         = {2022},
  url          = {https://doi.org/10.18653/v1/2022.acl-long.568},
  doi          = {10.18653/V1/2022.ACL-LONG.568},
  bibsource    = {dblp computer science bibliography, https://dblp.org}
}

@inproceedings{belrose2024statistics,
  author       = {Nora Belrose and
                  Quintin Pope and
                  Lucia Quirke and
                  Alex Mallen and
                  Xiaoli Z. Fern},
  title        = {Neural Networks Learn Statistics of Increasing Complexity},
  booktitle    = {Forty-first International Conference on Machine Learning, {ICML} 2024,
                  Vienna, Austria, July 21-27, 2024},
  publisher    = {OpenReview.net},
  year         = {2024},
  url          = {https://openreview.net/forum?id=IGdpKP0N6w},
  bibsource    = {dblp computer science bibliography, https://dblp.org}
}

@article{takahashi2017statistical,
  author       = {Shuntaro Takahashi and
                  Kumiko Tanaka{-}Ishii},
  title        = {Do Neural Nets Learn Statistical Laws behind Natural Language?},
  journal      = {CoRR},
  volume       = {abs/1707.04848},
  year         = {2017},
  url          = {http://arxiv.org/abs/1707.04848},
  eprinttype    = {arXiv},
  eprint       = {1707.04848},
  bibsource    = {dblp computer science bibliography, https://dblp.org}
}

@inproceedings{
    hanna2024faithfulness,
    title={Have Faith in Faithfulness: Going Beyond Circuit Overlap When Finding Model Mechanisms},
    author={Michael Hanna and Sandro Pezzelle and Yonatan Belinkov},
    booktitle={First Conference on Language Modeling},
    year={2024},
    url={https://openreview.net/forum?id=TZ0CCGDcuT}
}

@inproceedings{sundararajan2017attribution,
  author       = {Mukund Sundararajan and
                  Ankur Taly and
                  Qiqi Yan},
  editor       = {Doina Precup and
                  Yee Whye Teh},
  title        = {Axiomatic Attribution for Deep Networks},
  booktitle    = {Proceedings of the 34th International Conference on Machine Learning,
                  {ICML} 2017, Sydney, NSW, Australia, 6-11 August 2017},
  series       = {Proceedings of Machine Learning Research},
  volume       = {70},
  pages        = {3319--3328},
  publisher    = {{PMLR}},
  year         = {2017},
  url          = {http://proceedings.mlr.press/v70/sundararajan17a.html},
  bibsource    = {dblp computer science bibliography, https://dblp.org}
}

@inproceedings{zhao2021fewshot,
  author       = {Zihao Zhao and
                  Eric Wallace and
                  Shi Feng and
                  Dan Klein and
                  Sameer Singh},
  editor       = {Marina Meila and
                  Tong Zhang},
  title        = {Calibrate Before Use: Improving Few-shot Performance of Language Models},
  booktitle    = {Proceedings of the 38th International Conference on Machine Learning,
                  {ICML} 2021, 18-24 July 2021, Virtual Event},
  series       = {Proceedings of Machine Learning Research},
  volume       = {139},
  pages        = {12697--12706},
  publisher    = {{PMLR}},
  year         = {2021},
  url          = {http://proceedings.mlr.press/v139/zhao21c.html},
  bibsource    = {dblp computer science bibliography, https://dblp.org}
}

@inproceedings{gao2021fewshot,
  author       = {Tianyu Gao and
                  Adam Fisch and
                  Danqi Chen},
  editor       = {Chengqing Zong and
                  Fei Xia and
                  Wenjie Li and
                  Roberto Navigli},
  title        = {Making Pre-trained Language Models Better Few-shot Learners},
  booktitle    = {Proceedings of the 59th Annual Meeting of the Association for Computational
                  Linguistics and the 11th International Joint Conference on Natural
                  Language Processing, {ACL/IJCNLP} 2021, (Volume 1: Long Papers), Virtual
                  Event, August 1-6, 2021},
  pages        = {3816--3830},
  publisher    = {Association for Computational Linguistics},
  year         = {2021},
  url          = {https://doi.org/10.18653/v1/2021.acl-long.295},
  doi          = {10.18653/V1/2021.ACL-LONG.295},
  bibsource    = {dblp computer science bibliography, https://dblp.org}
}

@article{brown2020fewshot,
  author       = {Tom B. Brown and
                  Benjamin Mann and
                  Nick Ryder and
                  Melanie Subbiah and
                  Jared Kaplan and
                  Prafulla Dhariwal and
                  Arvind Neelakantan and
                  Pranav Shyam and
                  Girish Sastry and
                  Amanda Askell and
                  Sandhini Agarwal and
                  Ariel Herbert{-}Voss and
                  Gretchen Krueger and
                  Tom Henighan and
                  Rewon Child and
                  Aditya Ramesh and
                  Daniel M. Ziegler and
                  Jeffrey Wu and
                  Clemens Winter and
                  Christopher Hesse and
                  Mark Chen and
                  Eric Sigler and
                  Mateusz Litwin and
                  Scott Gray and
                  Benjamin Chess and
                  Jack Clark and
                  Christopher Berner and
                  Sam McCandlish and
                  Alec Radford and
                  Ilya Sutskever and
                  Dario Amodei},
  title        = {Language Models are Few-Shot Learners},
  journal      = {CoRR},
  volume       = {abs/2005.14165},
  year         = {2020},
  url          = {https://arxiv.org/abs/2005.14165},
  eprinttype    = {arXiv},
  eprint       = {2005.14165},
  bibsource    = {dblp computer science bibliography, https://dblp.org}
}

@misc{vanderwal2025polypythias,
      title={PolyPythias: Stability and Outliers across Fifty Language Model Pre-Training Runs}, 
      author={Oskar van der Wal and Pietro Lesci and Max Muller-Eberstein and Naomi Saphra and Hailey Schoelkopf and Willem Zuidema and Stella Biderman},
      year={2025},
      eprint={2503.09543},
      archivePrefix={arXiv},
      primaryClass={cs.CL},
      url={https://arxiv.org/abs/2503.09543}, 
}

@misc{penedo2024fineweb,
      title={The FineWeb Datasets: Decanting the Web for the Finest Text Data at Scale}, 
      author={Guilherme Penedo and Hynek Kydlíček and Loubna Ben allal and Anton Lozhkov and Margaret Mitchell and Colin Raffel and Leandro Von Werra and Thomas Wolf},
      year={2024},
      eprint={2406.17557},
      archivePrefix={arXiv},
      primaryClass={cs.CL},
      url={https://arxiv.org/abs/2406.17557}, 
}

@inproceedings{svete2024ngram,
  author       = {Anej Svete and
                  Ryan Cotterell},
  editor       = {Kevin Duh and
                  Helena G{\'{o}}mez{-}Adorno and
                  Steven Bethard},
  title        = {Transformers Can Represent n-gram Language Models},
  booktitle    = {Proceedings of the 2024 Conference of the North American Chapter of
                  the Association for Computational Linguistics: Human Language Technologies
                  (Volume 1: Long Papers), {NAACL} 2024, Mexico City, Mexico, June 16-21,
                  2024},
  pages        = {6845--6881},
  publisher    = {Association for Computational Linguistics},
  year         = {2024},
  url          = {https://doi.org/10.18653/v1/2024.naacl-long.381},
  doi          = {10.18653/V1/2024.NAACL-LONG.381},
  bibsource    = {dblp computer science bibliography, https://dblp.org}
}

@inproceedings{nguyen2024ngram,
  author       = {Timothy Nguyen},
  editor       = {Amir Globersons and
                  Lester Mackey and
                  Danielle Belgrave and
                  Angela Fan and
                  Ulrich Paquet and
                  Jakub M. Tomczak and
                  Cheng Zhang},
  title        = {Understanding Transformers via N-Gram Statistics},
  booktitle    = {Advances in Neural Information Processing Systems 38: Annual Conference
                  on Neural Information Processing Systems 2024, NeurIPS 2024, Vancouver,
                  BC, Canada, December 10 - 15, 2024},
  year         = {2024},
  url          = {http://papers.nips.cc/paper\_files/paper/2024/hash/b1c446eebd9a317dd0e96b16908c821a-Abstract-Conference.html},
  bibsource    = {dblp computer science bibliography, https://dblp.org}
}

@article{chen2024jetexpansionsngram,
  author       = {Yihong Chen and
                  Xiangxiang Xu and
                  Yao Lu and
                  Pontus Stenetorp and
                  Luca Franceschi},
  title        = {Jet Expansions of Residual Computation},
  journal      = {CoRR},
  volume       = {abs/2410.06024},
  year         = {2024},
  url          = {https://doi.org/10.48550/arXiv.2410.06024},
  doi          = {10.48550/ARXIV.2410.06024},
  eprinttype    = {arXiv},
  eprint       = {2410.06024},
  bibsource    = {dblp computer science bibliography, https://dblp.org}
}

@article{bigscience2024bloom,
  author       = {Teven Le Scao and
                  Angela Fan and
                  Christopher Akiki and
                  Ellie Pavlick and
                  Suzana Ilic and
                  Daniel Hesslow and
                  Roman Castagn{\'{e}} and
                  Alexandra Sasha Luccioni and
                  Fran{\c{c}}ois Yvon and
                  Matthias Gall{\'{e}} and
                  Jonathan Tow and
                  Alexander M. Rush and
                  Stella Biderman and
                  Albert Webson and
                  Pawan Sasanka Ammanamanchi and
                  Thomas Wang and
                  Beno{\^{\i}}t Sagot and
                  Niklas Muennighoff and
                  Albert Villanova del Moral and
                  Olatunji Ruwase and
                  Rachel Bawden and
                  Stas Bekman and
                  Angelina McMillan{-}Major and
                  Iz Beltagy and
                  Huu Nguyen and
                  Lucile Saulnier and
                  Samson Tan and
                  Pedro Ortiz Suarez and
                  Victor Sanh and
                  Hugo Lauren{\c{c}}on and
                  Yacine Jernite and
                  Julien Launay and
                  Margaret Mitchell and
                  Colin Raffel and
                  Aaron Gokaslan and
                  Adi Simhi and
                  Aitor Soroa and
                  Alham Fikri Aji and
                  Amit Alfassy and
                  Anna Rogers and
                  Ariel Kreisberg Nitzav and
                  Canwen Xu and
                  Chenghao Mou and
                  Chris Emezue and
                  Christopher Klamm and
                  Colin Leong and
                  Daniel van Strien and
                  David Ifeoluwa Adelani and
                  et al.},
  title        = {{BLOOM:} {A} 176B-Parameter Open-Access Multilingual Language Model},
  journal      = {CoRR},
  volume       = {abs/2211.05100},
  year         = {2022},
  url          = {https://doi.org/10.48550/arXiv.2211.05100},
  doi          = {10.48550/ARXIV.2211.05100},
  eprinttype    = {arXiv},
  eprint       = {2211.05100},
  bibsource    = {dblp computer science bibliography, https://dblp.org}
}

@misc{olmo20242olmo2furious,
      title={2 OLMo 2 Furious}, 
      author={Team OLMo and Pete Walsh and Luca Soldaini and Dirk Groeneveld and Kyle Lo and Shane Arora and Akshita Bhagia and Yuling Gu and Shengyi Huang and Matt Jordan and Nathan Lambert and Dustin Schwenk and Oyvind Tafjord and Taira Anderson and David Atkinson and Faeze Brahman and Christopher Clark and Pradeep Dasigi and Nouha Dziri and Michal Guerquin and Hamish Ivison and Pang Wei Koh and Jiacheng Liu and Saumya Malik and William Merrill and Lester James V. Miranda and Jacob Morrison and Tyler Murray and Crystal Nam and Valentina Pyatkin and Aman Rangapur and Michael Schmitz and Sam Skjonsberg and David Wadden and Christopher Wilhelm and Michael Wilson and Luke Zettlemoyer and Ali Farhadi and Noah A. Smith and Hannaneh Hajishirzi},
      year={2024},
      eprint={2501.00656},
      archivePrefix={arXiv},
      primaryClass={cs.CL},
      url={https://arxiv.org/abs/2501.00656}, 
}
\bibliographystyle{iclr2025_conference}

\newpage

\appendix

\section{Crosscoder Training Details}
\label{appendix:crosscoder-detail}

Section~\ref{sec:crosscoders} presents the mathematical definition of our cross-snapshot crosscoder. However, in practical applications, additional architectural design and training techniques are required to advance the Pareto frontier of sparsity versus reconstruction fidelity. In this section, we detail our selection of activation function $\sigma(\cdot)$ and regularization function $\Omega(\cdot)$, and present our training hyperparameters and their results. We also compare crosscoder performance to standard SAEs to evaluate how well crosscoders perform at sparse dictionary learning.

\subsection{Selection of Activation Function and Regularization Function}
\label{appendix:crosscoder-activation-regularization-selection}

The sparsity of natural features is the fundamental hypothesis underlying superposition~\citep{olshausen1997sparsecoding, elhage2022superposition, hubun2024sae}. To obtain crosscoder features with optimal sparsity for ideal interpretability and monosemanticity while maintaining reconstruction fidelity, we carefully select the activation function $\sigma(\cdot)$ and the regularization function $\Omega(\cdot)$. 

Previous SAE studies predominantly use ReLU activation with L1 regularization~\citep{bricken2023monosemanticity, hubun2024sae}. However, this configuration produces weak feature activations that are largely noise, compromising both interpretability and sparsity. We address this by adopting JumpReLU~\citep{rajamanoharan2024jumprelu} as the activation function, which eliminates activations below learned thresholds (trained via straight-through estimation~\citep{bengio2013ste}). 

To prevent features from becoming permanently inactive at certain snapshots, we incorporate decoder norms into the activation decision. Given pre-activation $z(x)$:

\begin{equation}
    z(x)=\sum_{\theta \in \Theta} W_\text{enc}^\theta a^\theta(x)+b_\text{enc}
\end{equation}

The $i$-th feature activation at snapshot $\theta$ is defined as:
\begin{equation}
    f^\theta_i(x)=z_i(x) \cdot H(z_i(x) \cdot \| W_{\text{dec}, i}^\theta \| - t_i)
\end{equation}
where $H(\cdot)$ is the Heaviside step function and $t_i\in\mathbb{R}$ is the JumpReLU threshold for feature $i$. 

This design ensures that features with small decoder norms require stronger pre-activations to activate, preventing complete feature death while maintaining sparsity. Although this means truly inactive features retain small positive decoder norms rather than zero values, this architectural choice significantly improves crosscoder performance and enables better feature tracking across snapshots.

For regularization, we employ a combination of tanh and quadratic frequency penalties~\citep{smith2025negativesae}: the tanh component provides a superior L0 approximation by reducing penalties on strong activations, while the quadratic frequency penalty suppresses high-frequency features. This yields the following batched sparsity loss:

\begin{equation}
    \begin{aligned}
        \omega_i^\theta(\mathcal{B})&=\frac{1}{\mathcal{B}}\sum_{x\in\mathcal{B}} \tanh\left( f_i(x) \cdot || W_{\text{dec}, i}^\theta || \right)\\
        \mathcal{L}_\text{sparsity}(\mathcal{B})&= \lambda_\text{sparsity}\sum_{\theta\in\Theta}\sum_{i=1}^{n_\text{features}} \omega_i^\theta (\mathcal{B})\cdot\left(1+\frac{\omega_i^\theta(\mathcal{B})}{\omega_0}\right)
    \end{aligned}
\end{equation}

where $\mathcal{B} \subset \mathcal{C} \times \mathbb{N}$ is an input batch, and $\omega_i^\theta(\mathcal{B})$ is the single-batch differentiable estimate for the activation frequency on snapshot $\theta$ of $i$-th crosscoder feature, using tanh as L0 approximation. A new hyperparameter $\omega_0$ is introduced to quadratically penalize feature activation when $\omega_i^\theta(\mathcal{B}) \gg \omega_0$.

\begin{figure*}[t]
    \centering
    \begin{subfigure}[b]{0.32\textwidth}
        \centering
        \includegraphics[width=\textwidth]{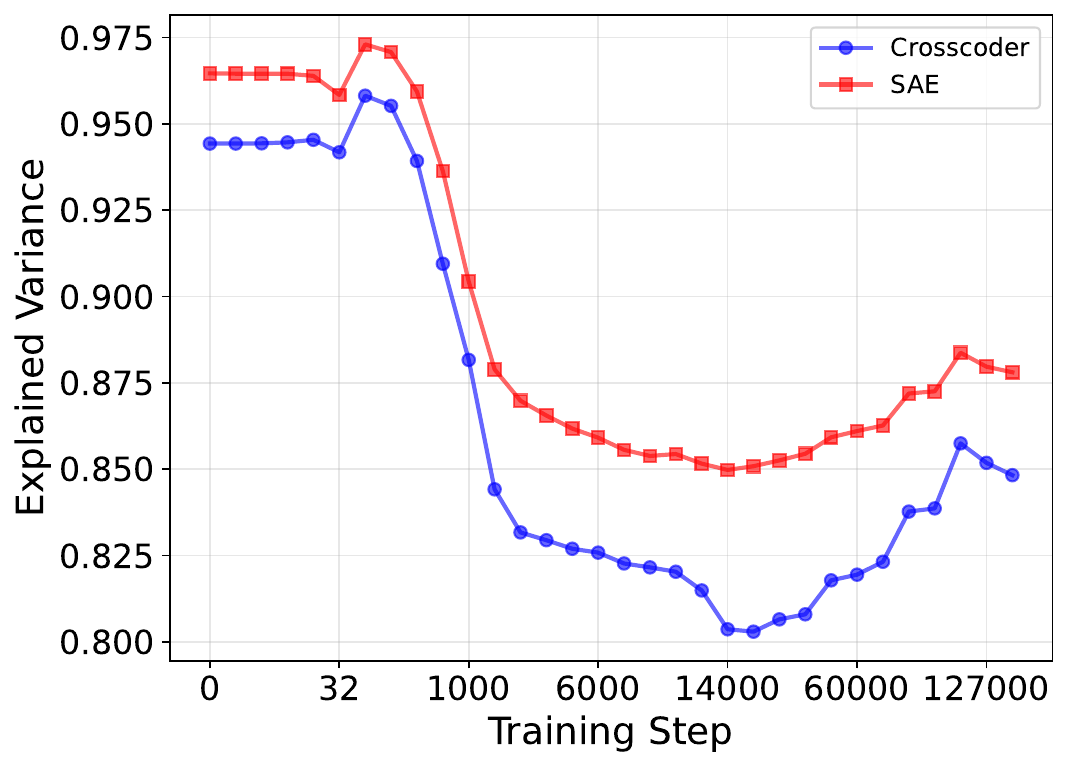}
        \caption{}
        \label{fig:crosscoder-ev-comparison}
    \end{subfigure}
    \hfill
    \begin{subfigure}[b]{0.32\textwidth}
        \centering
        \includegraphics[width=\textwidth]{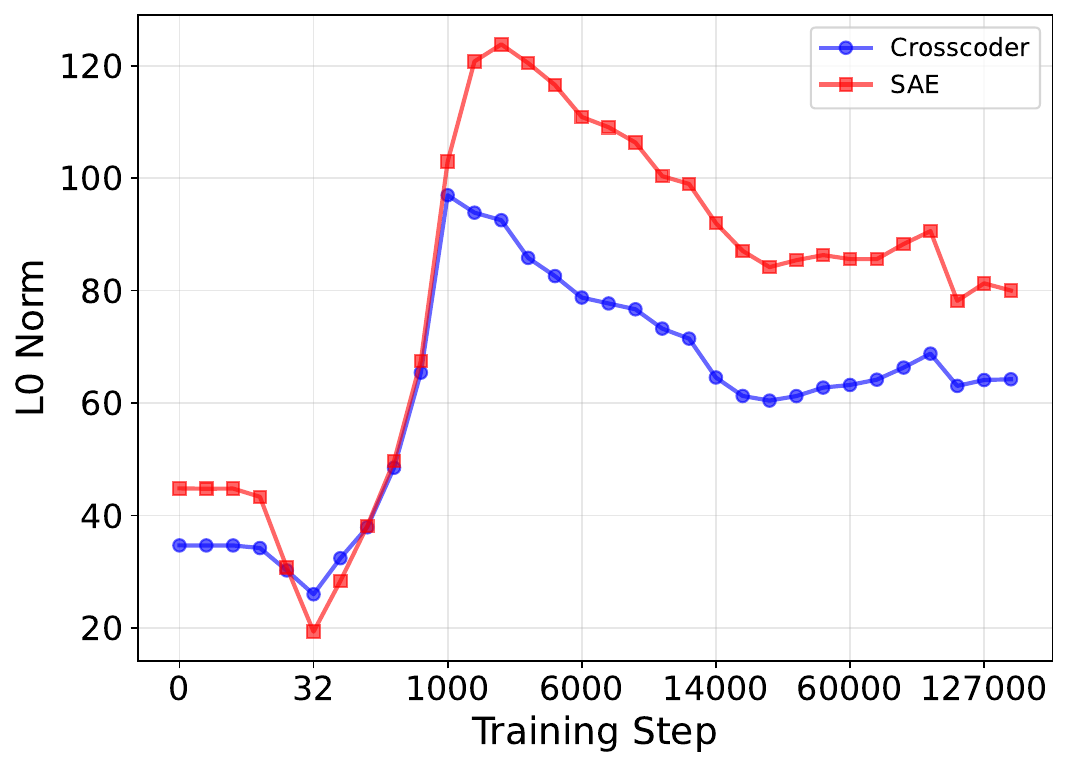}
        \caption{}
        \label{fig:crosscoder-l0-comparison}
    \end{subfigure}
    \hfill
    \begin{subfigure}[b]{0.32\textwidth}
        \centering
        \includegraphics[width=\textwidth]{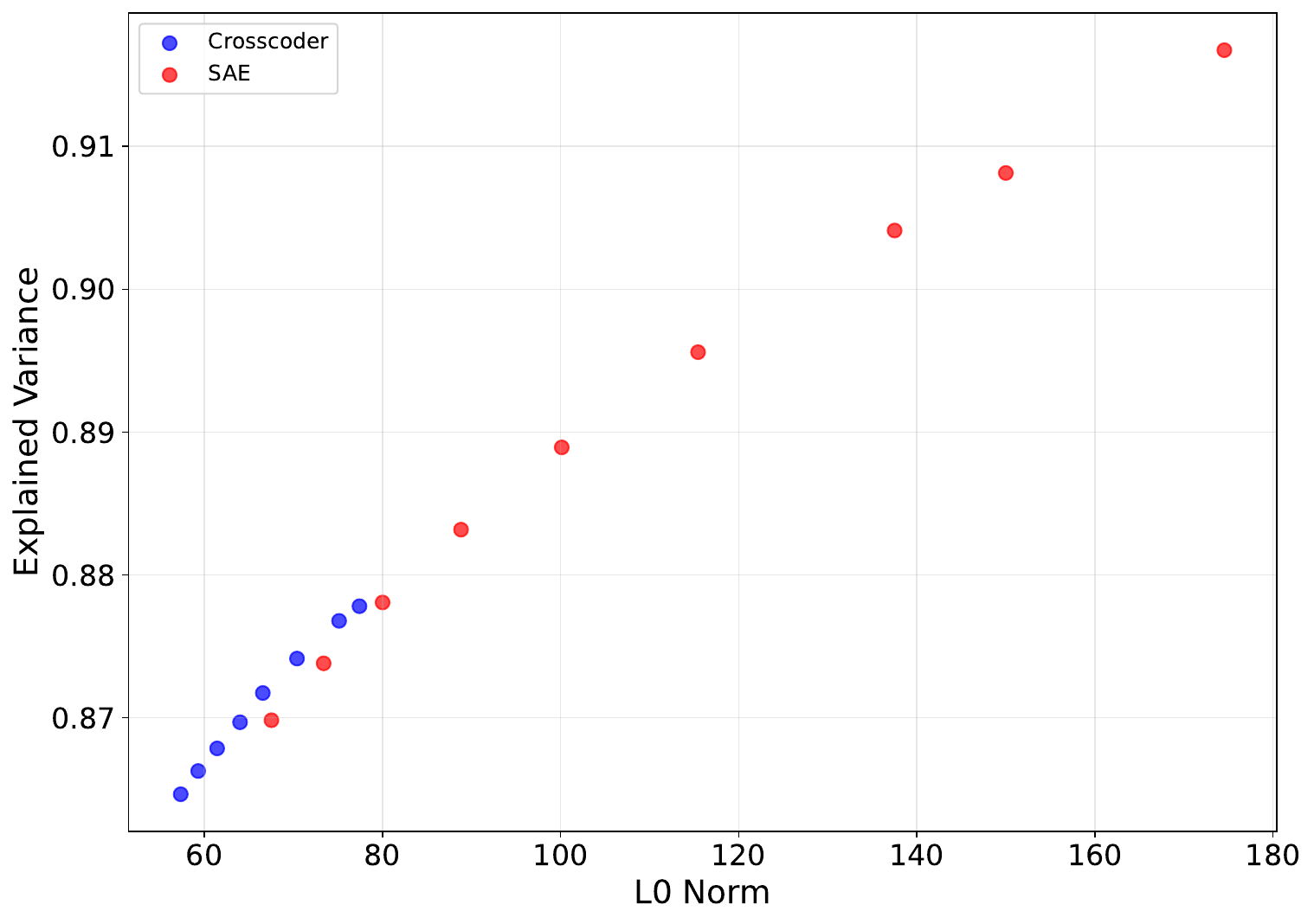}
        \caption{}
        \label{fig:crosscoder-pareto-comparison}
    \end{subfigure}
    \caption{Comparison between crosscoders and per-snapshot SAEs. (a) The explained variance of crosscoders versus SAEs at each snapshot. (b) The L0 norm of crosscoders versus SAEs at each snapshot. (c) The Pareto frontier comparison of crosscoders and SAEs trained on the final snapshot.}
\end{figure*}

\subsection{Selection of Pre-Training Snapshots}

\begin{figure*}[h]
    \centering
    \includegraphics[height=4cm]{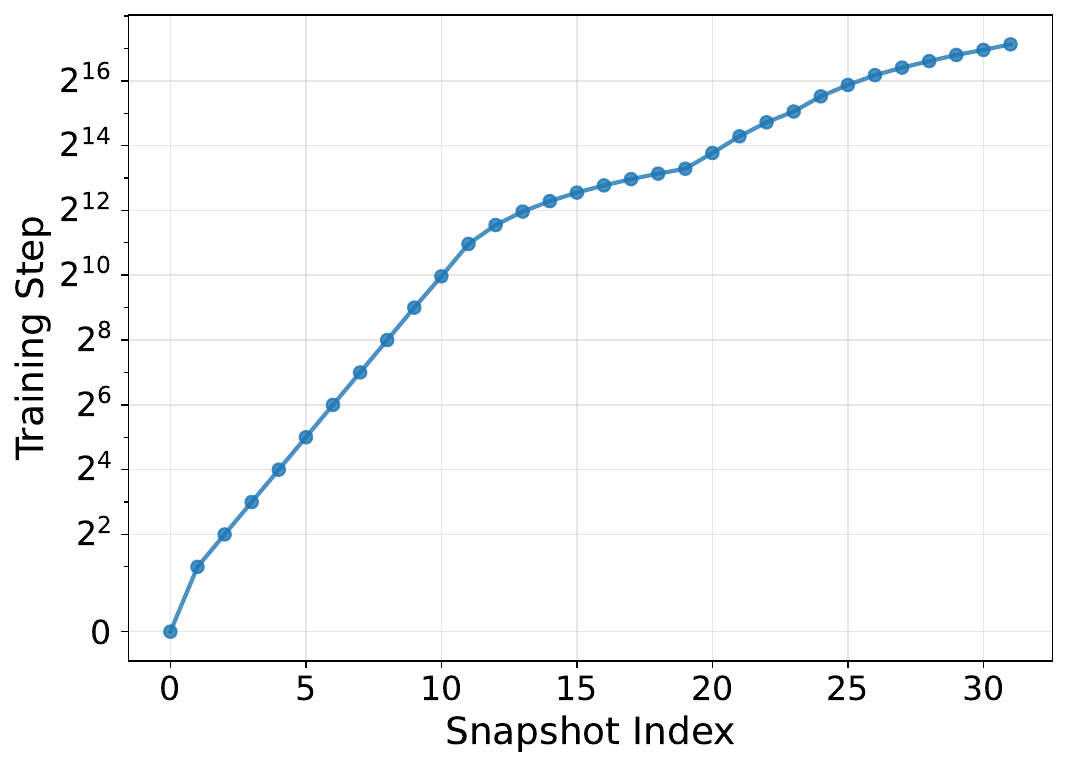}
    \caption{Selected snapshots from Pythia suite.}
    \label{fig:steps}
\end{figure*}

To balance training cost with the granularity of feature evolution analysis, we train crosscoders using $32$ source snapshots from the 154 open-source snapshots in the Pythia suite. Figure~\ref{fig:steps} shows the training steps of the selected snapshots.

\subsection{Distributed Training Strategy for Crosscoders}
Crosscoders require parameters that scale with the number of source snapshots $n_\text{snapshots}$, resulting in significantly higher memory and computational requirements compared to SAEs with equivalent feature counts, particularly when using many source snapshots.

To efficiently train crosscoders, we employ a \textbf{head parallelism} distributed training strategy, a variant of tensor parallelism~\citep{shoeybi2019megatron}. With $k$ processes where $k$ divides $n_\text{snapshots}$, each process handles encoding and decoding for $\frac{n_\text{snapshots}}{k}$ source snapshots. Pre-activations are computed via All-Reduce operations~\citep{sanders2019parallel}. Unlike standard tensor parallelism, our approach processes activations from each snapshot separately on individual processes, reducing I/O overhead for reading activations from disk.

\subsection{Experiments}
\label{appendix:experiments}

We train crosscoders on Pythia-160M and Pythia-6.9B snapshots at various scales (6,144 to 98,304 features on Pythia-160M, and 32,768 features on Pythia-6.9B). We primarily focus on middle layers (Layer 6 in Pythia-160M, and Layer 16 in Pythia-6.9B)\footnote{By ``Layer $i$", we refer to the activations at the output of the $i$-th transformer layer.}, but also train crosscoders of 24,576 features on all layers of Pythia-160M for comprehensive analysis.

All crosscoders are trained on 800M tokens from the SlimPajama corpus using the Adam optimizer~\citep{kingma2017adam} with $\beta$ values of $(0.9, 0.999)$. To prevent straight-through estimation from causing rapid threshold increases, we apply a reduced learning rate to JumpReLU threshold updates via a multiplier on the global learning rate. JumpReLU thresholds are initialized to $0.1$ for all features. The learning rate schedule includes 10\% warm-up steps followed by 20\% decay steps. We initialize encoders as transposes of their corresponding decoders, with identical initialization matrices across all snapshots. We employ initialization search to identify optimal decoder norms that minimize loss at initialization. Additional hyperparameters are listed in Table~\ref{tab:hyperparameters}.

\begin{table}[h]
\centering
\caption{Hyperparameters for crosscoder training}
\label{tab:hyperparameters}
\begin{tabular}{lcc}
\toprule
\textbf{Parameter} & \textbf{Pythia-160M} & \textbf{Pythia-6.9B} \\
\midrule
Learning Rate & 5e-5 & 1e-5 \\
Batch Size & 2048 & 2048 \\
Feature Expansion Ratio & 8×, 16×, 32×, 64×, 128× & 8× \\
Sparsity Coefficient ($\lambda_{\text{sparsity}}$) & 0.3 & 0.3 \\
JumpReLU Threshold LR Multiplier & 0.1 & 0.3 \\
\bottomrule
\end{tabular}
\end{table}

\subsection{Results on Other Layers}

\begin{figure*}[t]
    \centering
    \includegraphics[height=6cm]{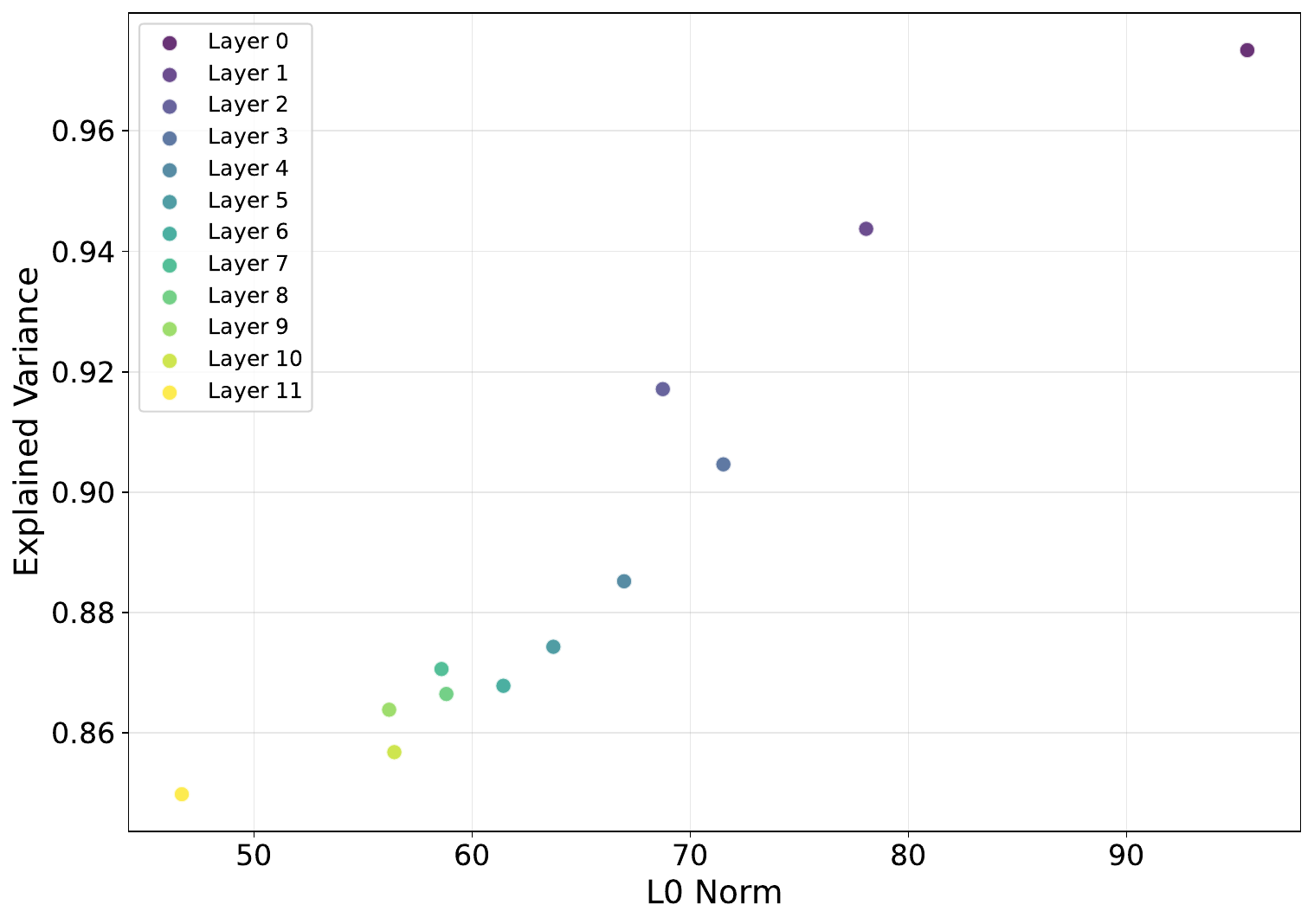}
    \caption{Explained variance versus L0 norm of crosscoders trained on all layers (0-11) of Pythia 160M}
    \label{fig:all-layer-ev-l0}
\end{figure*}

\begin{figure*}[p]
    \centering
    \includegraphics[width=\textwidth]{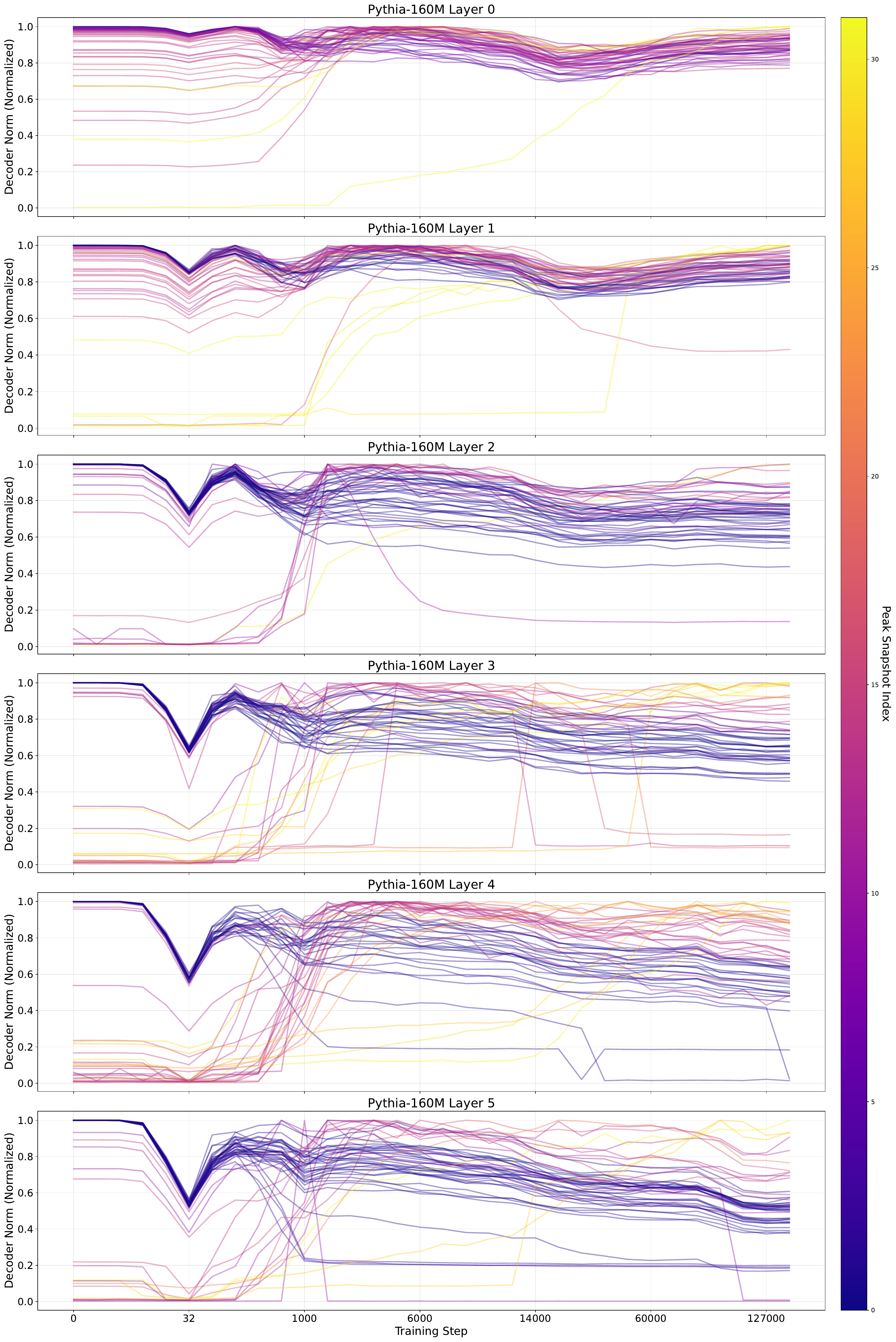}
    \caption{Feature decoder norm evolution of Layer 0 to Layer 5 in a 24,576-feature crosscoder trained on Pythia-160M.}
    \label{fig:all-layer-norm-evolution-1}
\end{figure*}

\begin{figure*}[p]
    \centering
    \includegraphics[width=\textwidth]{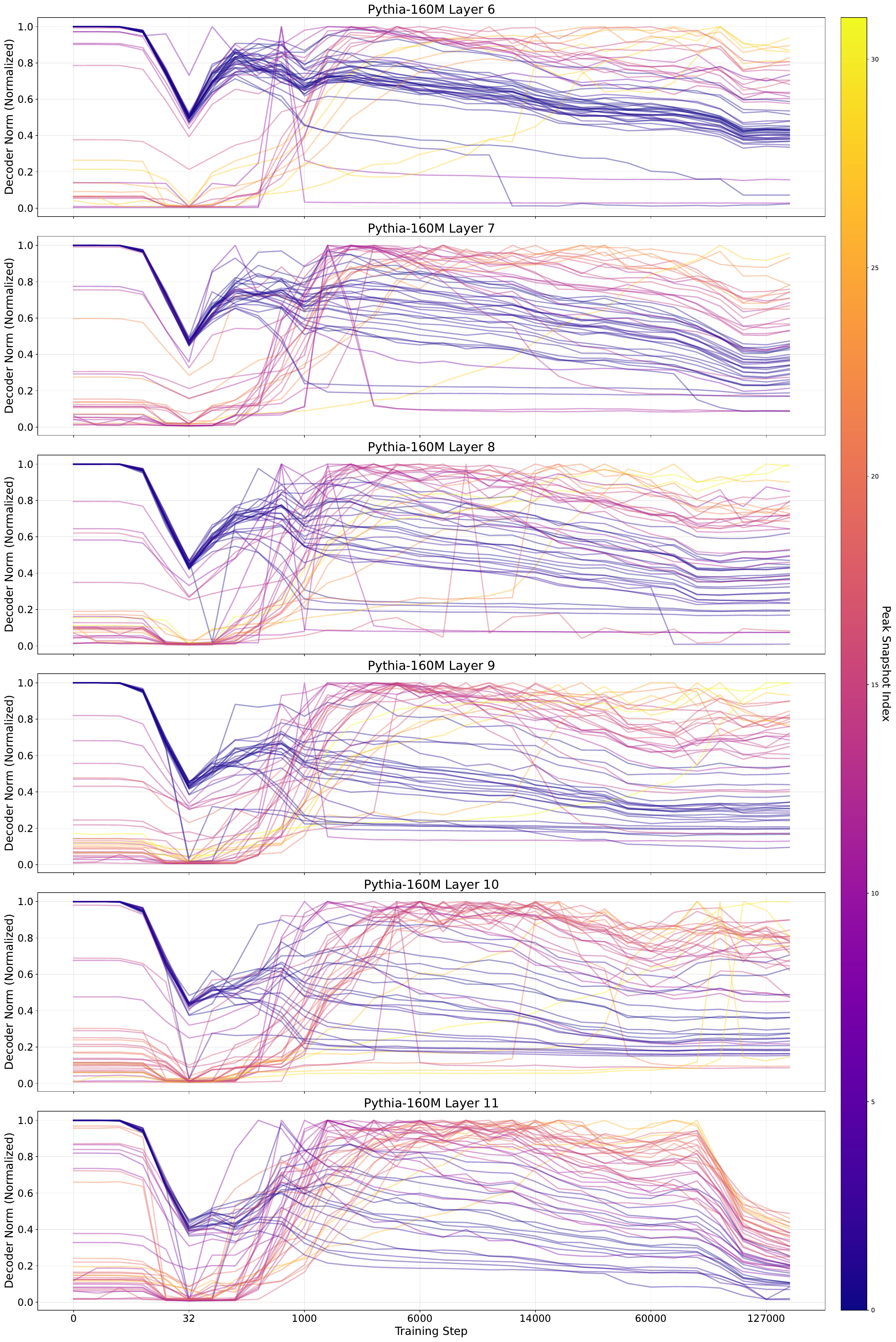}
    \caption{Feature decoder norm evolution of Layer 6 to Layer 11 in a 24,576-feature crosscoder trained on Pythia-160M.}
    \label{fig:all-layer-norm-evolution-2}
\end{figure*}

We train additional crosscoders with 24,576 features on all layers (0-11) of Pythia-160M using the same hyperparameters as Layer 6 (Table~\ref{tab:hyperparameters}). Figure~\ref{fig:all-layer-ev-l0} shows the explained variance and L0 norm for these crosscoders, which exhibit similar performance but different trade-offs between sparsity and reconstruction quality. 

To demonstrate feature evolution across all layers, we apply the same visualization strategy from Section~\ref{sec:assessing}, plotting decoder norms for 50 randomly selected features per layer (Figures~\ref{fig:all-layer-norm-evolution-1} and~\ref{fig:all-layer-norm-evolution-2}). Most middle layers exhibit the same evolutionary patterns as Layer 6, supporting our findings in Section~\ref{sec:assessing}. We would like to note that the majority of features in Layers 0 and 1 exist from initialization, which aligns with the observation that early layers implements more low-level, especially single-token features~\citep{he2024llamascope}.

\section{Automated Scoring of Complexity}
\label{appendix:complexity}

We leverage Claude Sonnet 4 for automated complexity score evaluation. For each feature, we select the top 10 activating samples with 100 surrounding tokens around the strongest activating tokens and apply the following prompt:

\footnotesize
\begin{verbatim}
# Neural Network Feature Analysis Instructions
We're analyzing features in a neural network. Each feature activates on 
specific words, substrings, or concepts within short documents. Activating 
words are marked as `<<text, {activation}>>` where `{activation}` indicates 
the strength of activation (higher values = stronger activation). You'll 
receive documents containing highest activations tokens and tokens 
surrounding them.

## Your task:
### 1. Summarize the Activation (<20 words)
Examine the marked activations and summarize what the feature detects in 
one sentence. 
- Avoid being overly specific—your explanation should cover most/all 
  activating words
- If all words in a sentence activate, focus on the sentence's concept 
  rather than individual words
- Note relevant patterns in capitalization, punctuation, or formatting
- Prioritize strongly activated tokens
- Keep explanations simple and concise
- Avoid long word lists

### 2. Assess Feature Complexity (1-5 scale, with decimal precision allowed)
Rate the feature's complexity:
- **5**: Rich feature with diverse contexts unified by an interesting theme
- **4**: High-level semantic structure with potentially dense activation
- **3**: Moderate complexity—phrases, categories, or sentence structures
- **2**: Synonyms or words at a same class
- **1**: Single specific word or token
You may use decimal values (e.g., 3.7) for more precise assessment.

### Output Format
Your output should be in JSON format, with two fields: summarization and 
complexity. You should directly output the JSON object, without any other 
text.
\end{verbatim}
\normalsize

\section{Assessing Crosscoder Decoder Norm}
\label{appendix:decoder-norm}

\begin{figure*}[h]
    \centering
    \includegraphics[height=8cm]{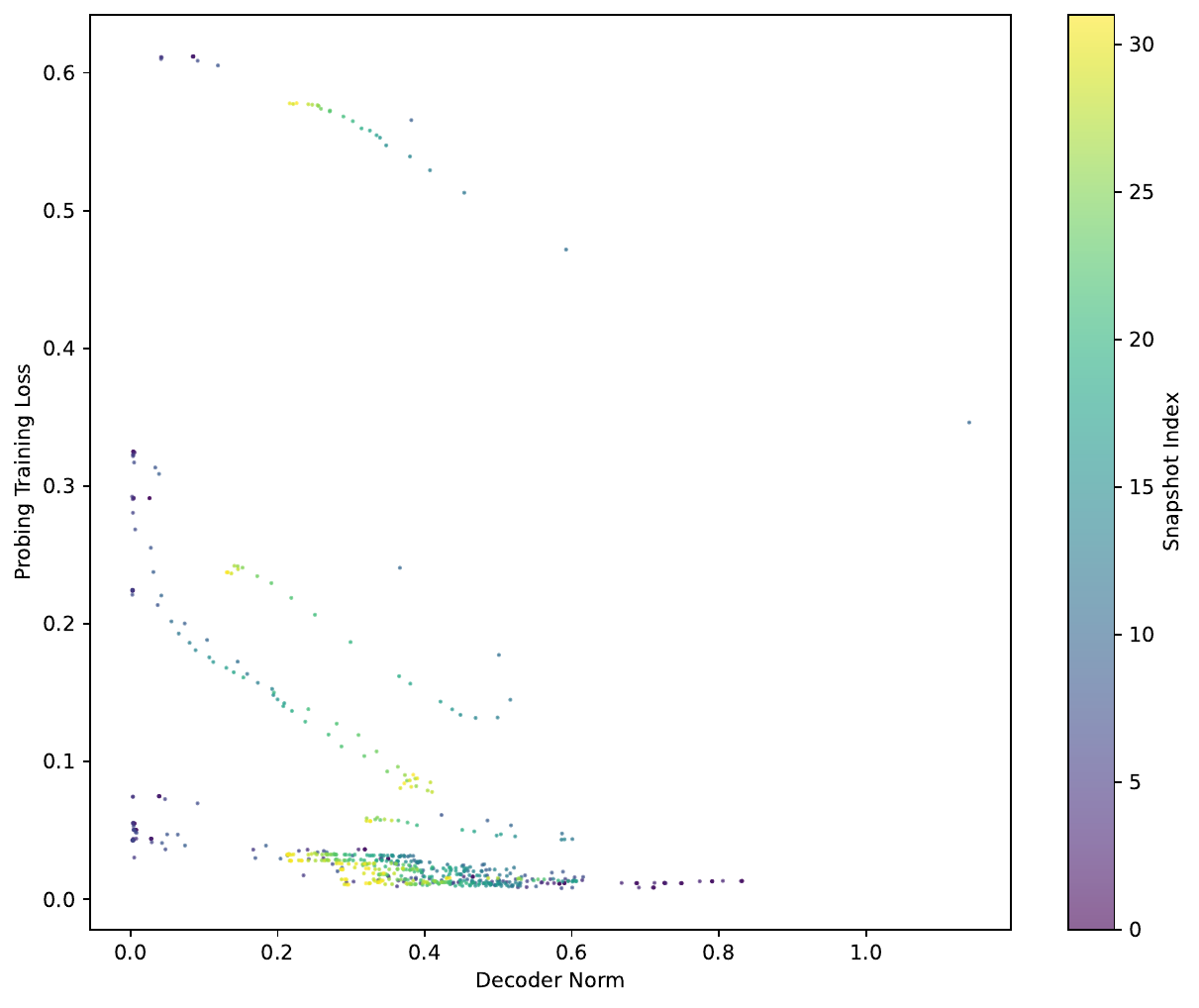}
    \caption{Linear probe training errors versus feature decoder norms of 20 randomly sampled features from a crosscoder with 6,144 features on Pythia-160M.}
    \label{fig:linear-probing}
\end{figure*}

Section~\ref{sec:assessing} takes advantage of the crosscoder decoder norms to study feature evolution. However, are the crosscoder decoder norms well quantitative indicators of the extent to which features evolved? Aside from theoretical statements, we conduct experiments to examine whether crosscoder decoder norms can reflect the intensity of features. We leverage linear probing to test the correlation between the feature decoder norm and the linear separability of the feature activations on each crosscoder feature.

We train linear probing classifiers (referred to as probes) separately on activations of each model snapshot to classify whether each crosscoder feature activates. For each feature $i$ and each snapshot $\theta$, our probes map model activations to the predicted feature activating probability $p_i^\theta(x)$ as:

\begin{equation}
    \begin{aligned}
        p_i^\theta(x)=\operatorname{sigmoid}(w_{\text{probe},i}^\theta \cdot a^\theta(x)+b_{\text{probe},i}^\theta)
    \end{aligned}
\end{equation}

where $w_{\text{probe},i}^\theta \in \mathbb{R}^{d_\text{model}}$ and $b_{\text{probe},i}^\theta\in\mathbb{R}$ is the weight and bias of the probe w.r.t. feature $i$ and snapshot $\theta$. Each probe is trained to minimize the binary cross-entropy loss $y_i \cdot \log p_i^\theta(x) + (1-y_i) \cdot \left(\log(1-p_i^\theta(x))\right)$, given the label $y_i=\operatorname{sgn}\left(f_i(x)\right)$. The training loss of each probe should be a direct measure of the linear separability of the feature activations.

We train probes for 100M tokens for each feature in a 6,144-feature crosscoder on Pythia-160M. The probe errors of each feature show a mean Pearson correlation with the corresponding decoder norms by $-0.867$, with a standard deviation of $0.153$. We demonstrate example probe errors versus feature decoder norms of 20 randomly sampled features in Figure~\ref{fig:linear-probing}. This indicates a strong negative linear relationship between probe errors and crosscoder decoder norms, demonstrating the effectiveness of the crosscoder decoder norms as indicators of feature evolution.

\section{Rules for Finding Typical Cross-Snapshot Features}
\label{appendix:evolution}

We define the rules used to identify previous token features, induction features, and context-sensitive features as follows:

\begin{enumerate}
    \item \textbf{Previous Token Features:} We collect the directly preceding tokens of all activating tokens in the top 20 activating samples and assess their consistency. Token consistency is defined as the proportion of the largest single group when all tokens are normalized by stemming (removing leading and trailing spaces and ignoring case). To exclude bigram or multigram features, we also evaluate the consistency of the activating tokens themselves. A feature is classified as a previous token feature if it exhibits high consistency in previous tokens (above 0.8) and low consistency in activating tokens (below 0.3).
    \item \textbf{Induction Features:} Induction features should activate on the second [A] in patterns [A][B]...[A][B]. For each activating token [A], we collect its following token [B] and search for previous occurrences of the bigram [A][B]. To distinguish induction features from simpler features that merely activate on any bigram [A][B], we require that the feature does not activate on the first appearance of [A][B]. A feature exhibiting this behavior in at least 20 instances within the top 20 activating samples is classified as an induction feature.
    \item \textbf{Context-sensitive Features:} We identify context-sensitive features using a simpler rule based on activation density within specific contexts. Context-sensitive features should activate frequently in highly specific contexts, so we require features to have high activation counts within the top 20 activating samples (exceeding 4,000 activations). To exclude features that activate ubiquitously (such as positional or bias features), we filter out features with excessive total activations (below 2M total activations across 100M analyzed tokens).
\end{enumerate}

\section{Details in Downstream Task Attribution}
\label{appendix:downstream-tasks}

\subsection{Formalization of IG Attribution Score}

\begin{figure*}[t]
   \centering
   \begin{subfigure}[b]{0.32\textwidth}
       \centering
       \includegraphics[width=\textwidth]{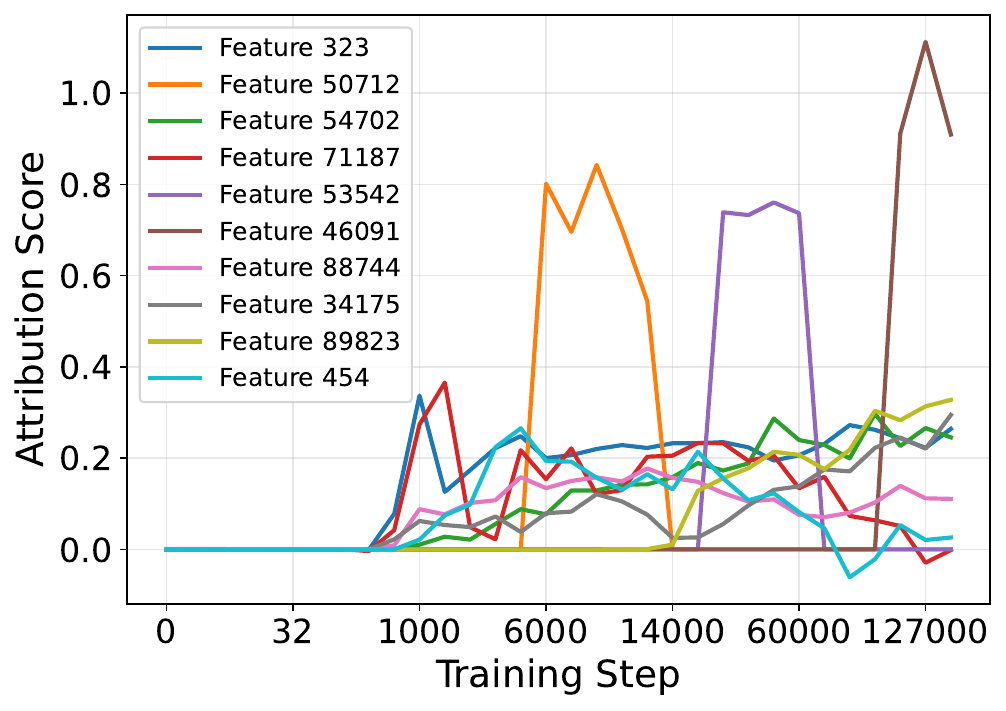}
       \caption{}
   \end{subfigure}
   \hfill
   \begin{subfigure}[b]{0.32\textwidth}
       \centering
       \includegraphics[width=\textwidth]{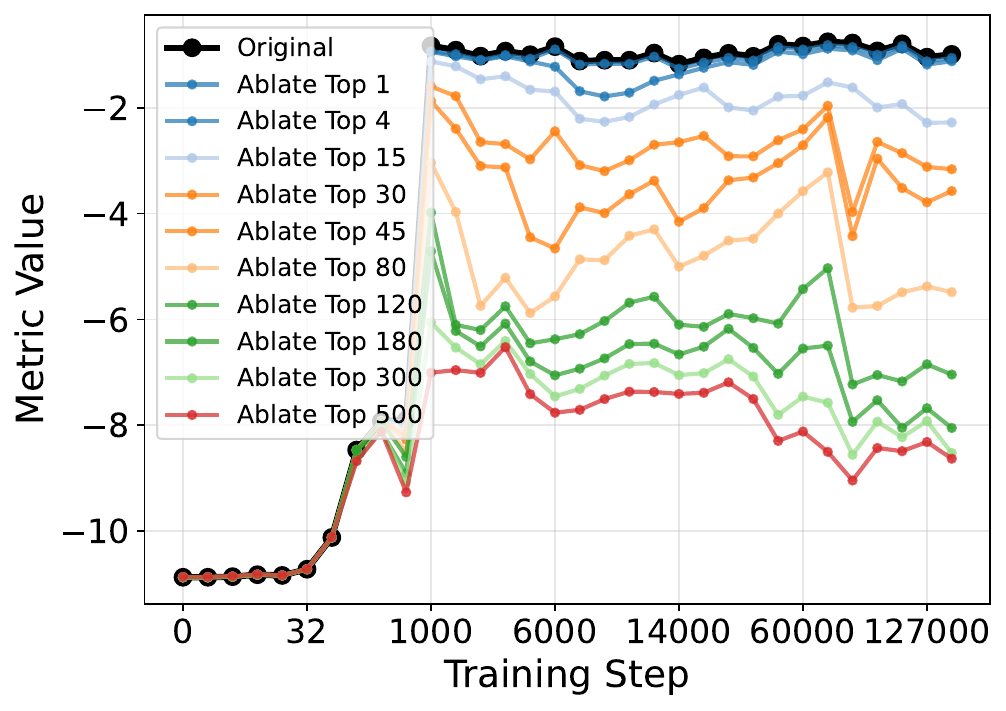}
       \caption{}
   \end{subfigure}
   \hfill
   \begin{subfigure}[b]{0.32\textwidth}
       \centering
       \includegraphics[width=\textwidth]{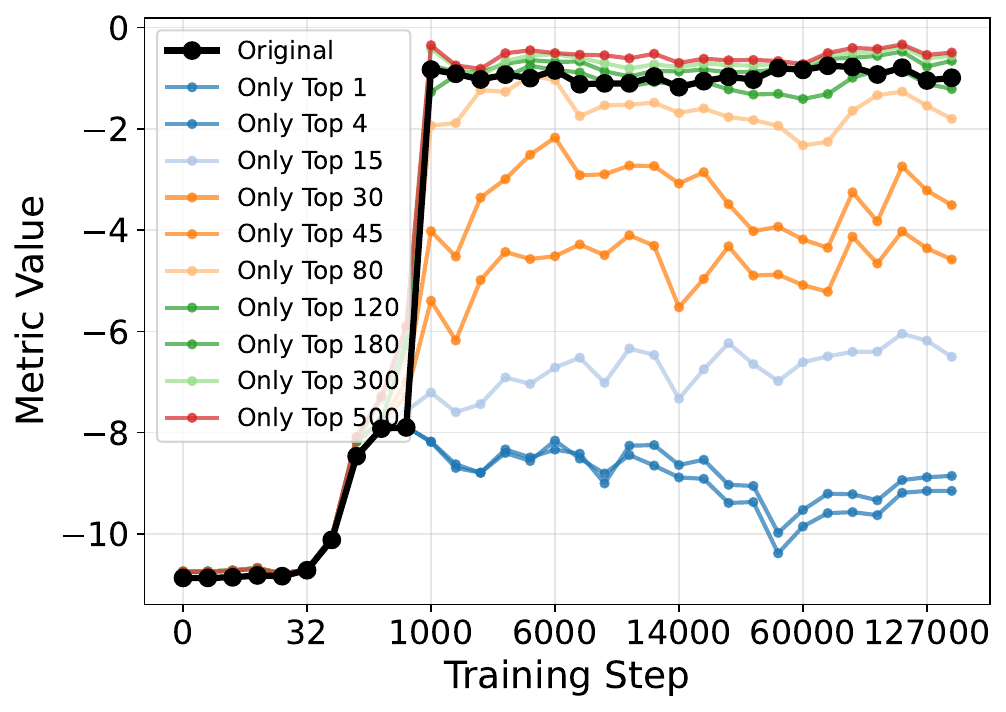}
       \caption{}
   \end{subfigure}

   \caption{Crosscoder feature attribution on the induction task, using a crosscoder with 98,304 features trained on Pythia-160M.}
   \label{fig:attribution-induction}
\end{figure*}

\begin{figure*}[t]
   \centering
   \begin{subfigure}[b]{0.32\textwidth}
       \centering
       \includegraphics[width=\textwidth]{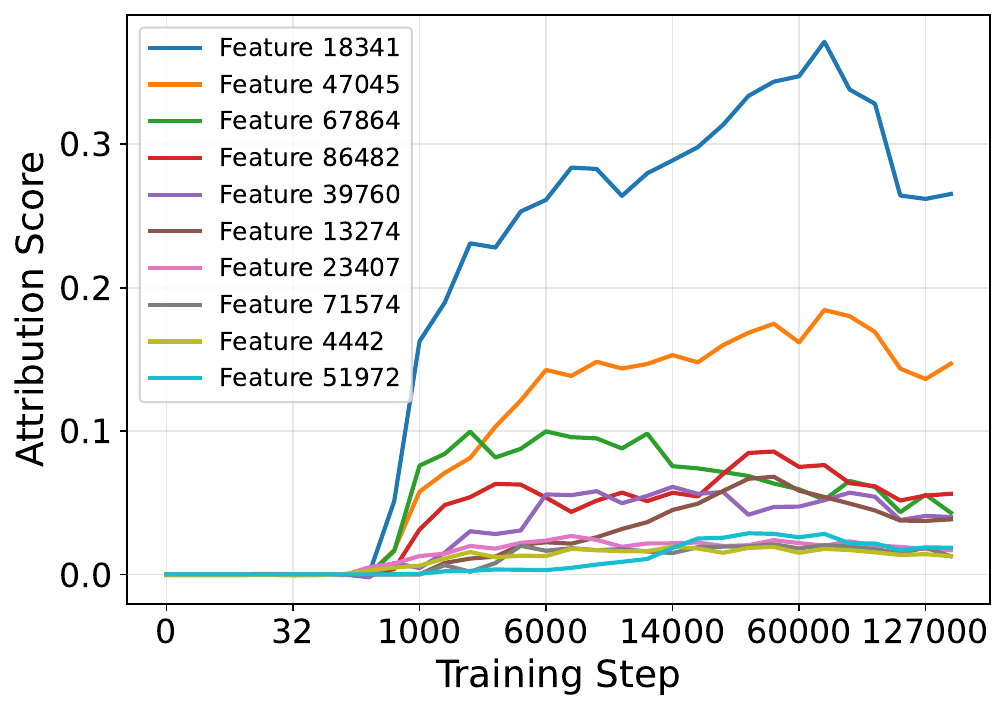}
       \caption{}
   \end{subfigure}
   \hfill
   \begin{subfigure}[b]{0.32\textwidth}
       \centering
       \includegraphics[width=\textwidth]{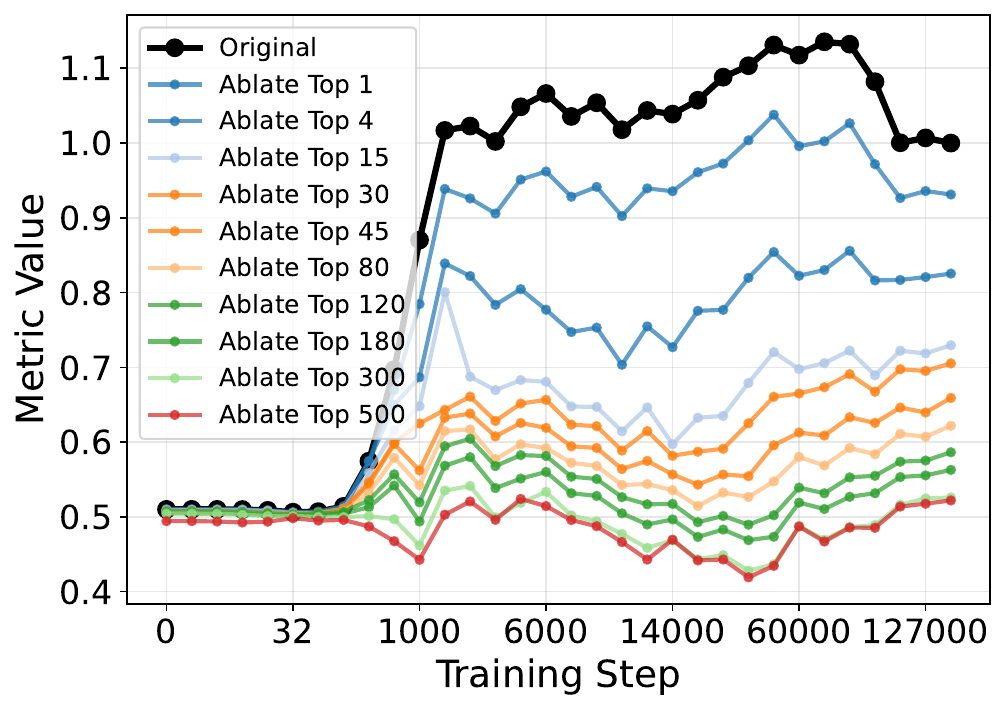}
       \caption{}
   \end{subfigure}
   \hfill
   \begin{subfigure}[b]{0.32\textwidth}
       \centering
       \includegraphics[width=\textwidth]{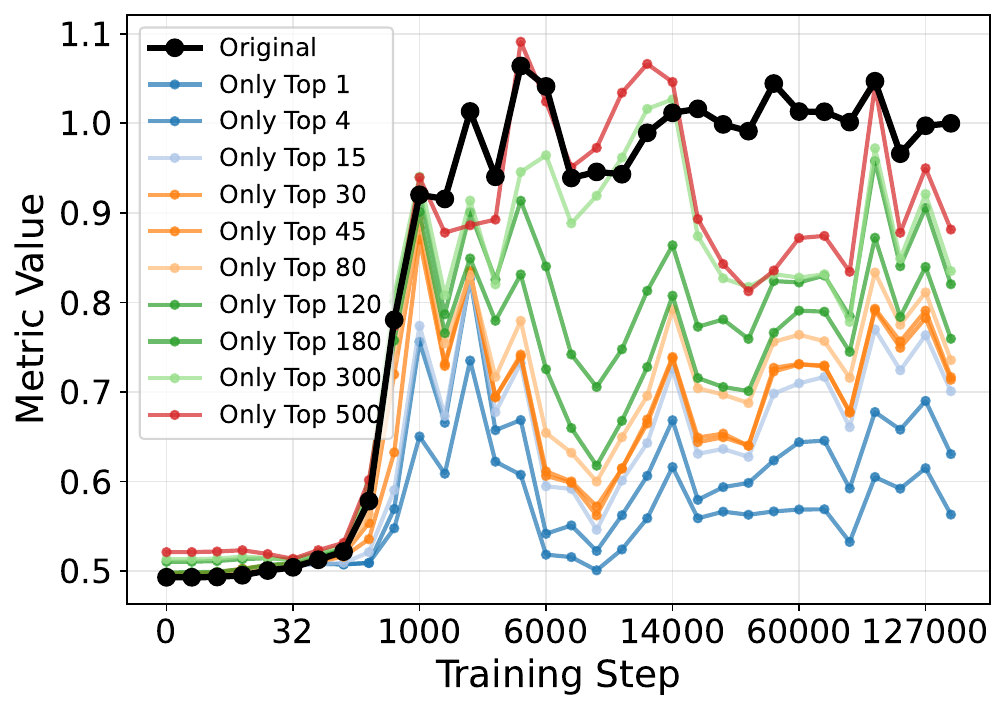}
       \caption{}
   \end{subfigure}

   \caption{Crosscoder feature attribution on the Simple variant of the SVA task, using a crosscoder with 98,304 features trained on Pythia-160M.}
   \label{fig:attribution-sva-simple}
\end{figure*}

\begin{figure*}[t]
   \centering
   \begin{subfigure}[b]{0.32\textwidth}
       \centering
       \includegraphics[width=\textwidth]{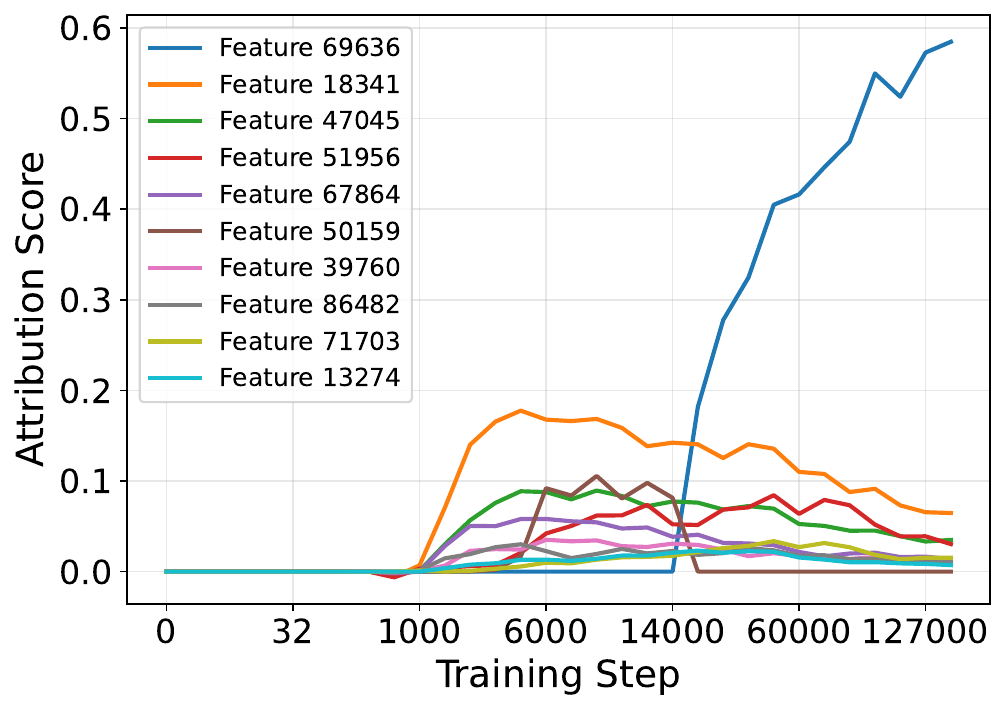}
       \caption{}
   \end{subfigure}
   \hfill
   \begin{subfigure}[b]{0.32\textwidth}
       \centering
       \includegraphics[width=\textwidth]{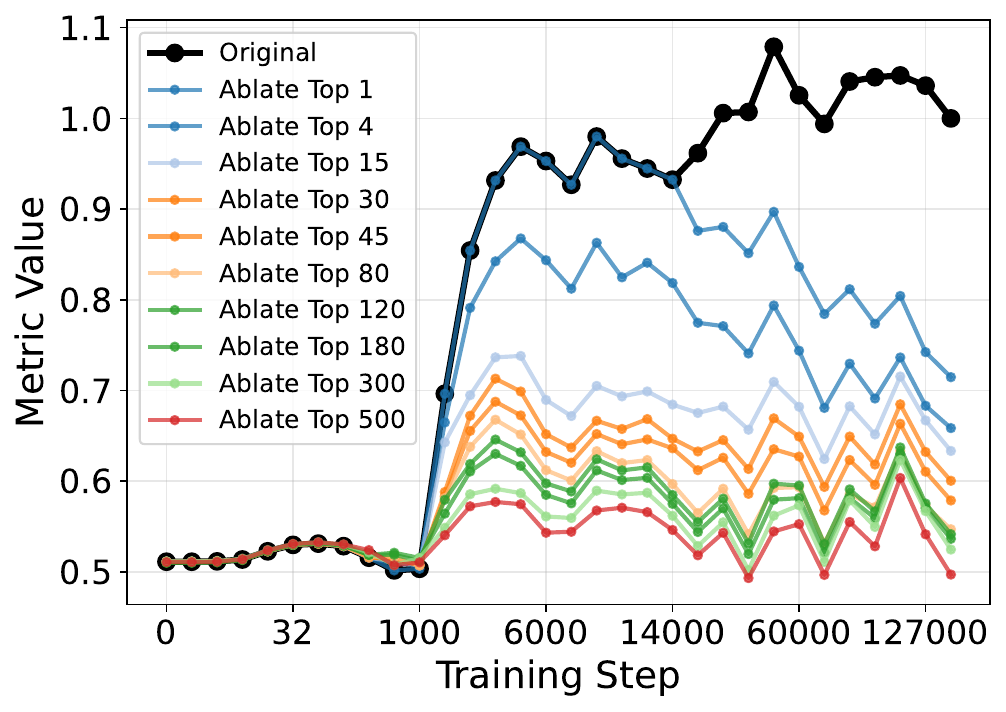}
       \caption{}
   \end{subfigure}
   \hfill
   \begin{subfigure}[b]{0.32\textwidth}
       \centering
       \includegraphics[width=\textwidth]{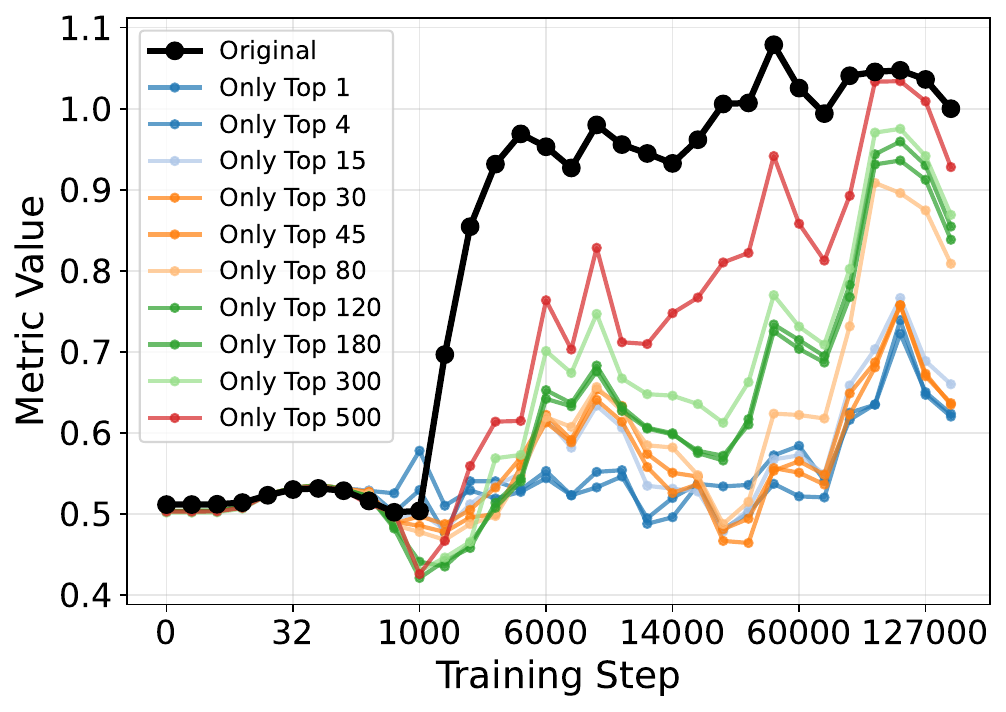}
       \caption{}
   \end{subfigure}

   \caption{Crosscoder feature attribution on the Across-RC variant of the SVA task, using a crosscoder with 98,304 features trained on Pythia-160M.}
   \label{fig:attribution-sva-rc}
\end{figure*}

\begin{figure*}[t]
   \centering
   \begin{subfigure}[b]{0.32\textwidth}
       \centering
       \includegraphics[width=\textwidth]{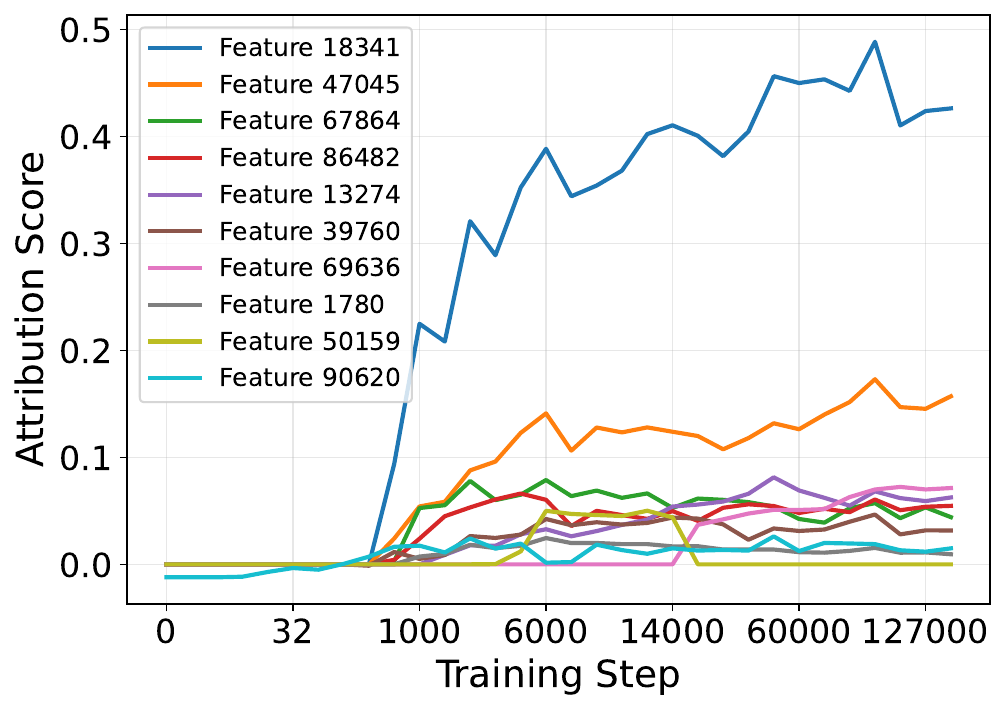}
       \caption{}
   \end{subfigure}
   \hfill
   \begin{subfigure}[b]{0.32\textwidth}
       \centering
       \includegraphics[width=\textwidth]{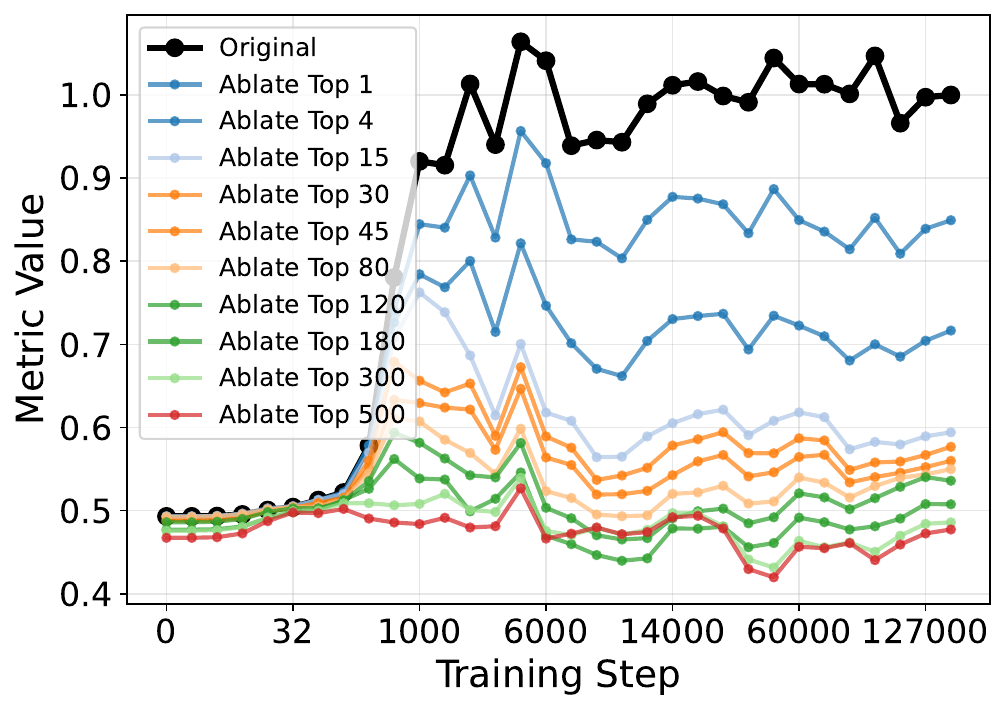}
       \caption{}
   \end{subfigure}
   \hfill
   \begin{subfigure}[b]{0.32\textwidth}
       \centering
       \includegraphics[width=\textwidth]{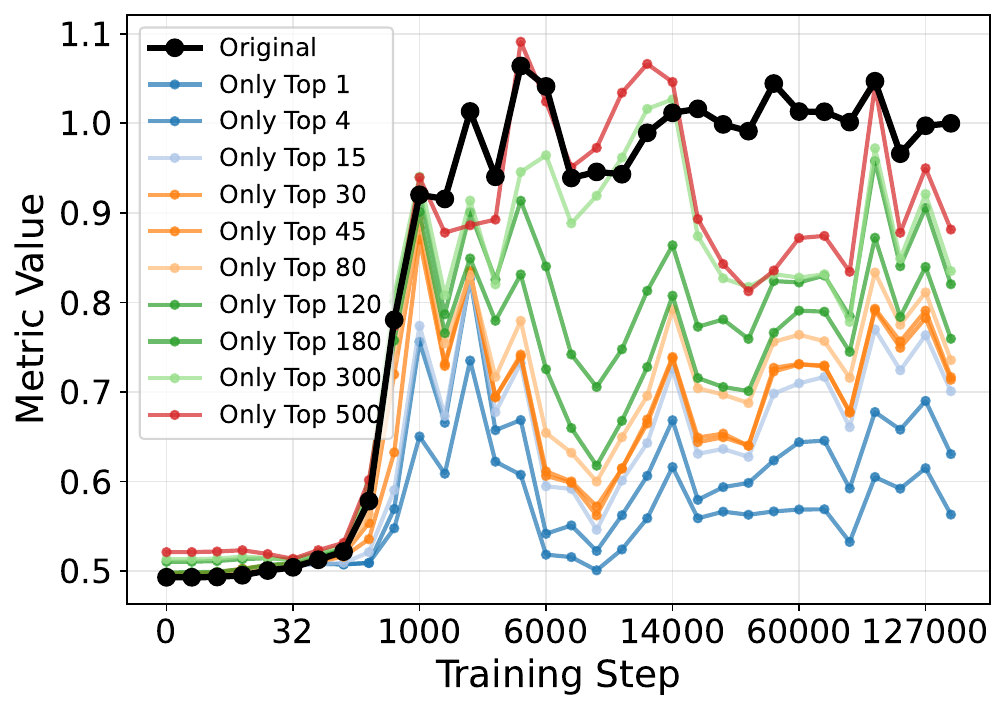}
       \caption{}
   \end{subfigure}

   \caption{Crosscoder feature attribution on the Within-RC variant of the SVA task, using a crosscoder with 98,304 features trained on Pythia-160M.}
   \label{fig:attribution-sva-within-rc}
\end{figure*}

\begin{figure*}[t]
   \centering
   \begin{subfigure}[b]{0.32\textwidth}
       \centering
       \includegraphics[width=\textwidth]{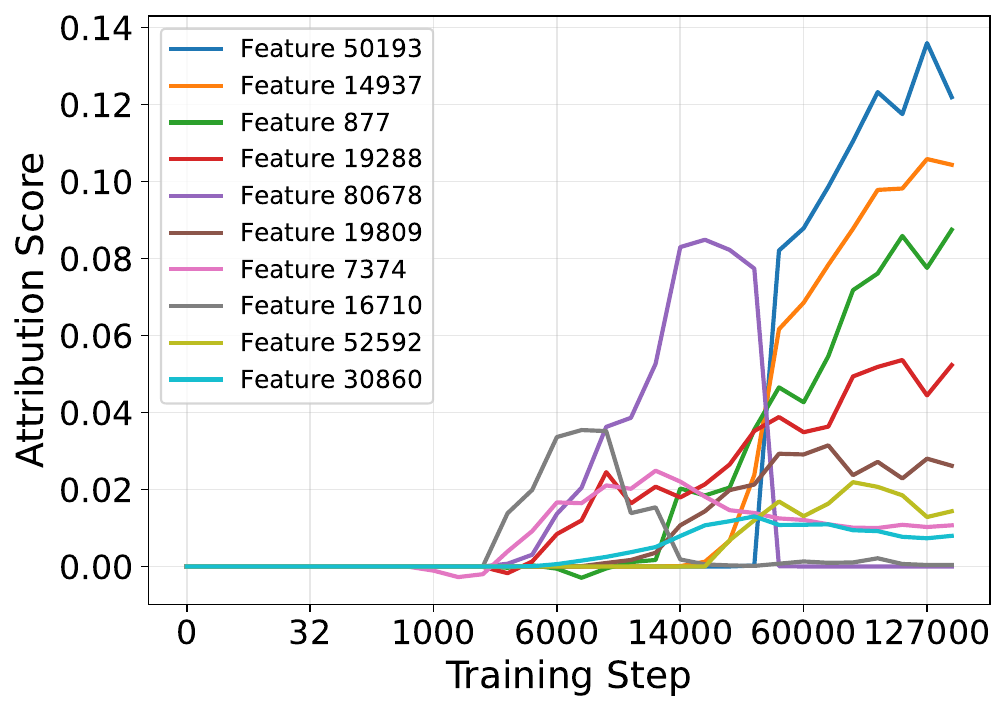}
       \caption{}
   \end{subfigure}
   \hfill
   \begin{subfigure}[b]{0.32\textwidth}
       \centering
       \includegraphics[width=\textwidth]{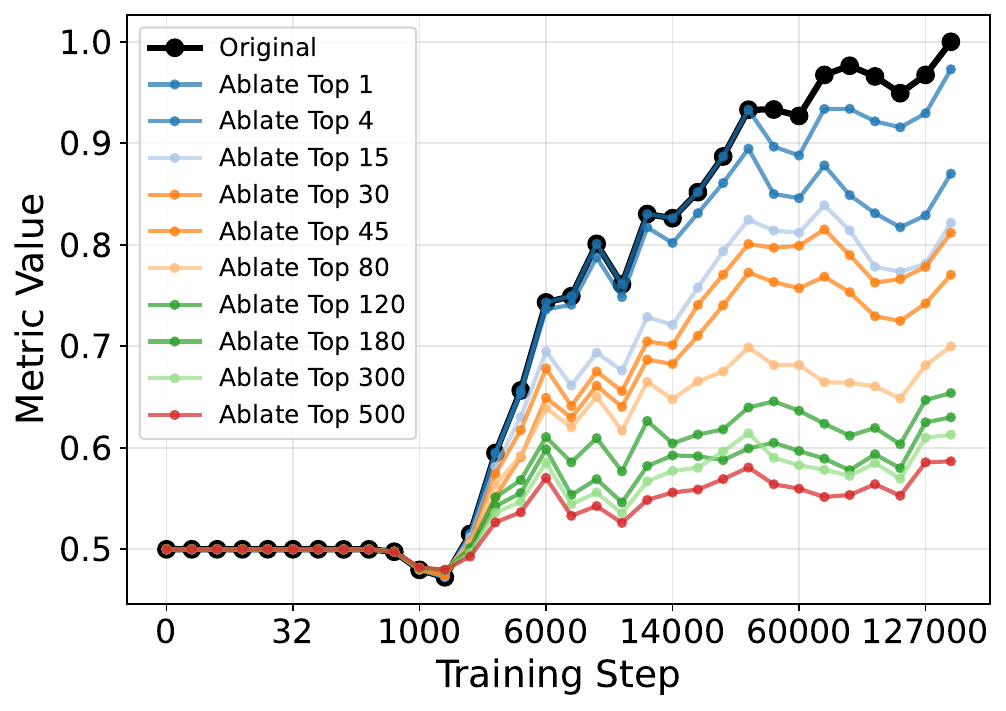}
       \caption{}
   \end{subfigure}
   \hfill
   \begin{subfigure}[b]{0.32\textwidth}
       \centering
       \includegraphics[width=\textwidth]{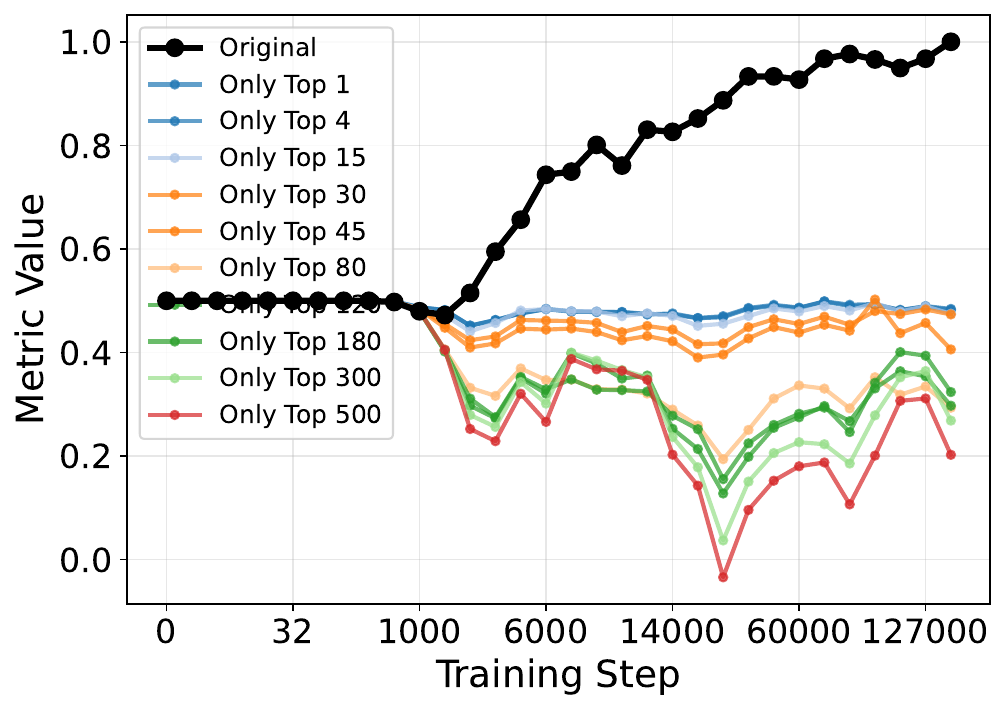}
       \caption{}
   \end{subfigure}

   \caption{Crosscoder feature attribution on the IOI task, using a crosscoder with 98,304 features trained on Pythia-160M.}
   \label{fig:attribution-ioi}
\end{figure*}

To more accurately estimate the causal effect of features, we employ the integrated gradient (IG) version of the attribution score~\citep{sundararajan2017attribution, hanna2024faithfulness, marks2025sparsefeaturecircuits}. IG attribution computes gradients along an interpolation path between baseline and target activations, providing more robust estimates than single-point gradients.

For the standard attribution score without clean/corrupted input pairs, we compute the IG version as:
\begin{equation}
    \operatorname{attr}_{\text{ig}, i}^\theta(x) = \frac{1}{N}\sum_{\alpha} f_i(x) \cdot \frac{\partial m(a^\theta(x))}{\partial (\alpha f_i(x))}
    \label{eq:attribution-ig}
\end{equation}

For attribution patching with clean/corrupted input pairs, the IG version is:
\begin{equation}
    \operatorname{attr}_{\text{ig}, i}^\theta(x,\tilde{x}) = \frac{1}{N}\sum_{\alpha} \left[f_i(x) - f_i(\tilde{x})\right] \cdot \frac{\partial m(a^\theta(x))}{\partial (\alpha f_i(x)+(1-\alpha)f_i(\tilde{x}))}
    \label{eq:attribution-patching-ig}
\end{equation}

where $\alpha \in \{0, \frac{1}{N}, \ldots, \frac{N-1}{N}\}$ linearly interpolates between the baseline and target feature activations. Consistent with \citet{marks2025sparsefeaturecircuits}, we use $N=10$ interpolation steps for the IG attribution score.

\subsection{Induction Task}

Transformers exhibit in-context learning capabilities~\citep{brown2020fewshot, zhao2021fewshot, gao2021fewshot} through induction heads~\citep{olsson2022induction}—circuits that look back over the sequence for previous instances of the current token (A), identify the subsequent token (B), and predict the same completion will occur again (forming sequences [A][B] … [A] → [B]). 

To evaluate models' induction abilities and trace them at the feature level, we construct samples with random tokens where identical patterns appear in the middle and at the end of sequences. We test whether models can correctly copy previous text as the next token. For precise feature-level analysis, we restrict next tokens to single capital letters with leading spaces. Since the induction task lacks natural corrupted counterparts, we use the log probability of the correct token as our evaluation metric. Results are shown in Figure~\ref{fig:attribution-induction}.

\subsection{Subject-Verb Agreement Tasks}

Subject-verb agreement tasks evaluate whether models can predict verbs in the appropriate grammatical form based on their subjects. We test four variants: (1) Simple—the verb directly follows the subject, e.g., ``The parents are"; (2) Across-RC—a relative clause intervenes between subject and verb, e.g., ``The athlete that the managers like does"; (3) Within-RC—both subject and verb appear within the relative clause, e.g., ``The athlete that the managers like"; (4) Across-PP—a prepositional phrase separates the subject and verb, e.g., `` The secretaries near the cars have". We use data provided by \citet{marks2025sparsefeaturecircuits}, sampling 1000 examples from each variant. We use the verb in the wrong form as the counterpart in attribution patching. Results are shown in Figure~\ref{fig:attribution}, \ref{fig:attribution-sva-simple}, \ref{fig:attribution-sva-rc}, and \ref{fig:attribution-sva-within-rc}.

\subsection{Indirect Object Identification Task}

Indirect Object Identification (IOI) evaluates whether models can correctly predict the indirect object (IO) based on repeated occurrence of subjects~\citep{wang2023ioi}. In IOI tasks, sentences such as ``When Mary and John went to the store, John gave a drink to" should be completed with ``Mary." We generate IOI samples following the same strategy as \citet{wang2023ioi}, using template sentences with random person names. We use the subject name as the corrupted counterpart in attribution patching. Results are shown in Figure~\ref{fig:attribution-ioi}.

Note that while ablating top features significantly degrades performance, we cannot recover the original metric using only these top features. This likely occurs because IOI is a complex task requiring multiple feature interactions. The features that distinguish between indirect objects and subjects represent only part of the full computational requirements, making isolated feature sets insufficient for complete task execution.

\subsection{Random Baseline of Ablation}

\begin{figure*}
   \centering
   \begin{subfigure}[b]{0.32\textwidth}
       \centering
       \includegraphics[width=\textwidth]{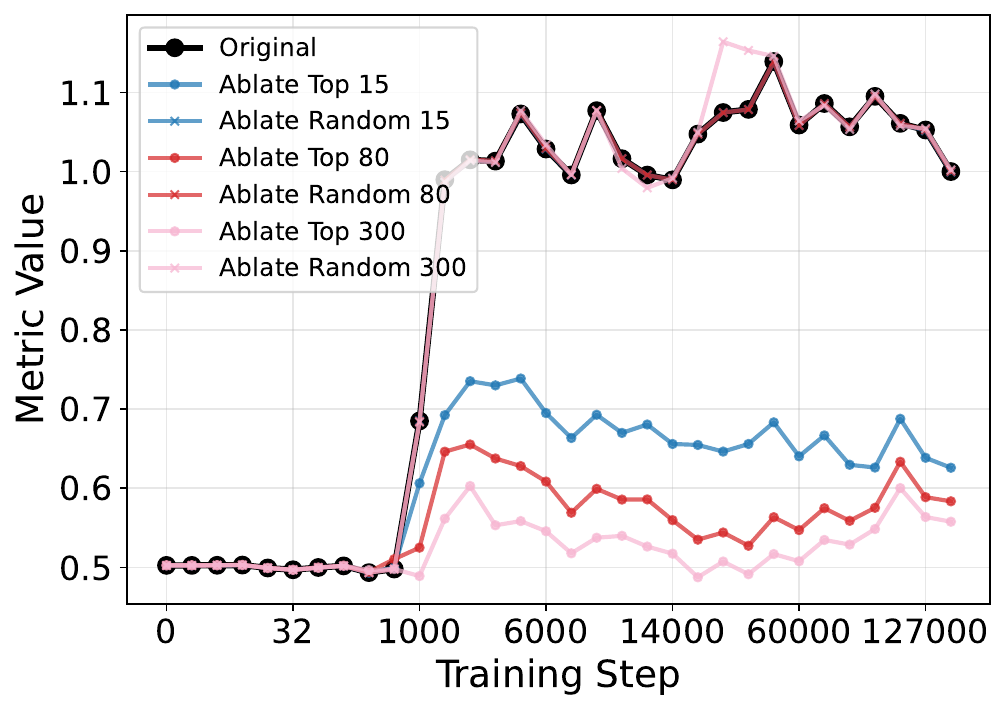}
       \caption{}
   \end{subfigure}
   \begin{subfigure}[b]{0.32\textwidth}
       \centering
       \includegraphics[width=\textwidth]{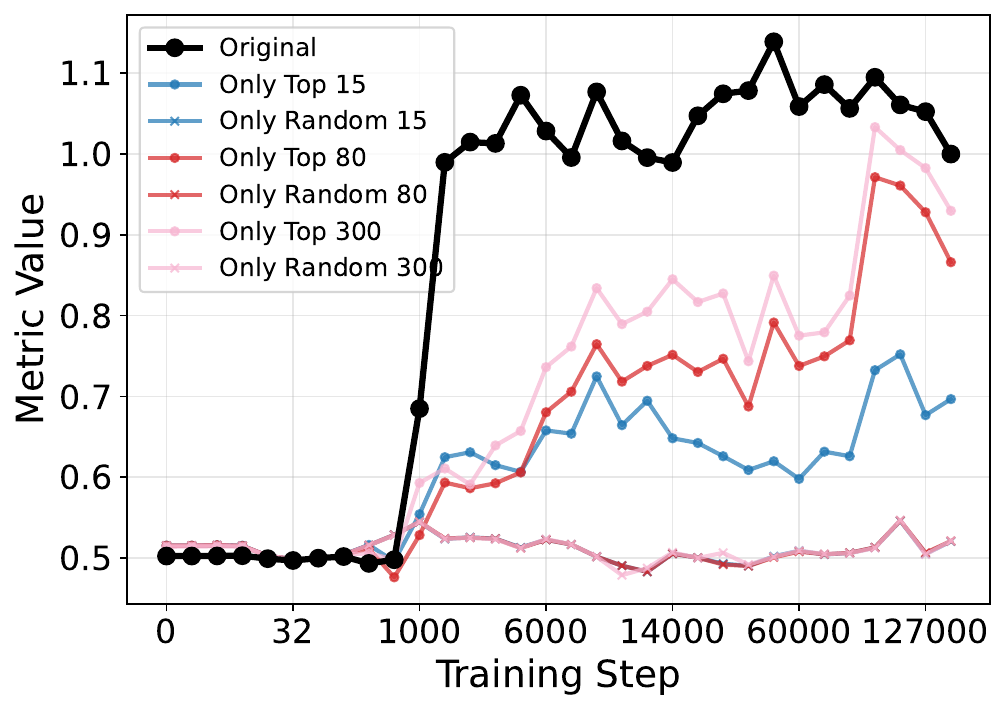}
       \caption{}
   \end{subfigure}

   \caption{Crosscoder feature attribution on the Across-PP variant of the SVA task, with random baselines.}
   \label{fig:random-baseline}
\end{figure*}

In Figure~\ref{fig:random-baseline}, we add baselines to Figure~\ref{fig:attribution} by ablating/preserving $k$ random features. The results show that our feature selected by attribution patching is far more effective.

\section{Crosscoder Feature Splitting}
\label{appendix:feature-splitting}

Feature splitting is a well-known phenomenon in SAEs where increasing the dictionary size $n_\text{features}$ causes features from smaller SAEs to fragment into multiple distinct features. Under feature splitting, a single concept may not be represented by one feature, but rather by multiple specialized features that activate on the same concept in different contexts.

We observe similar feature splitting phenomenon in crosscoders, but across the temporal dimension. We find in rare cases, features with similar semanticity and direction in different snapshots may not be encoded in the same latent of crosscoder feature space, but result in separate features.

\begin{figure*}[t]
    \centering
    \includegraphics[width=\textwidth]{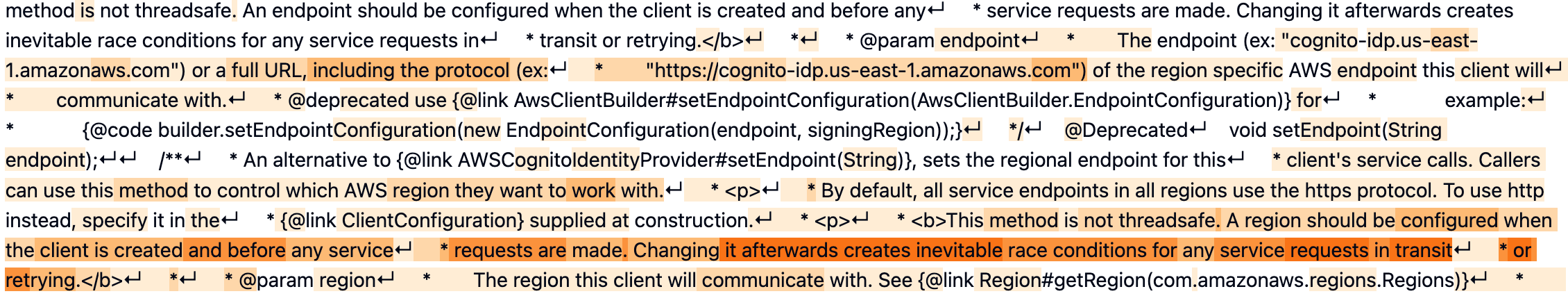}
    \caption{Top activation of Feature 66688 in the 98,304-feature crosscoder on Pythia-160M.}
    \label{fig:feature-activation-66688}
\end{figure*}

\begin{figure*}[t]
    \centering
    \includegraphics[width=\textwidth]{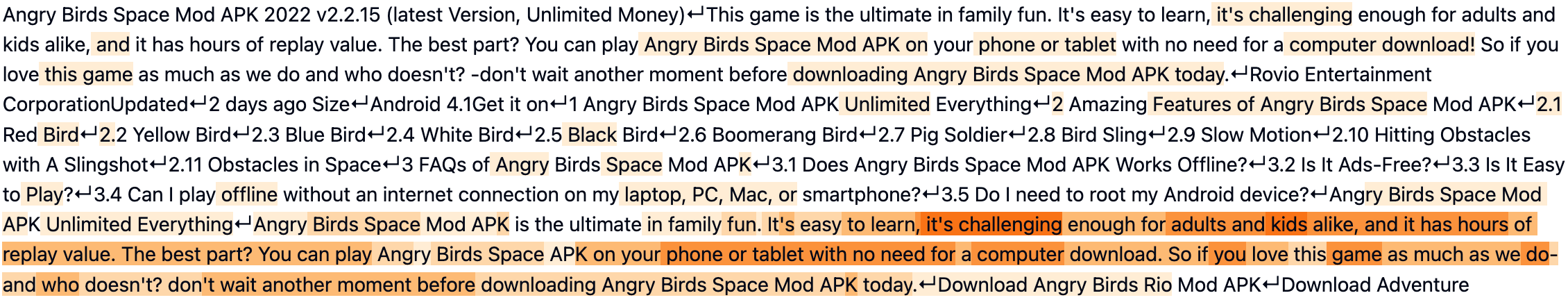}
    \caption{Top activation of Feature 53542 in the 98,304-feature crosscoder on Pythia-160M.}
    \label{fig:feature-activation-53542}
\end{figure*}

\begin{figure*}[t]
    \centering
    \includegraphics[width=\textwidth]{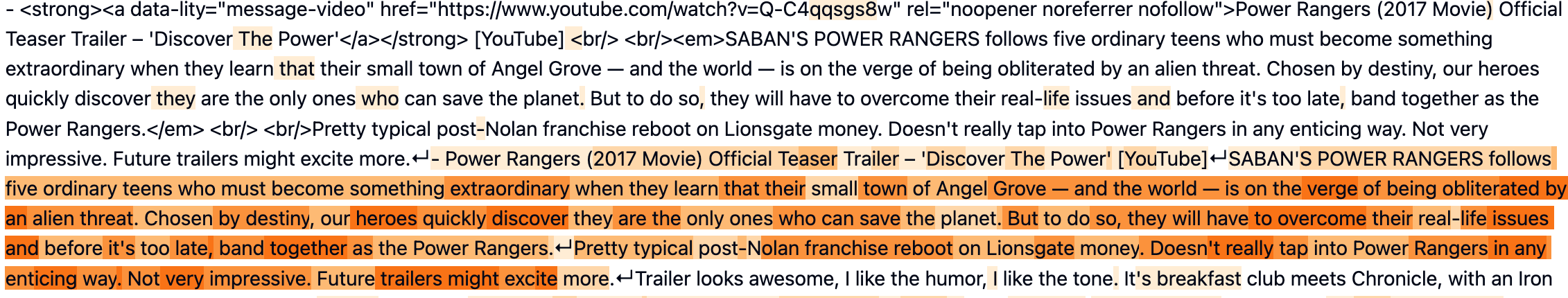}
    \caption{Top activation of Feature 42307 in the 98,304-feature crosscoder on Pythia-160M.}
    \label{fig:feature-activation-42307}
\end{figure*}

\begin{figure*}[t]
    \centering
    \includegraphics[width=\textwidth]{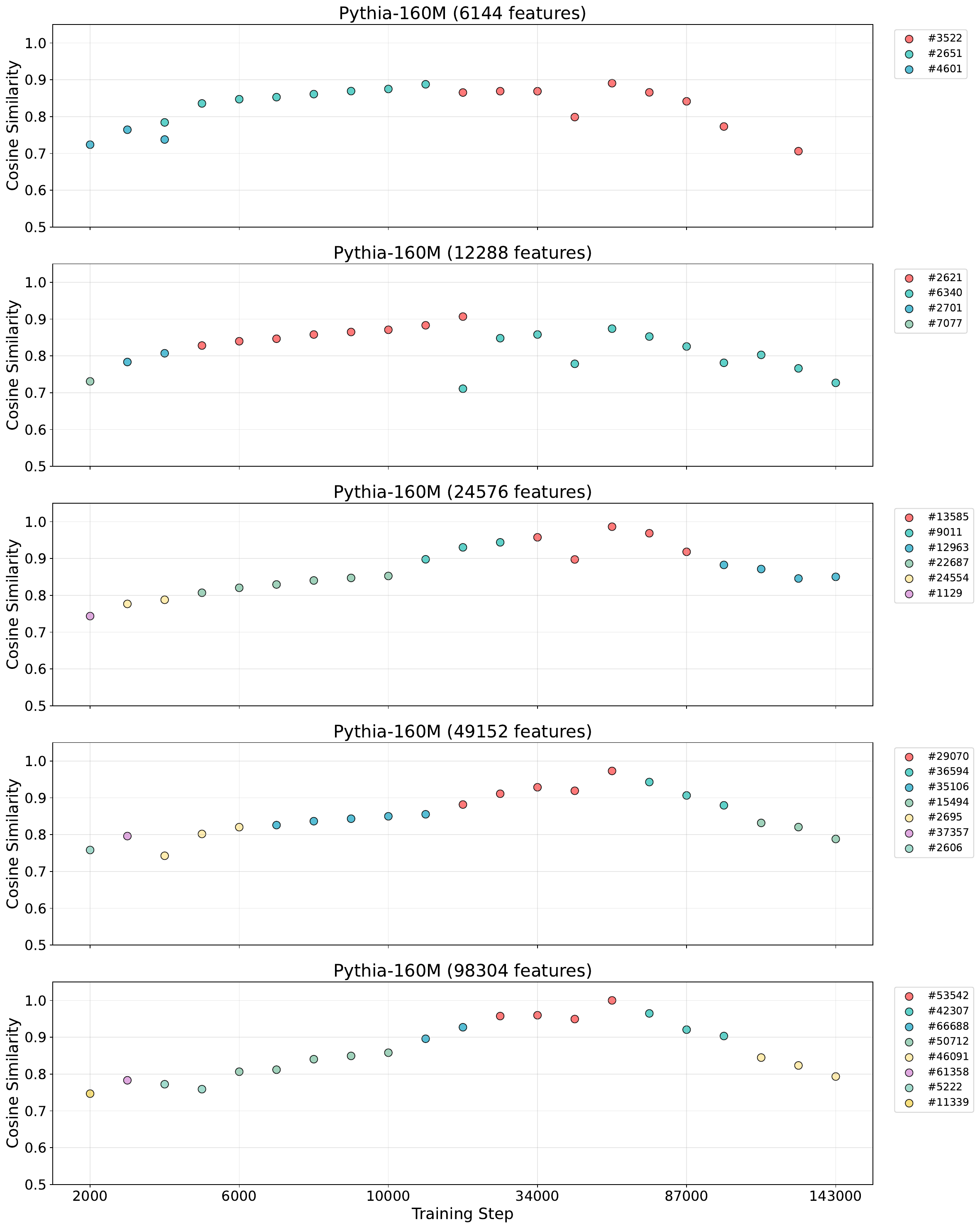}
    \caption{Crosscoder features similar to Feature 53542 of a 98,304-feature crosscoder. The number of features split increases with dictionary size.}
    \label{fig:feature-splitting}
\end{figure*}

\begin{figure*}[t]
    \centering
    \includegraphics[width=0.8\textwidth]{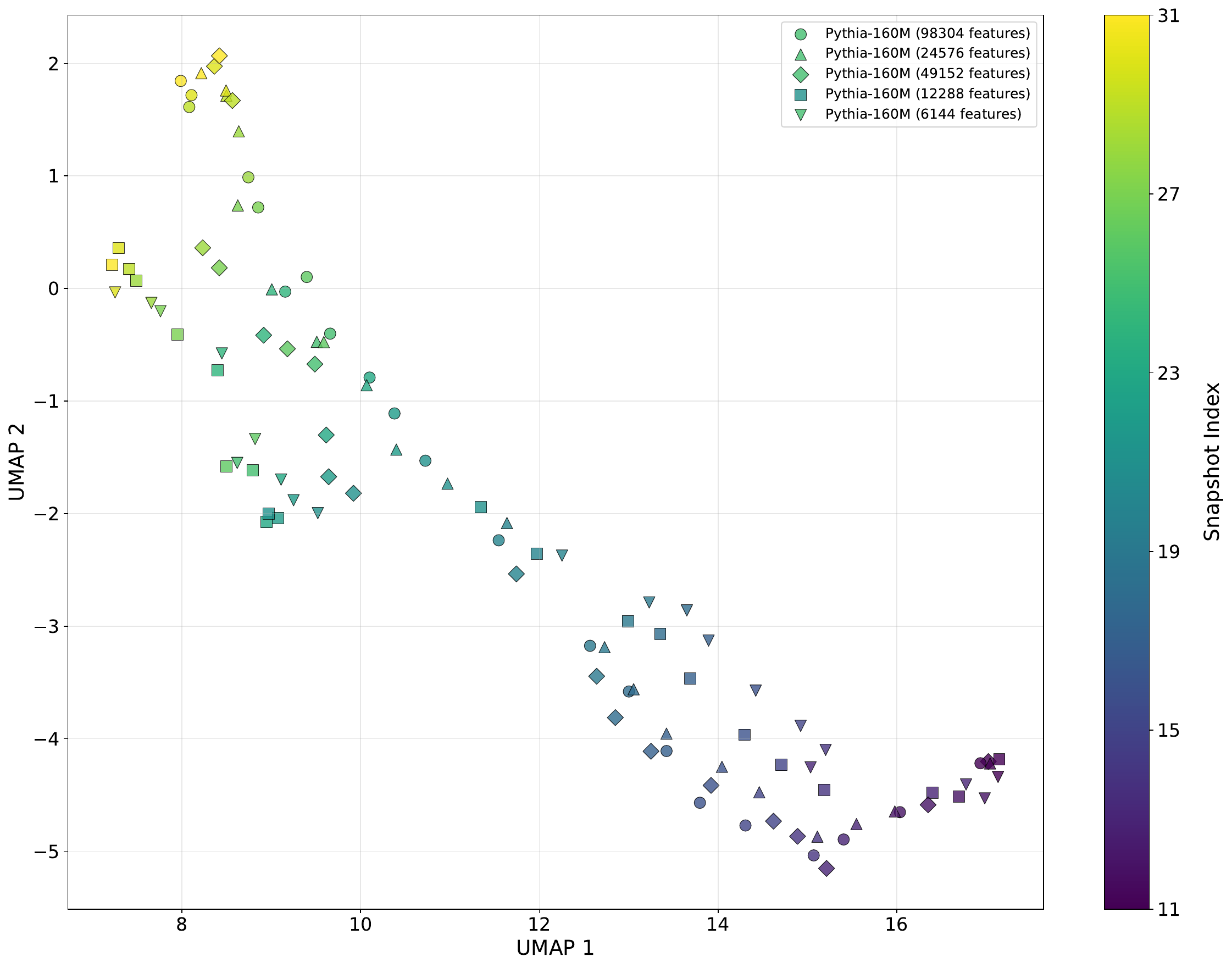}
    \caption{UMAP visualization of features shown in Figure~\ref{fig:feature-splitting}.}
    \label{fig:feature-splitting-umap}
\end{figure*}

For example, we identify feature 66688, 53542, and 42307 that all activate on long sequences containing repeated text patterns, contributing to the induction task (Figures~\ref{fig:feature-activation-66688}, \ref{fig:feature-activation-53542}, and \ref{fig:feature-activation-42307}). Each feature exists only during non-overlapping continuous periods across snapshots, with a drastic emergence and disappearance.

To examine whether feature splitting across snapshots relates to dictionary size, we search for feature decoder vectors (across all snapshots) with cosine similarity above 0.7—a high threshold in such high-dimensional space—in crosscoders with varying $n_\text{features}$. Across dictionary sizes ranging from 6,144 to 98,304 features, we observe that while each snapshot after step 2,000 activates almost exactly one feature representing this concept, the total number of distinct crosscoder features increases with dictionary size (Figure~\ref{fig:feature-splitting}). This confirms that larger dictionaries lead to temporal feature splitting, where a single underlying concept splits into multiple features active at different training stages.

We also observe that temporal feature splitting predominantly occurs among densely activating features, i.e., features that activate frequently across many contexts. This crosscoder feature splitting likely arises from subtly different activation patterns that emerge over training. These findings suggest that language model features may evolve by refining their activation patterns, leading to more specialized representations that warrant separate feature assignments at different training stages.

\section{Generalizability Studies}

In this section, we include more experiments on training and analyzing crosscoders on different base models/initialization/datasets and different numbers of snapshot selection, to examine the generalizablity of our methods.

\subsection{Generalization across Model Families/Seeds}

\begin{figure*}
   \centering
   \includegraphics[width=\textwidth]{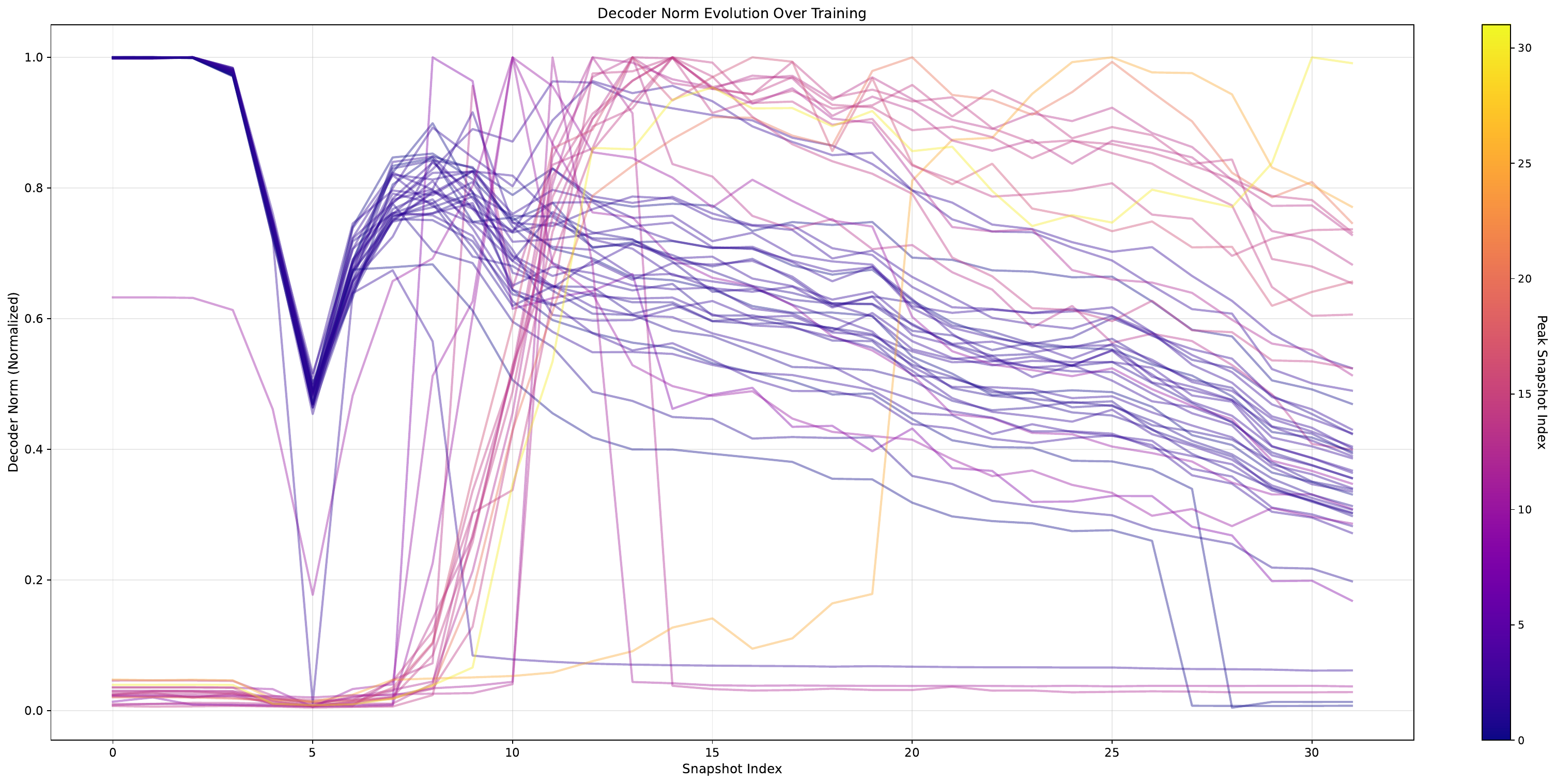}
   \caption{Overview of cross-snapshot feature decoder norm evolution in Pythia-160M Seed1.}
   \label{fig:norm-evolution-pythia-seed1}
\end{figure*}

\begin{figure*}
   \centering
   \includegraphics[width=\textwidth]{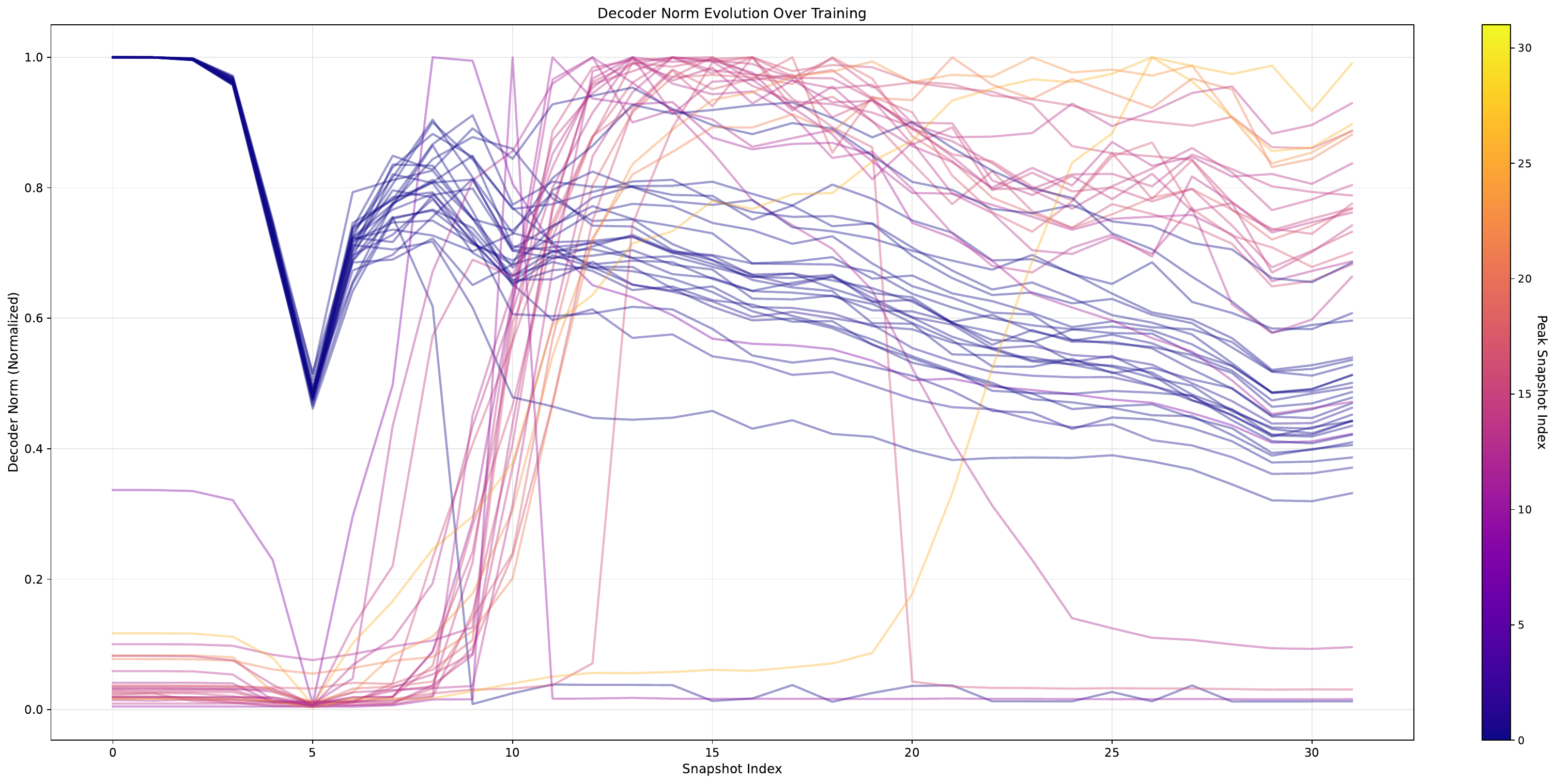}
   \caption{Overview of cross-snapshot feature decoder norm evolution in Pythia-160M Seed2.}
   \label{fig:norm-evolution-pythia-seed2}
\end{figure*}

\begin{figure*}
   \centering
   \includegraphics[width=\textwidth]{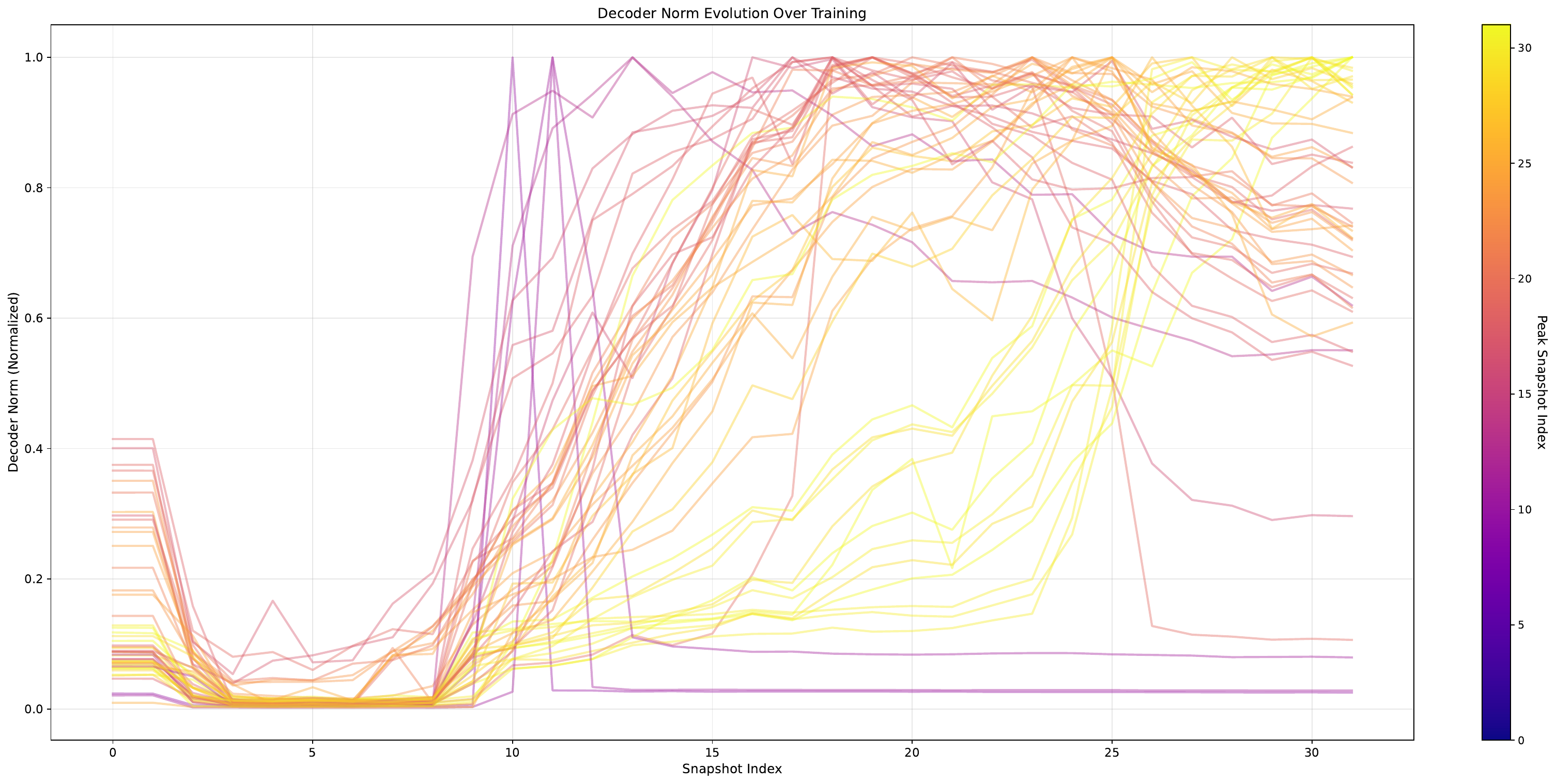}
   \caption{Overview of cross-snapshot feature decoder norm evolution in Alias run of Stanford CRFM GPT-2.}
   \label{fig:norm-evolution-stanford-crfm}
\end{figure*}

\begin{figure*}
   \centering
   \begin{subfigure}[b]{0.32\textwidth}
       \centering
       \includegraphics[width=\textwidth]{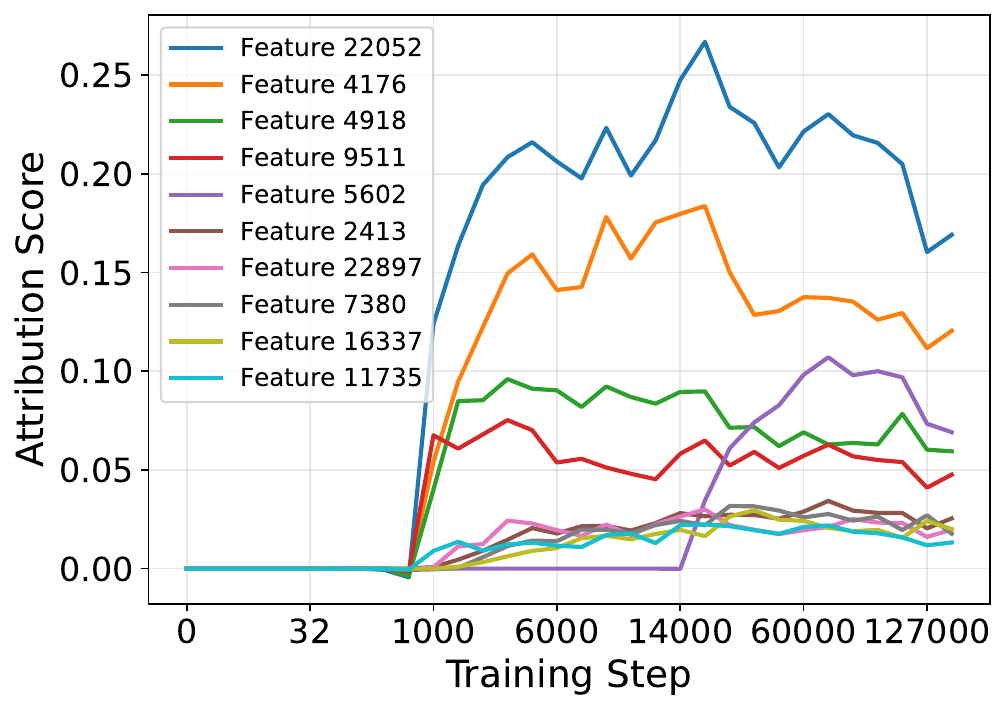}
       \caption{}
   \end{subfigure}
   \hfill
   \begin{subfigure}[b]{0.32\textwidth}
       \centering
       \includegraphics[width=\textwidth]{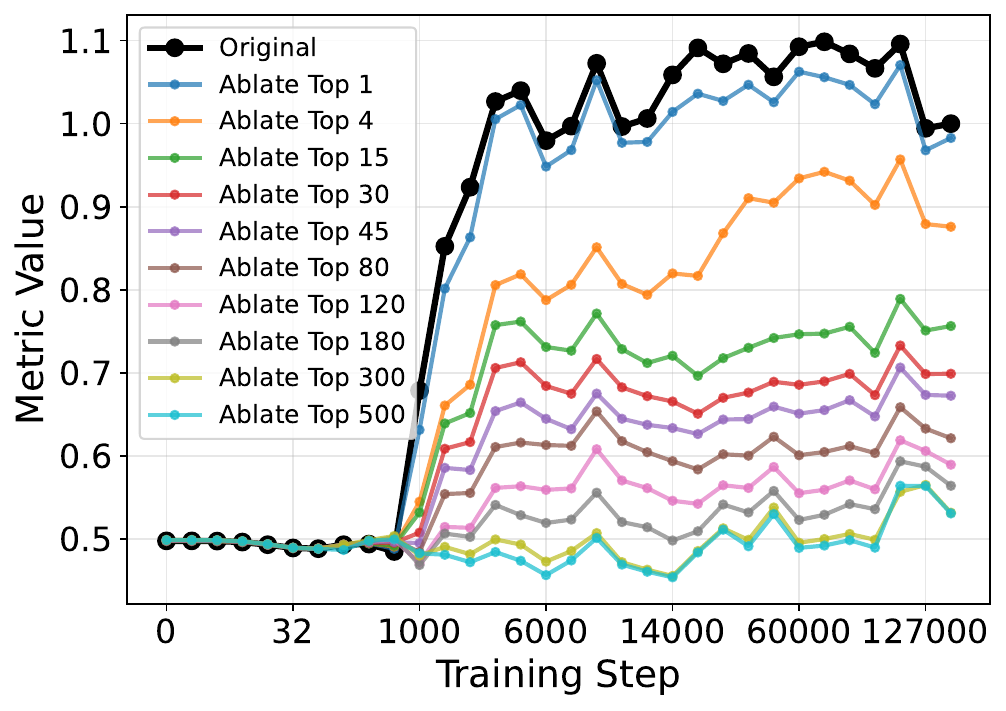}
       \caption{}
   \end{subfigure}
   \hfill
   \begin{subfigure}[b]{0.32\textwidth}
       \centering
       \includegraphics[width=\textwidth]{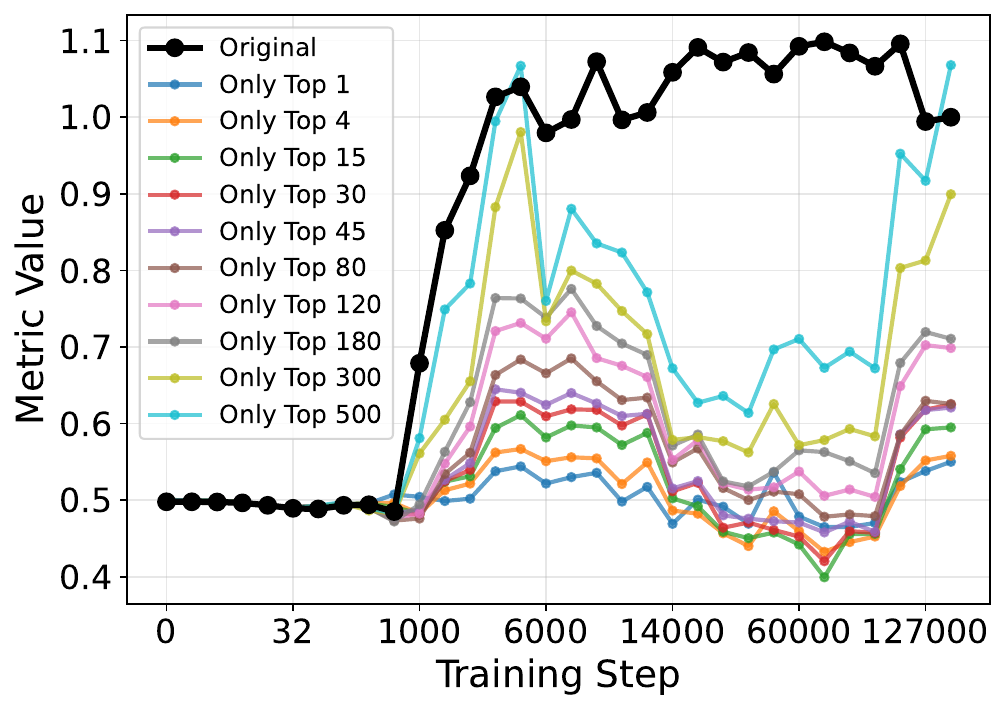}
       \caption{}
   \end{subfigure}

   \caption{Crosscoder feature attribution on the Across-PP variant of the SVA task, using a crosscoder with 24,576 features trained on Pythia-160M Seed1.}
   \label{fig:attribution-pythia-seed1}
\end{figure*}

\begin{figure*}
   \centering
   \begin{subfigure}[b]{0.32\textwidth}
       \centering
       \includegraphics[width=\textwidth]{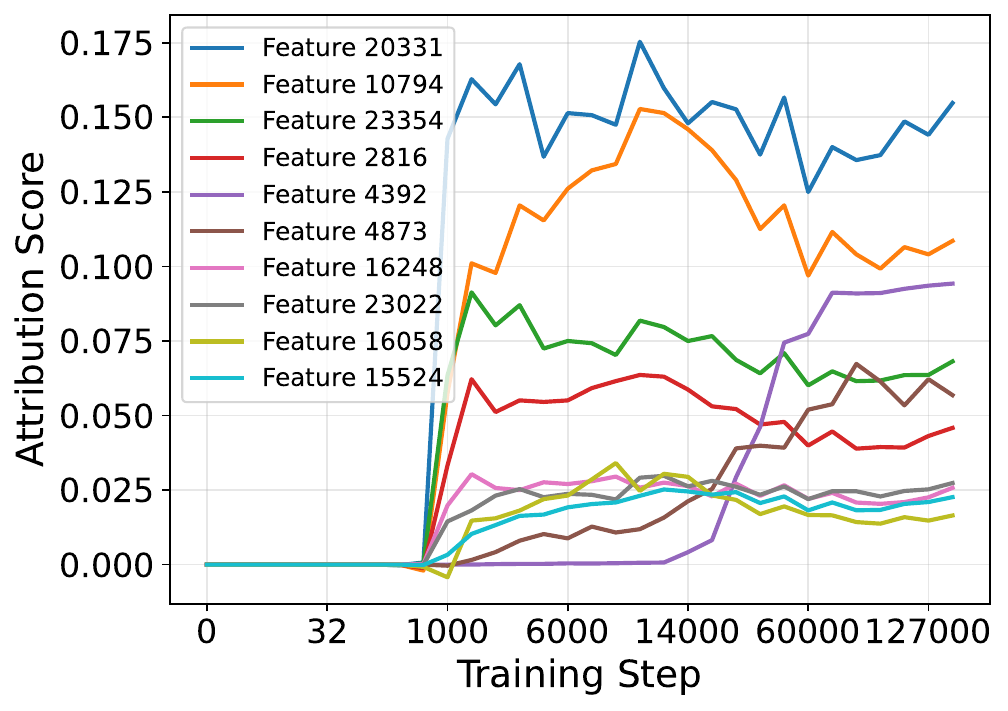}
       \caption{}
   \end{subfigure}
   \hfill
   \begin{subfigure}[b]{0.32\textwidth}
       \centering
       \includegraphics[width=\textwidth]{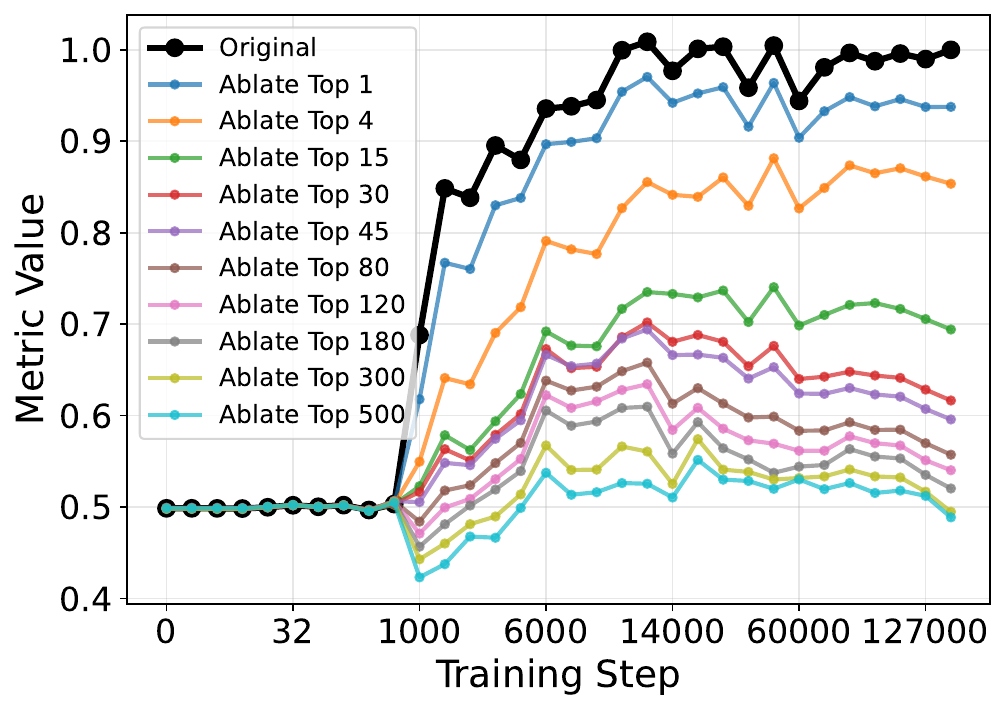}
       \caption{}
   \end{subfigure}
   \hfill
   \begin{subfigure}[b]{0.32\textwidth}
       \centering
       \includegraphics[width=\textwidth]{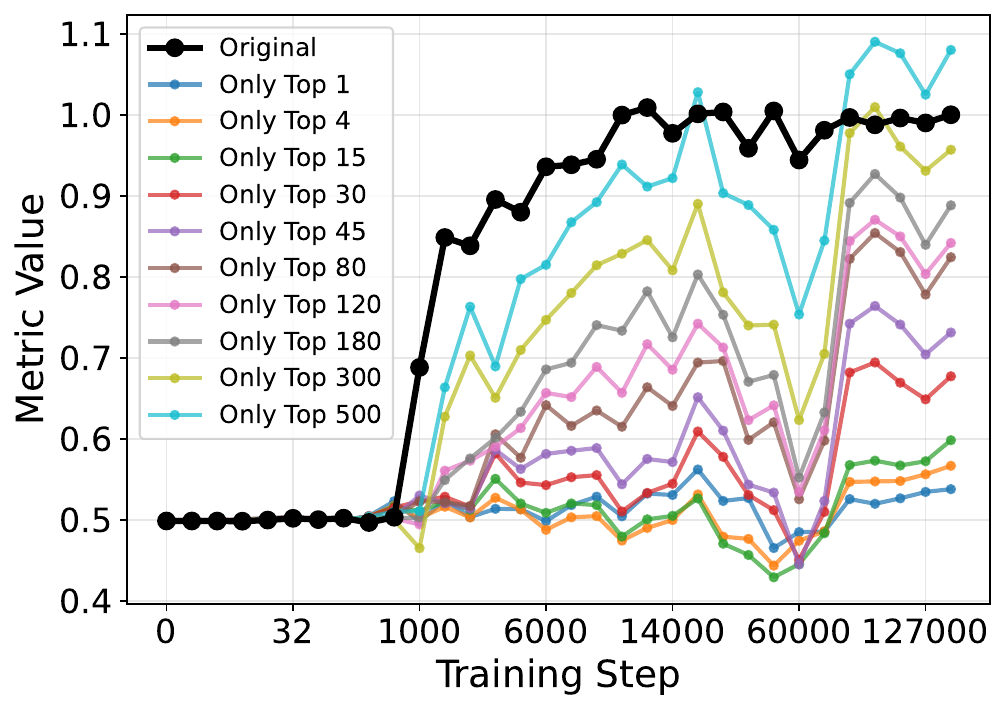}
       \caption{}
   \end{subfigure}

   \caption{Crosscoder feature attribution on the Across-PP variant of the SVA task, using a crosscoder with 24,576 features trained on Pythia-160M Seed2.}
   \label{fig:attribution-pythia-seed2}
\end{figure*}

\begin{figure*}
   \centering
   \begin{subfigure}[b]{0.32\textwidth}
       \centering
       \includegraphics[width=\textwidth]{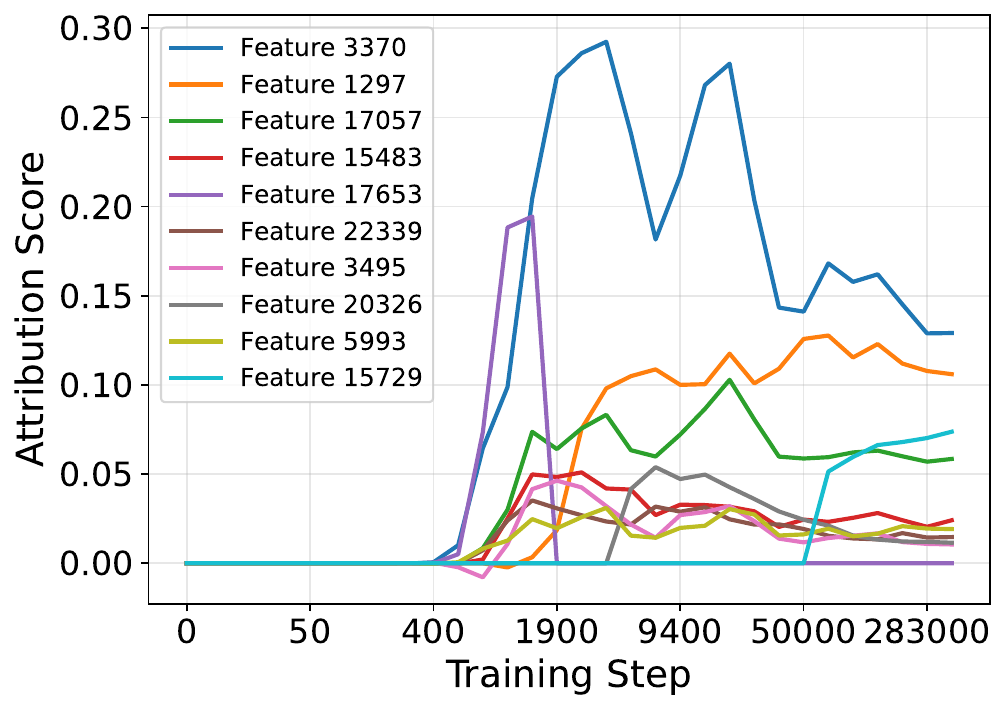}
       \caption{}
   \end{subfigure}
   \hfill
   \begin{subfigure}[b]{0.32\textwidth}
       \centering
       \includegraphics[width=\textwidth]{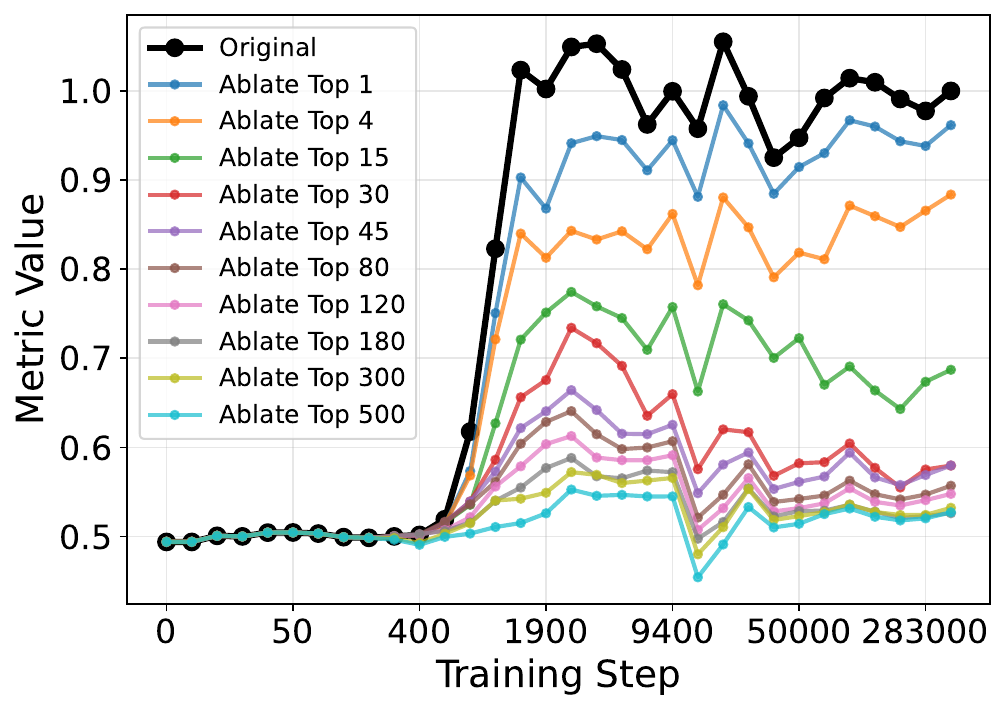}
       \caption{}
   \end{subfigure}
   \hfill
   \begin{subfigure}[b]{0.32\textwidth}
       \centering
       \includegraphics[width=\textwidth]{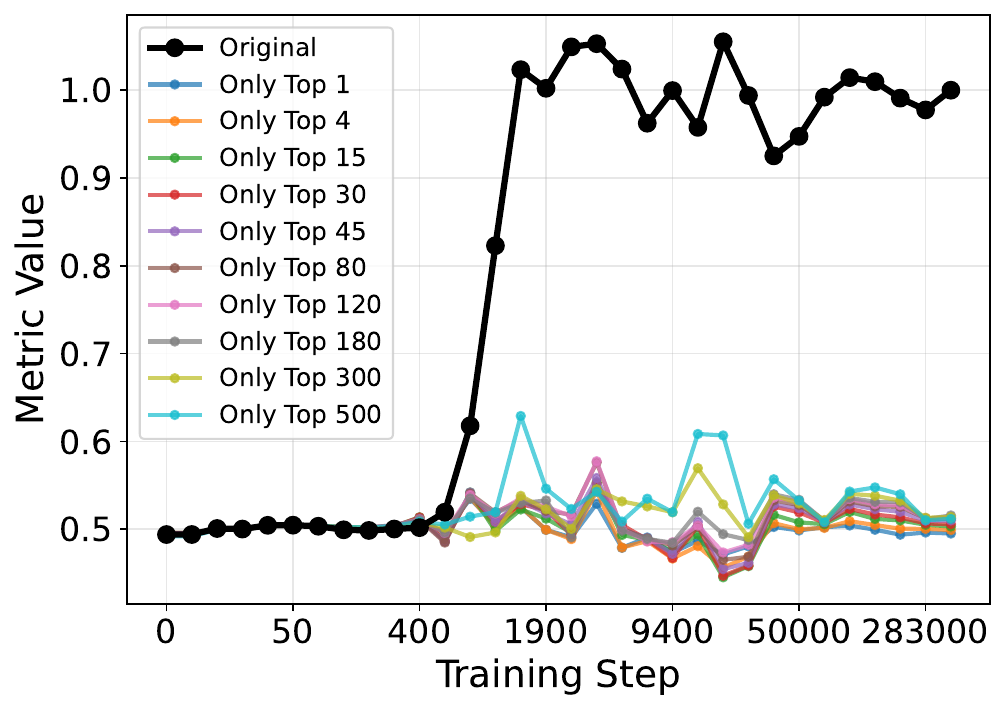}
       \caption{}
   \end{subfigure}

   \caption{Crosscoder feature attribution on the Across-PP variant of the SVA task, using a crosscoder with 24,576 features trained on the Alias run of Stanford CRFM GPT-2.}
   \label{fig:attribution-stanford-crfm}
\end{figure*}

We conduct experiments on two additional Pythia 160M models with different random seeds (Pythia 160M Seed1 and Seed2)~\citep{vanderwal2025polypythias} and the Alias run of Stanford CRFM's GPT-2 model~\footnote{Obtained from \url{https://huggingface.co/stanford-crfm/alias-gpt2-small-x21}.}. All of these models have an activation space of 768 dimensions. We train crosscoders of 24,576 features (32x) on Layer 6 of them, and inspect their feature evolution.

\paragraph{Different initializations (Pythia 160M variants).} We observe highly consistent results across differently initialized versions of Pythia 160M. Decoder norms of initialization features follow identical trajectories, and emergent features begin rising around step 1000, exhibiting a clear two-phase pattern with the same transition point (Figure~\ref{fig:norm-evolution-pythia-seed1} and~\ref{fig:norm-evolution-pythia-seed2}). Attribution experiments confirm that we successfully identify sparse crosscoder features with significant contributions to downstream tasks (Figure~\ref{fig:attribution-pythia-seed1} and~\ref{fig:attribution-pythia-seed2}).

\paragraph{Different base model (GPT-2).} For Stanford CRFM's GPT-2 model, we observe a clear two-phase pattern, with emergent features rising around step 1,000. Upon manual inspection of these features, we still find that initialization features exhibit token-level information, while more complex patterns emerge in emergent features.

However, we find a notable difference: due to absolute positional embeddings, initialization feature norms do not peak at the beginning but rather at later steps (with above-threshold norm at beginning). This difference arises because the absolute positional embeddings initially exhibit large norms (larger than the word embeddings) and dominate the activation norm. Since we normalize all activation sources to the same average activation norm $\sqrt{d_\text{model}}$, this large positional contribution substantially reduces the observed norms of initialization features. (Figure~\ref{fig:norm-evolution-stanford-crfm}).

To sum up, the statistical-to-feature-learning phase transition still exists across models, despite small differences. Attribution experiments (Figure~\ref{fig:attribution-stanford-crfm}) again confirm the effectiveness of sparse crosscoder features, with generally similar feature evolution patterns for specific downstream tasks.

\paragraph{Other models with accessible checkpoints.} We also investigated other model families, including BLOOM~\citep{bigscience2024bloom} and OLMo~\citep{olmo20242olmo2furious}. Unfortunately, these models either lack sufficient intermediate checkpoints for fine-grained analysis of pretraining dynamics (BLOOM), or lack sufficiently early checkpoints to capture the two-stage transition (BLOOM and OLMo).

\subsection{Generalizability across Datasets}

\begin{figure*}
   \centering
   \includegraphics[width=\textwidth]{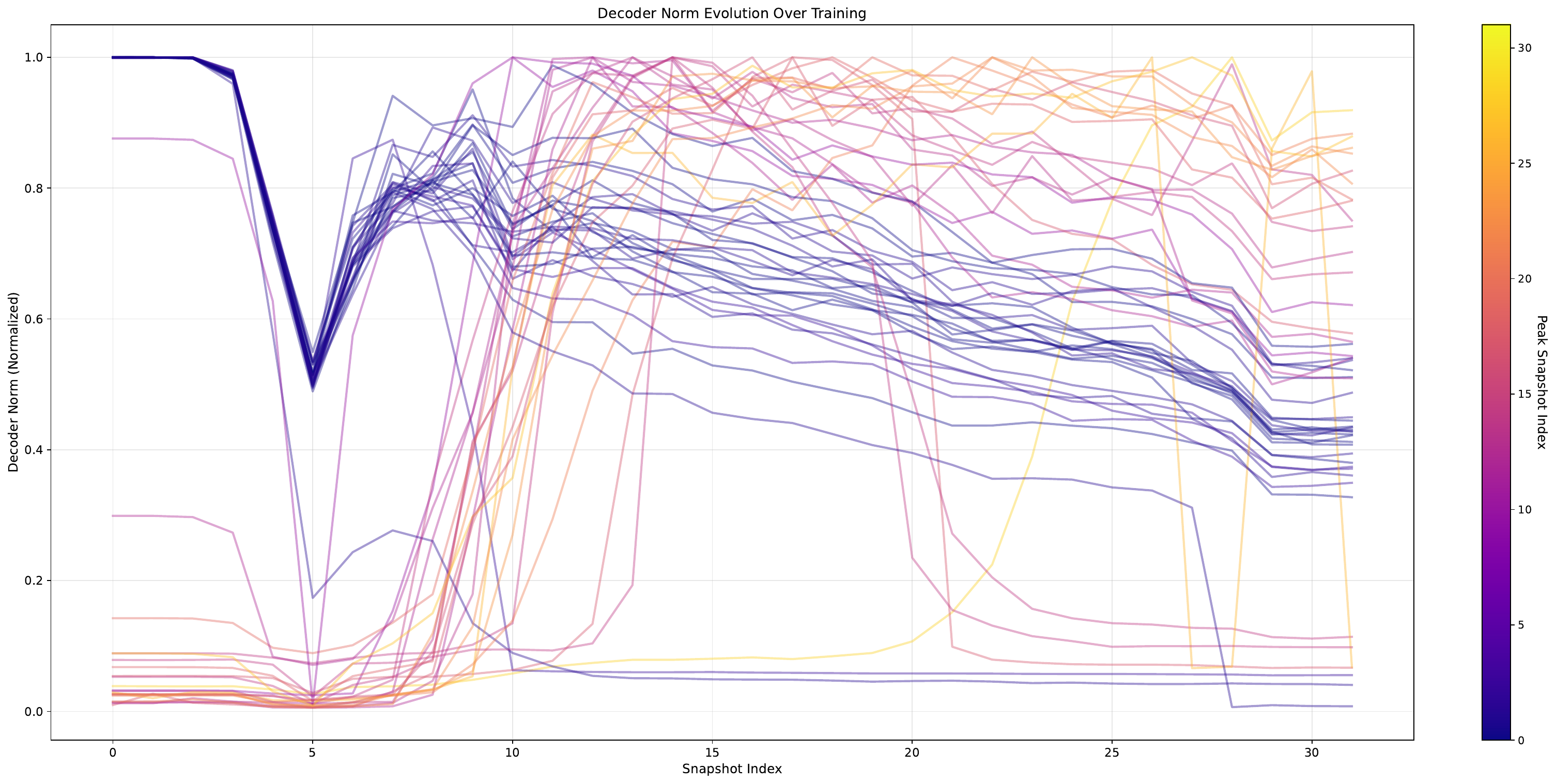}
   \caption{Overview of cross-snapshot feature decoder norm evolution in Pythia-160M, trained on FineWeb-Edu dataset.}
   \label{fig:norm-evolution-fineweb}
\end{figure*}

We conduct experiments on the original Pythia 160M snapshots, but train and analyze 24,576-feature crosscoders on the FineWeb-Edu dataset~\citep{penedo2024fineweb}, which consists of educational webpages. Our results (Figure~\ref{fig:norm-evolution-fineweb}) show nearly identical patterns to those observed with Pythia 160M crosscoders trained on SlimPajama, exhibiting very similar feature evolution dynamics.

\subsection{Sensitivity to Snapshot Selection}

\begin{figure*}
   \centering
   \includegraphics[width=\textwidth]{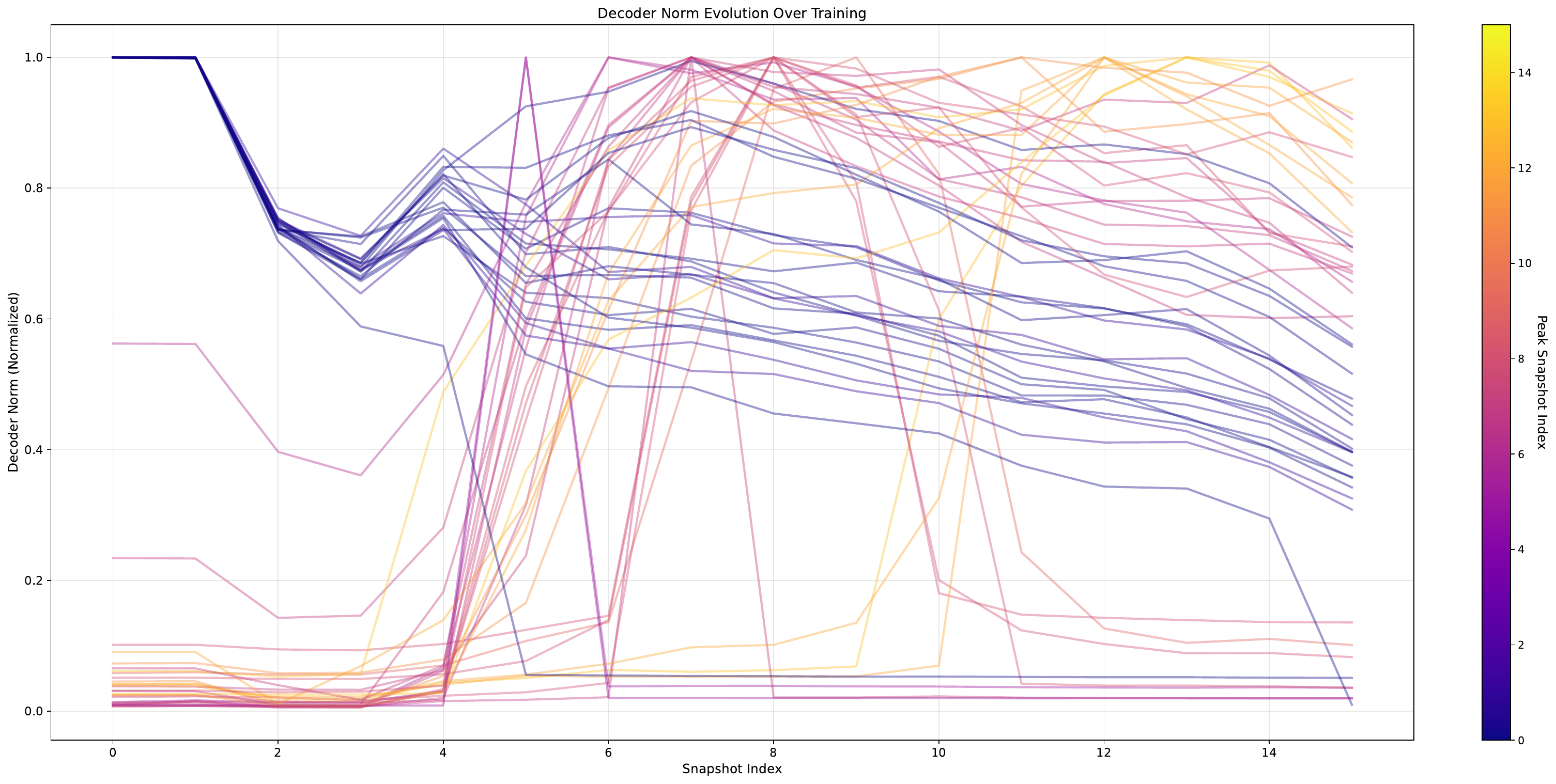}
   \caption{Overview of cross-snapshot feature decoder norm evolution in Pythia-160M, trained on 16 snapshots.}
   \label{fig:norm-evolution-16heads}
\end{figure*}

All of our above experiments train crosscoders on 32 pre-training snapshots of the language model. To further examine whether snapshot selection affects the trained crosscoder and the captured feature evolution, we train a 24,576-feature crosscoder on 16 pre-training snapshots of Pythia 160M, downsampled from the original set by selecting every other snapshot. The result (Figure~\ref{fig:norm-evolution-16heads}) shows that the overall feature evolution exhibits nearly identical trends, with only a reduction in temporal resolution.

\section{More Examples of Feature}

\begin{figure*}[p]
    \centering
    \includegraphics[width=\textwidth]{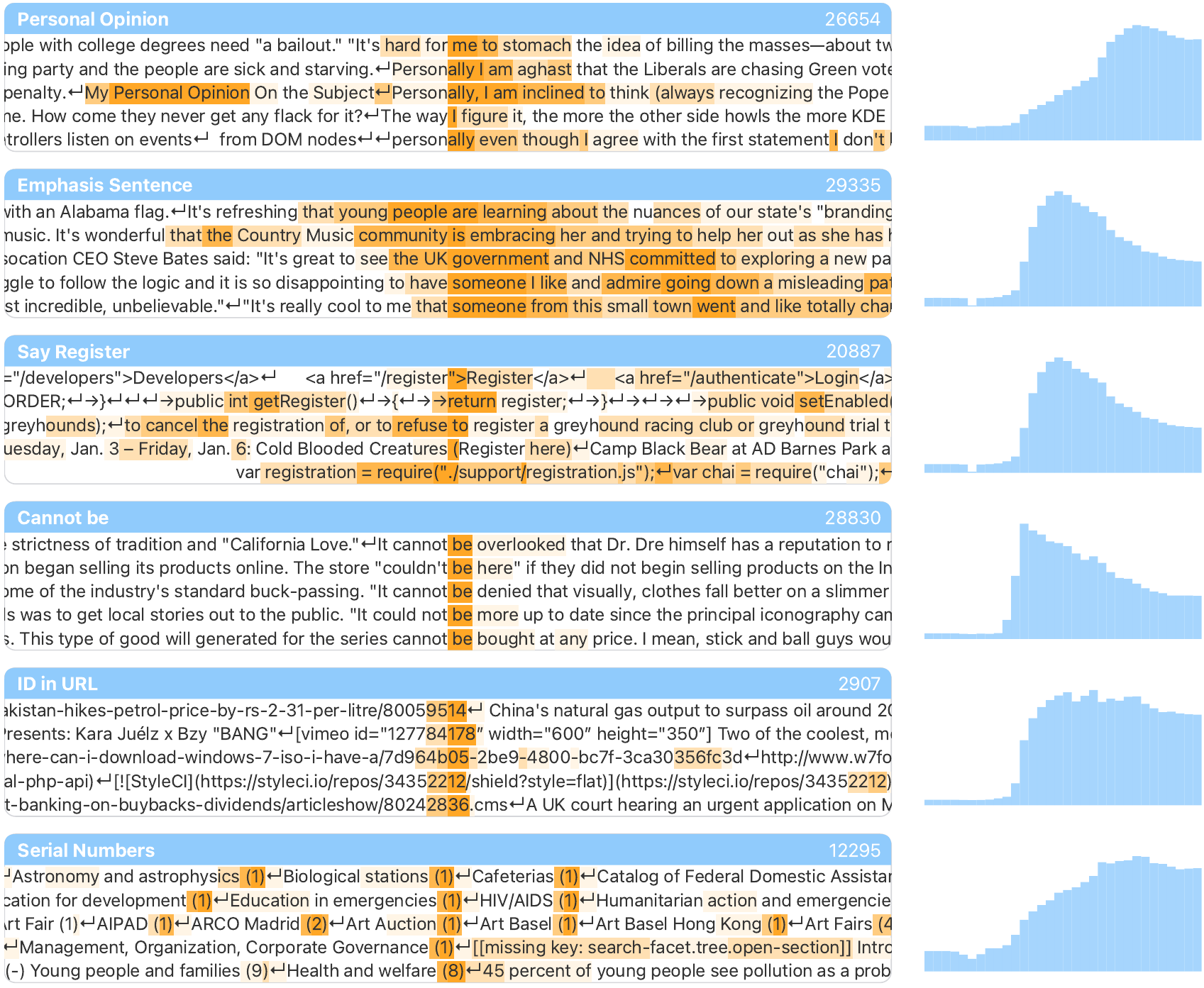}
    \caption{More features in the 32,768-feature crosscoder on Pythia-6.9B}
    \label{fig:more-feature-cards}
\end{figure*}

We additionally list some emergent features and demonstrate their top activating samples and decoder norm evolution in Figure~\ref{fig:more-feature-cards}.

\section{The Use of Large Language Models}

The use of Large Language Models as an assistive tool in this paper is limited to the following two aspects:

\begin{enumerate}
    \item The automated generation of feature complexity scores, as detailed in Section~\ref{sec:complexity} and Appendix~\ref{appendix:complexity}.
    \item Grammar correction and stylistic refinement of the manuscript.
\end{enumerate}

\end{document}